%% file: ms.tex
\pdfoutput=1 
\documentclass[10pt,twocolumn,letterpaper]{article}

\usepackage{wacv}
\usepackage{times}
\usepackage{epsfig}
\usepackage{graphicx}
\usepackage{amsmath}
\usepackage{amssymb}
\usepackage{booktabs}

\usepackage{svg}
\usepackage{tikz}
\usepackage{xcolor}
\usepackage{array,multirow,textcomp}
\usepackage{makecell}
\usepackage{epigraph}
\usepackage{adjustbox}

\usepackage{titling}

\RequirePackage[format=plain,labelformat=simple,labelsep=period,font=small,compatibility=false]{caption}
\RequirePackage[font=footnotesize,skip=3pt,subrefformat=parens]{subcaption}

\newcommand{\tableFont}{\fontsize{7pt}{8.4pt} \selectfont}

\def\wacvPaperID{404} 

\wacvalgorithmstrack   

\wacvfinalcopy 

\ifwacvfinal
\usepackage[breaklinks=true,bookmarks=false]{hyperref}
\else
\usepackage[pagebackref=true,breaklinks=true,colorlinks,bookmarks=false]{hyperref}
\fi

\begin{document}

\title{Spatially Multi-conditional Image Generation}

\author{Ritika~Chakraborty$^{*1}$,  Nikola~Popovic$^{*1}$, Danda~Pani~Paudel$^{1}$, Thomas~Probst$^{1}$, Luc~Van~Gool$^{1,2}$\\
$^{1}$Computer Vision Laboratory, ETH Zurich, Switzerland\\
$^{2}$VISICS, ESAT/PSI, KU Leuven, Belgium\\
{\tt\small \{critika, nipopovic, paudel, probstt, vangool\}@vision.ee.ethz.ch}
}
\date{}



\maketitle

\def\thefootnote{*}\footnotetext{Equal contributions.}\def\thefootnote{\arabic{footnote}}

\thispagestyle{empty}

\input{latex_submission/paper/00_abstract}
\input{latex_submission/paper/01_introduction}

\input{latex_submission/paper/02_related_work}
\input{latex_submission/paper/03_method}

\input{latex_submission/paper/04_experiments}

\input{latex_submission/paper/0x_conclusion}


\appendix

\title{
Spatially Multi-conditional Image Generation \\ Supplementary material
}

\maketitle

\input{latex_submission/supplementary/all-tlam}

\clearpage

{\small
\bibliographystyle{ieee_fullname}
\bibliography{ms}
}

\end{document}

%% file: latex_submission/paper/00_abstract.tex
\begin{abstract}
In most scenarios, conditional image generation can be thought of as an inversion of the image understanding process. Since generic image understanding involves solving multiple tasks, it is natural to aim at generating images via multi-conditioning. However, multi-conditional image generation is a very challenging problem due to the heterogeneity and the sparsity of the (in practice) available conditioning labels. In this work, we propose a novel neural architecture to address the problem of heterogeneity and sparsity of the spatially multi-conditional labels. Our choice of spatial conditioning, such as by semantics and depth, is driven by the promise it holds for better control of the image generation process. The proposed method uses a transformer-like architecture operating pixel-wise, which receives the available labels as input tokens to merge them in a learned homogeneous space of labels. The merged labels are then used for image generation via conditional generative adversarial training. In this process, the sparsity of the labels is handled by simply dropping the input tokens corresponding to the missing labels at the desired locations, thanks to the proposed pixel-wise operating architecture. Our experiments on three benchmark datasets demonstrate the clear superiority of our method over the state-of-the-art and compared baselines. 
The source code will be made publicly available.            
\end{abstract}

%% file: latex_submission/paper/01_introduction.tex

\section{Introduction}
\label{sec:introduction}

In recent years, automated image generation under user control has become more and more of a reality. 
Such processes are typically based on so-called conditional image generation methods~\cite{mirza2014conditional,pix2pix2017}. 
One could see these as an intermediate between fully unconditional~\cite{goodfellow2014generative} and purely rendering based~\cite{cook1987reyes} generation, respectively. Methods belonging to these two extreme cases either offer no user control (the former) or need to get all necessary information of image formation supplied by the user (the latter).  In many cases, neither extreme is desirable. 
Therefore, several conditional image generation methods, that circumvent the rendering process altogether, have been proposed. These methods usually receive the image descriptions either in the form of text~\cite{mirza2014conditional,reddydall} or as spatially localized semantic classes~\cite{pix2pix2017,park2019semantic,RottShaham2020ASAP}. 

In this paper, we aim to condition the image generation process beyond text descriptions and the desired semantics. In this regard, we make a generic practical assumption that any semantic or geometric aspects of the desired image may be available for conditioning.  For example, an augmented reality application may require to render a known object using its 3D model and specified pose, but with missing texture and lighting details. In such cases, attributes such as semantic mask, depth, normal, curvature at that object's location, become instantly available for the image generation process, offering us the multi-conditioning inputs. The geometric and semantic aspects of the other parts of the same image may also be available partially or completely. Incorporating all such information in a multi-conditional manner to generate the desired image, is the main challenge that we undertake. In some sense, our approach bridges the gap between generative and rendering based image synthesis.

Two major challenges of  multi-conditional image generation, in the context of this paper, are the heterogeneity and the sparsity of the available labels. The heterogeneity refers to differences in representations of different labels, for e.g. depth and semantics. On the other hand, the sparsity is either simply caused by the label definition (e.g. sky has no normal) or due to missing annotations~\cite{popovic2021compositetasking}. 
It is important to note that some geometric aspects of images, such as an object's depth and orientation, can be introduced manually (e.g. by sketching), without requiring any 3D model with its pose. This allows users to geometrically control images (beyond the semantics based control) both in the the presence or absence of 3D models. It goes without saying that the geometric manipulation of any image can be carried out by first inferring its geometric attributes using existing methods~\cite{popovic2021compositetasking,sun2021task,yu2019inverserendernet}, followed by generation after manipulation.

To address the problems of both heterogeneity and diversity, we propose a label merging network that learns to merge the provided conditioning labels pixel-wise. To this end, we introduce a novel transformer-based architecture, that is designed to operate on each pixel individually. The provided labels are first processed by label-specific multilayer perceptrons (MLPs) to generate a token for each label. The tokens are then processed by the transformer module. In contrast to popular vision transformers that perform spatial attention~\cite{dosovitskiy2020vit,Yuan2021TokenstoTokenVT}, our transformer module applies self-attention across the label dimension, thus avoiding the high computational complexity. The pixel-wise interaction of available labels homogenizes the different labels to a common representation for the output tokens. The output tokes are then averaged to obtain the fused labels in the homogeneous space to form the local \emph{concept}. This is performed efficiently for all pixels in parallel by sliding the pixel-wise transformer over the input label maps. Finally, the concepts are used for the image generation via conditional generative adversarial training, using a state-of-the-art method~\cite{RottShaham2020ASAP}. During the process of label merging, the spatial alignment is always preserved. The sparsity of the labels is handled by simply dropping the input tokens of the missing labels, at the corresponding pixel locations. This way, the transformer learns to reconstruct the concept for each pixel, also in the case when not all labels are available. 

We study the influence of several spatially conditioning labels including semantics, depth, normal, curvature, edges, in three different benchmark datasets. The influence of the labels is studied both in the case of sparse and dense label availability.  In both cases, the proposed method provides outstanding results by clearly demonstrating the benefit of conditioning labels beyond the commonly used image semantics. The major contribution of this paper can be summarized as follow:

\begin{enumerate}
    \item We study the problem of spatially multi-conditional image generation, for the first time.
    \item We propose a novel neural network architecture to fuse the heterogeneous multi-conditioning labels provided for the task at hand, while also handling the sparsity of the labels at the same time.   
    \item We analyse the utility of various conditioning types for image generation and present outstanding results obtained by the proposed method in benchmark datasets.
\end{enumerate}

\input{figures/paper/teaser}



%% file: figures/paper/teaser.tex
\begin{figure}[t!]
    \centering
    \captionsetup{font=small}
    \includegraphics[width=0.9\linewidth]{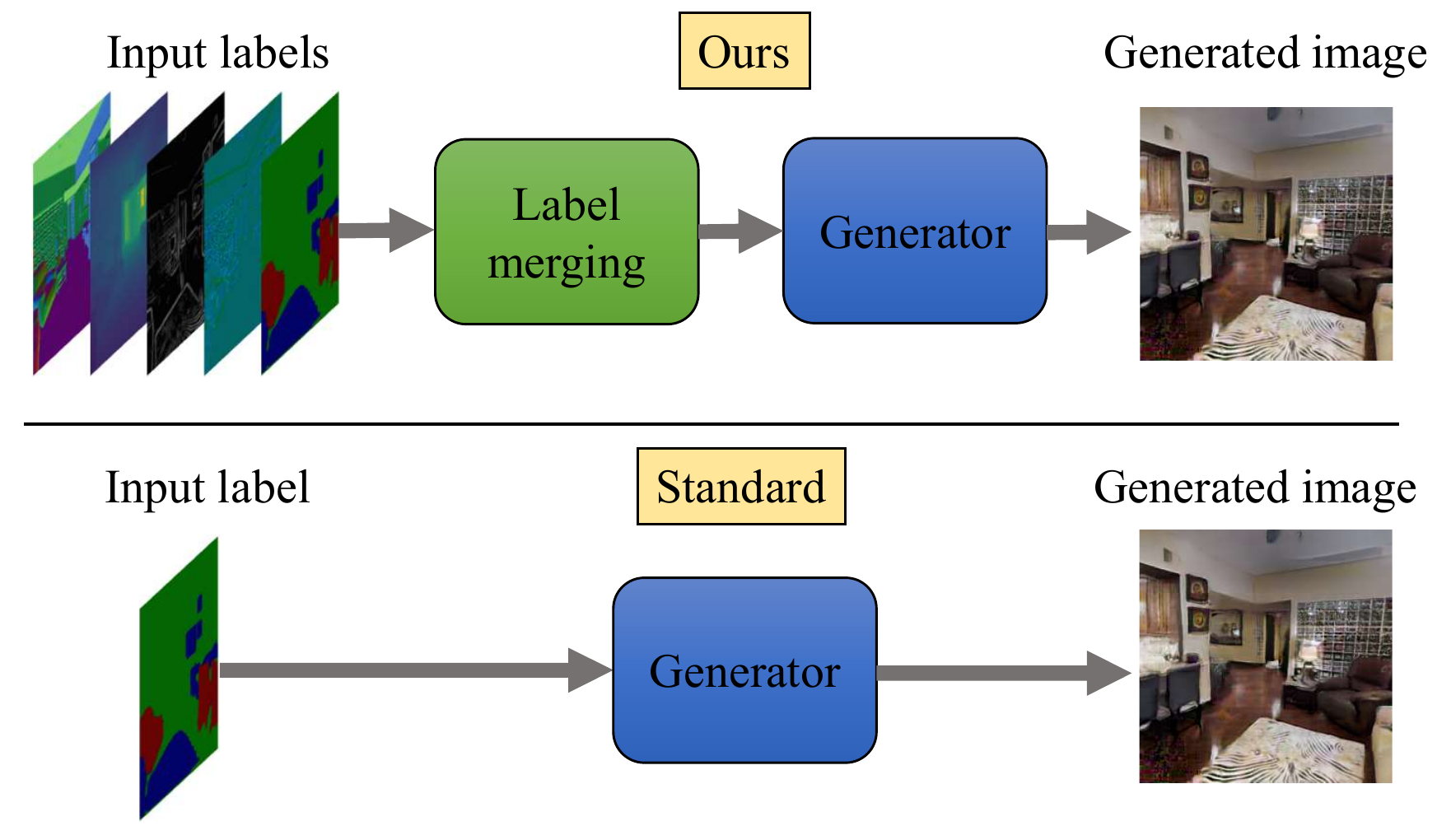}
    \vspace{-6pt}
    \caption{{\textbf{Spatially multi-conditional image generation}. Our model uses multiple labels to generate an image, compared to standard approaches which use only the semantic segmentation label. Multiple input labels, coming from different sources, are handled by the proposed label merging block.}}
    \vspace{-6pt}
    \label{fig:main_teaser}
\end{figure}

%% file: latex_submission/paper/02_related_work.tex
\section{Related Work}
\label{sec:related}

\noindent\textbf{Conditional Image Synthesis.}
A description of the desired image to be generated can be provided in various forms, from 
class conditions~\cite{Mirza2014ConditionalGA,brock2018large}, text~\cite{mirza2014conditional,reddydall}, spatially localized semantic classes~\cite{pix2pix2017,park2019semantic,RottShaham2020ASAP}, sketches~\cite{pix2pix2017}, style information~\cite{Gatys2016ImageST,johnson2016style,karras2019stylebased} to human poses~\cite{Ma2017PoseGP}. 
Recently, handling of different data structures (e.g. text sequences and images) has received attention in the literature~\cite{zhang2021m6,Xia2021TediGANTD}. 
The problem of spatially multi-conditional image generation is orthogonal to the problem of unifying different (non-spatial) modalities, as it seeks to fuse heterogeneous spatially localized labels into concepts, while preserving the spatial layout for image generation.
%
%
In the area of image-to-image translation, Ronneberger et al. introduced the \emph{UNet}~\cite{ronneberger2015unet} architecture, which has been used in several important follow up works. Isola et al.~\cite{pix2pix2017} later introduced the \emph{Pix2Pix} paradigm to convert sketches into photo-realistic images, leveraging the UNet backbone as a generator combined with a convolutional discriminator. This work was improved  by Wang et al. to support high resolution image translation in~\emph{Pix2PixHD} \cite{wang2018pix2pixHD} and video translation in \emph{Vid2Vid}~\cite{wang2018vid2vid}.
Recently, Shaham et al. introduced the \emph{ASAP-Net}~\cite{RottShaham2020ASAP} which achieves a superior trade-off of inference time and performance on several image translation tasks. We employ the \emph{ASAP-Net} as one component in our architecture.

\input{figures/paper/network_diagram}

\noindent\textbf{Conditioning Mechanisms.}
The conditioning mechanism is at the core of semantically controllable neural networks, and is often realized in conjunction with normalization techniques \cite{dumoulin2018feature-wise,huang2020normalization}.  Perez et al. introduced a simple  feature-wise linear modulation FiLM~\cite{perez2017film} for  visual reasoning. In the context of neural style transfer~\cite{gatys2015neural}, Huang et al. introduced Adaptive Instance Normalization AdaIN~\cite{Huang_2017_ICCV}.
Park et al. extended AdaIN for spatial control in SPADE~\cite{park2019semantic}, where the normalization parameters are derived from a semantic segmentation.   Zhu et al.~\cite{Zhu_2020_CVPR} further extend SPADE to allow for independent application of global and local styles. Finally, generic normalisation schemes that make use of kernel prediction networks to achieve arbitrarily global and local control include Dynamic Instance Normalization (DIN)~\cite{jing2019dynamic} and Adaptive Convolutions (AdaConv)~\cite{Chandran_2021_CVPR}. While being greatly flexible, they also increase the resulting inference time, unlike ASAP-Net that uses adaptive implicit functions~\cite{sitzmann2019scene} for efficiency. 

\noindent\textbf{Differentiable Rendering Methods.}
Since rendering is a complex and computationally expensive process, several approximations were proposed to facilitate its use for training neural networks. Methods starting from simple approximations of the rasterization function to produce silhouettes~\cite{kato2018renderer,liu2019soft,Laine2020diffrast} to more complex approximations modelling indirect lighting effects~\cite{Li2018,NimierDavid2020Radiative,Loubet2019Reparameterizing} have been proposed. 
These algorithms have also been incorporated into popular deep learning frameworks~\cite{ravi2020pytorch3d,Li2018,NimierDavidVicini2019Mitsuba2}. Differentiable renderers have been successfully used in several neural networks including those used for face~\cite{tewari17MoFA,tewari2018} and human body~\cite{lin2021endtoend,pavlakos2019expressive} reconstruction. We refer the interested reader to the excellent surveys of Tewari et al.~\cite{tewari2020state} and Kato et al.~\cite{kato2020differentiable} for more details.
{In contrast to rendering approaches which require setting numerous scene parameters, our method directly generates realistic images from only a sparse set of chosen labels.}

%% file: figures/paper/network_diagram.tex
\begin{figure*}[t!]
    \centering
    \captionsetup{font=small}
    \def\svgwidth{0.77\textwidth}
    \adjustbox{trim=0.0cm 1.5cm 0cm 1.0cm, clip}{%
    \includesvg{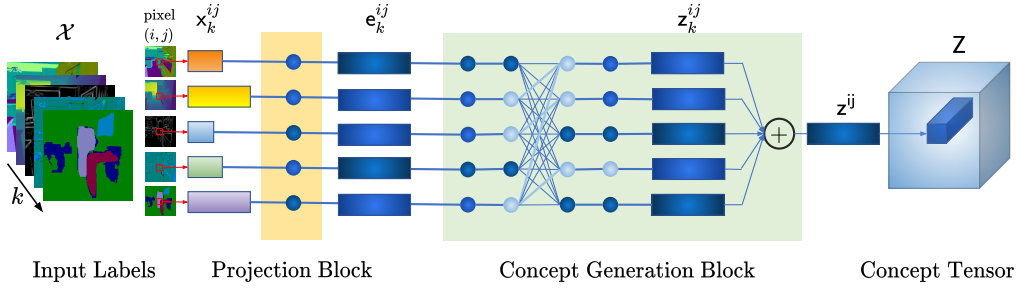}}
    \vspace{-10mm}
    \caption{{\textbf{Pixel-wise Transformer Label Merging block (TLAM)}. 
    The heterogeneous labels of each pixel are first projected into the same dimensionality, and then passed to the concept generation block.
    A transformer module promotes the interaction between labels, before finally distilling them to a concept representation by averaging the homogeneous label embeddings at every pixel location.
    }}
    \label{fig:LAME_block}
\end{figure*}

%% file: latex_submission/paper/03_method.tex
\section{Method}


We start by introducing a few formal notations. As an input, we have a collection of labels $\mathcal{X}=\{\mathsf{X}_1, \mathsf{X}_2, ..., \mathsf{X}_N\}$, where each label $\mathsf{X}_k \in \mathbb{R}^{H \times W \times C_k}$ has height $H$, width $W$ and $C_k$  channels. The element of the label $\mathsf{X}_k$ corresponding  to the pixel location  $(i,j)$  is denoted as $\mathsf{x}_{k}^{ij} \in \mathbb{R}^{C_k}$. Hence, elements of the label set $\mathcal{X}$ corresponding to the pixel location $(i,j)$ form a set $\mathcal{X}^{ij}=\{\mathsf{x}_{1}^{ij}, \mathsf{x}_{2}^{ij}, ..., \mathsf{x}_{N}^{ij}\}$. The model takes the set of labels $\mathcal{X}$ as an input and produces $\mathsf{I} = \phi (\mathcal{X})$, where $\mathsf{I} \in \mathbb{R}^{H \times W \times 3}$ is the generated image.

\subsection{Label Merging}
We describe the mechanism of the label merging component, as illustrated in Figure~\ref{fig:LAME_block}. This component processes different spatial locations of the input labels $\mathcal{X}^{ij}$ independently, and is efficiently performed on all pixels in parallel. We empirically found this to be sufficient. To this end, we first collapse the spatial dimensions of every label $\mathsf{X}_k$ to obtain $\mathsf{E}_k \in \mathbb{R}^{H W \times d}$, which contains $HW$  embedding vectors $\mathsf{e}_{k}^{ij} \in \mathbb{R}^{d}$.
In this process, we are interested to merge all embedding vectors $\{\mathsf{e}_{k}^{ij}\}$ into a common concept vector $\mathsf{z}^{ij}$, for each pixel location. 
The label merging is performed by two blocks: the projection block and the concept generation block, which we described in the following.

\subsubsection{Projection Block}
\label{sec:projection_block}
This block projects every heterogeneous input label $\mathsf{X}_k$ into an embedding space $\mathsf{E}_k \in \mathbb{R}^{H \times W \times d}$, where $d$ is the dimensionality of the embedding space.
It does so by transforming every token $\mathsf{x}_k^{ij}$ with the projection function $f_k$, to project it into the embedding space $\mathsf{e}_{k}^{ij} = f_k (\mathsf{x}^{ij})$, where $\mathsf{e}_{k}^{ij} \in \mathbb{R}^{d} $. All the elements of a given label $\mathsf{X}_k$ share the same projection function $f_k$.  Different input labels have different projection functions $f_k$. We use an affine transformation followed by the GeLU nonlinearity $a(\cdot)$~\cite{hendrycks2020GELU} to serve as the embedding function,
\begin{equation}
    \label{eq:projection_block}
    \mathsf{e}_{k}^{ij} = f_k (\mathsf{x}_{k}^{ij}) = a(\mathsf{A}_{k}\mathsf{x}_{k}^{ij} + \mathsf{b}_k),
\end{equation}
where $\mathsf{A}_k \in \mathbb{R}^{d \times C_k }$ and $\mathsf{b}_k \in \mathbb{R}^{d}$. We implement this by using fully connected layers as projection functions for each input label $\mathsf{X}_k$, followed by the GeLU activation function.  Our method also works with sparse input labels, which can be caused by label definition or missing labels. When the value of a label $\mathsf{X}_k$ is missing at a certain spatial location, its element token $\mathsf{x}_{k}^{ij}$ is  dropped by setting it to a zero vector. Thus, a token with the embedding $\mathsf{e}_k^{ij} = a(\mathsf{b}_k)$ will send a signal to the concept generation block that the label $k$ is not present at this location, and that it should extract the information from other labels. Also, the absence of the whole label $\mathsf{X}_k$ is handled similarly.

\begin{figure}[t!]
    \centering
    \includegraphics[width=8cm]{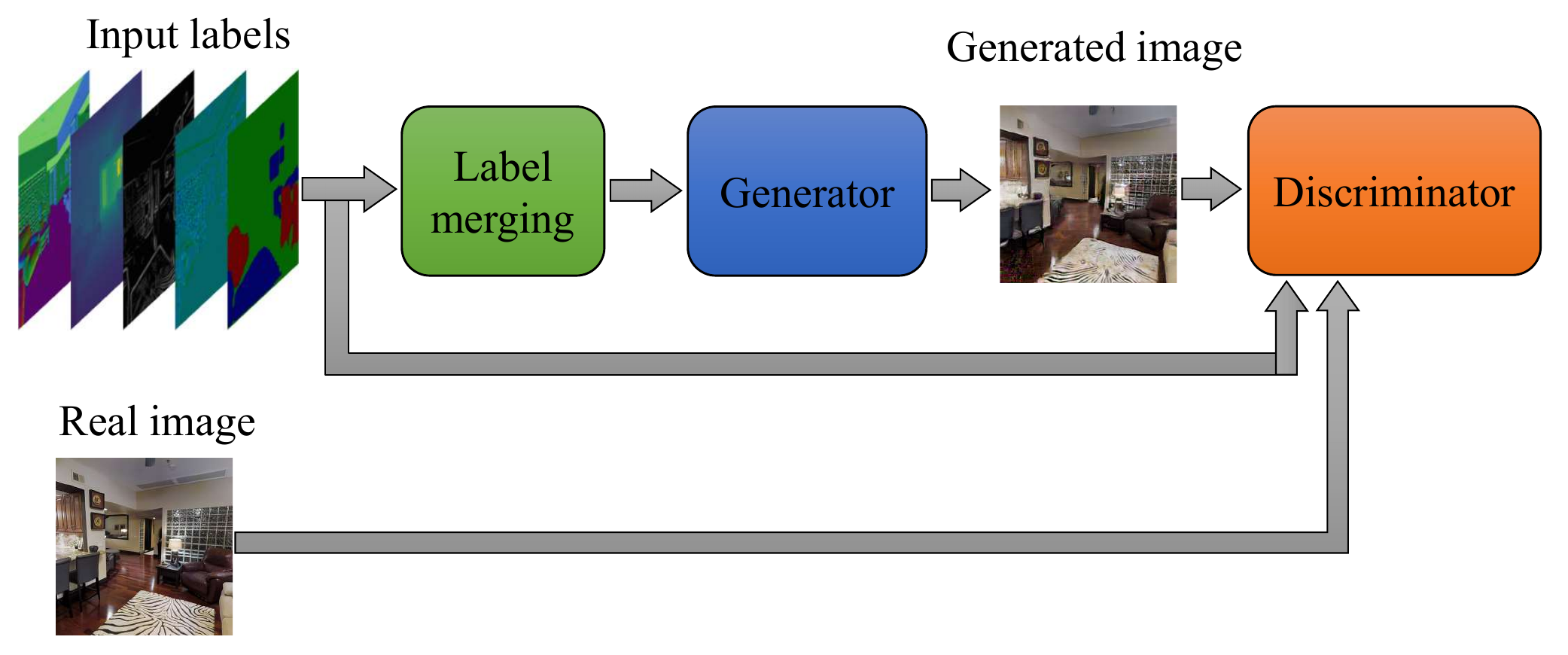}
    \vspace{-4pt}
    \caption{\textbf{Network overview.} Different input labels are embedded into a homogeneous space with the label merging module. The generator uses this embedding to generate an image. During training, the discriminator takes the labels, real and generated images, and uses them to optimize the label merger and the generator.}
    \label{fig:netOverview}
    \vspace{-6pt}
\end{figure}

\subsubsection{Concept Generation Block}
\label{ssec:concept_generation}
The collection of the embedding vectors corresponding to different labels $\mathcal{E} = \{ \mathsf{e}_{1}, .., \mathsf{e}_{k}, ..., \mathsf{e}_{N}\}$ serve as input tokens for our concept generation block. 
This block uses a novel attention mechanism to model the interaction across labels, which is shared across all spatial locations. It therefore does not require expensive spatial attention as used in vision transformer architectures~\cite{Vaswani2017,dosovitskiy2020vit}. In other words,  we apply the same label-transformer on each pixel individually. Transformers are naturally suited here to encourage interaction between the embedded label tokens to share their label specific information while merging it into the final label representation. Before giving $\mathcal{E}$ to the transformer, we add label specific encodings $\mathsf{p_k}$ to the embedded tokens $\mathsf{e}_k$ to obtain $\mathsf{z}_k^{(0)} = \mathsf{e}_{k} + \mathsf{p}_k$. Then we pass $\mathcal{Z}^{(0)} = \{ \mathsf{z}_{1}^{(0)}, \mathsf{z}_{2}^{(0)}, ... ,\mathsf{z}_{N}^{(0)} \}$ through $l$ transformer blocks $\mathsf{B}_m$. Each block $\mathsf{B}_m$ takes the output $\mathcal{Z}^{(m-1)}$ of the previous block and produces $\mathcal{Z}^{(m)}= \mathsf{B}_m(\mathcal{Z}^{(m-1)})$, where $\mathcal{Z}^{(m)} = \{ \mathsf{z}_{1}^{(m)}, \mathsf{z}_{2}^{(m)}, \ldots, \mathsf{z}_{N}^{(m)} \}$. 

Each transformer block $\mathsf{B}_m$ is composed of a Multi-head Self Attention block (MSA), followed by a Multilayer Perceptron block (MLP)~\cite{Vaswani2017,dosovitskiy2020vit}.
Self-attention is a global operation, where every token interacts with every other token and thus information is shared across them.
Multiple heads in the MSA block are used for more efficient computation and for extracting richer and more diverse features.
The MSA block executes the following operations:
\begin{equation}
    \label{eq:MSA}
    \mathcal{\hat{Z}}^{(m)} = 
    \mathsf{MSA}(\mathsf{LN}(\mathcal{Z}^{(m-1)})) + 
    \mathcal{Z}^{(m-1)},
\end{equation}
where $\mathsf{LN}$ represents Layer Normalization ~\cite{ba2016LayerNorm}.
The MSA block is followed by the MLP block, which processes each token separately using the same Multilayer Perceptron. 
This block further processes the token features, after their global interactions in the MSA block, by sharing and refining the representations of each token across all their channels. 
The MLP block executes the following operations:
\begin{equation}
    \label{eq:MLP}
    \mathcal{Z}^{(m)} = 
    \mathsf{MLP}(\mathsf{LN}(\mathcal{\hat{Z}}^{(m)})) + 
    \mathcal{\hat{Z}}^{(m)}.
\end{equation}

Note that, for vision transformers~\cite{dosovitskiy2020vit} operating on $M$ spatial elements (e.g. pixels/patches), the computational complexity of the self-attention is $\mathcal{O}(M^2)$. Our pixel-wise transformer, operating only on $N$ label tokens, reduces the complexity of self-attention to $\mathcal{O}(N^2)$ (per pixel) with the number of labels $N \ll M.$ 

Finally, all the elements of the output set produced by the final transformer block $\mathcal{{Z}} = \{ \mathsf{z}_{1}^{(l)}, \mathsf{z}_{2}^{(l)}, ... , \mathsf{z}_{N}^{(l)} \}$ are averaged to obtain $\mathsf{z} = \frac{1}{N} \sum_{k=1:N}\mathsf{z}_{k}$. They are reshaped back into the original spatial dimensions to obtain the \emph{Concept Tensor} $\mathsf{Z} \in \mathbb{R}^{H \times W \times d}$.  We name this label merging block as the Transformer LAbel Merging (\textbf{TLAM}) block.

\begin{figure*}[t]
\addtolength{\tabcolsep}{-4.5pt}
\centering
\footnotesize
\begin{tabular} {cc|cc|cc}
 All-dense & All-sparse & SPADE~\cite{park2019semantic} & ASAP~\cite{RottShaham2020ASAP} & \makecell{TLAM \\(Ours)} & \makecell{Sparse-TLAM \\(Ours)} \\
 \includegraphics[width=0.10\linewidth]{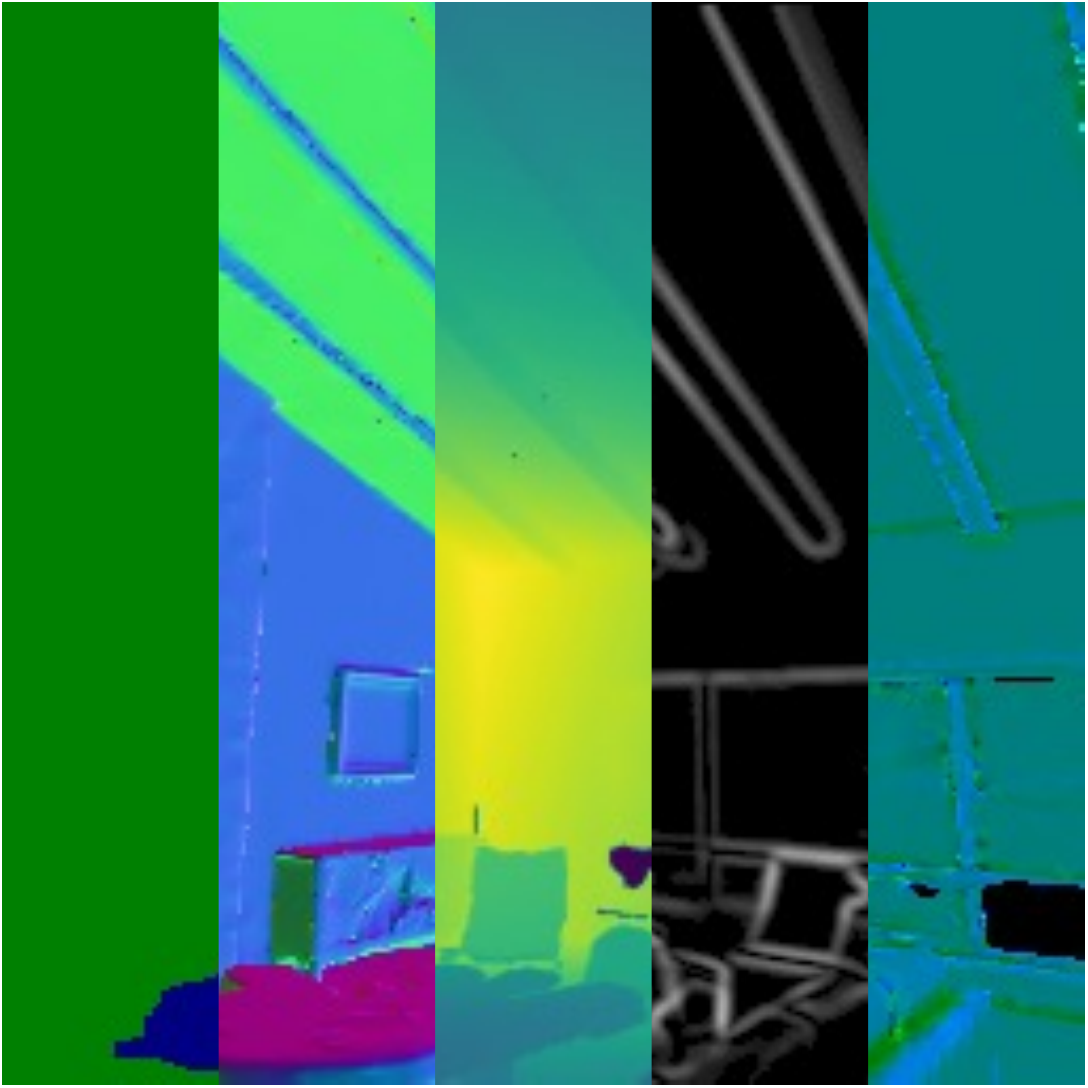}
 & \includegraphics[width=0.10\linewidth]{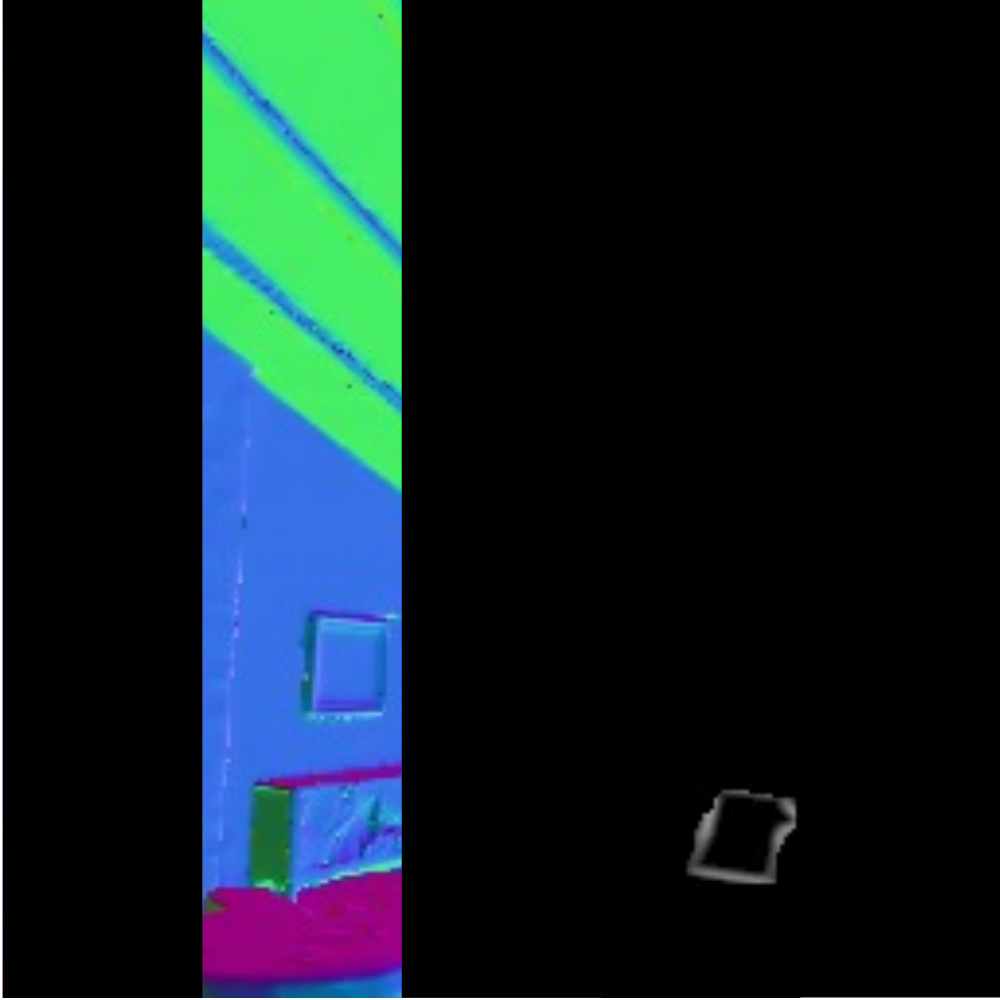} &
\includegraphics[width=0.10\linewidth]{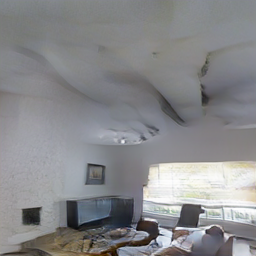} &   \includegraphics[width=0.10\linewidth]{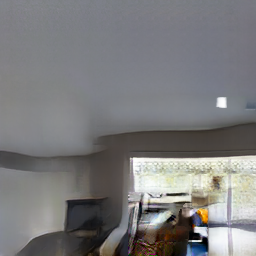} &
\includegraphics[width=0.10\linewidth]{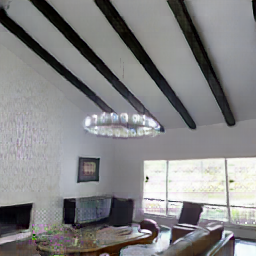} &  \includegraphics[width=0.10\linewidth]{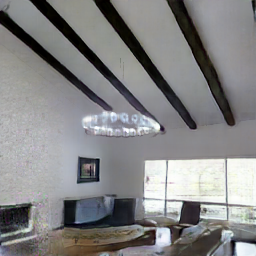} \\

\includegraphics[width=0.10\linewidth]{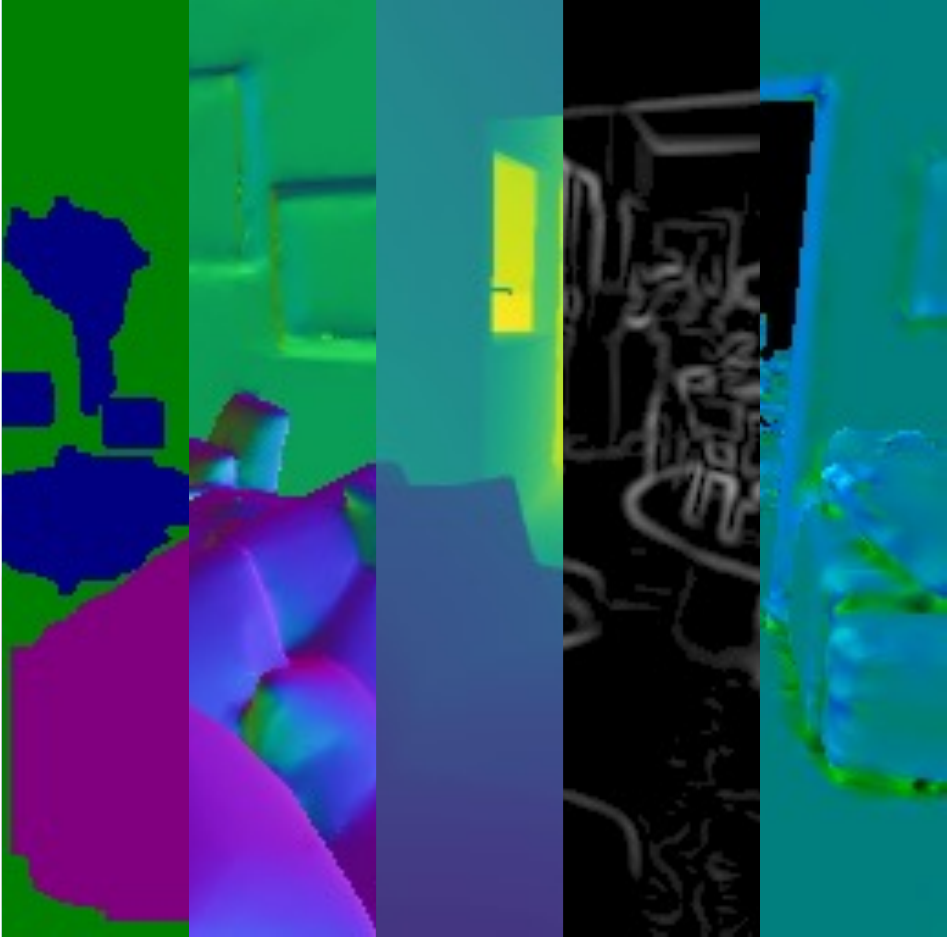} &
 \includegraphics[width=0.10\linewidth]{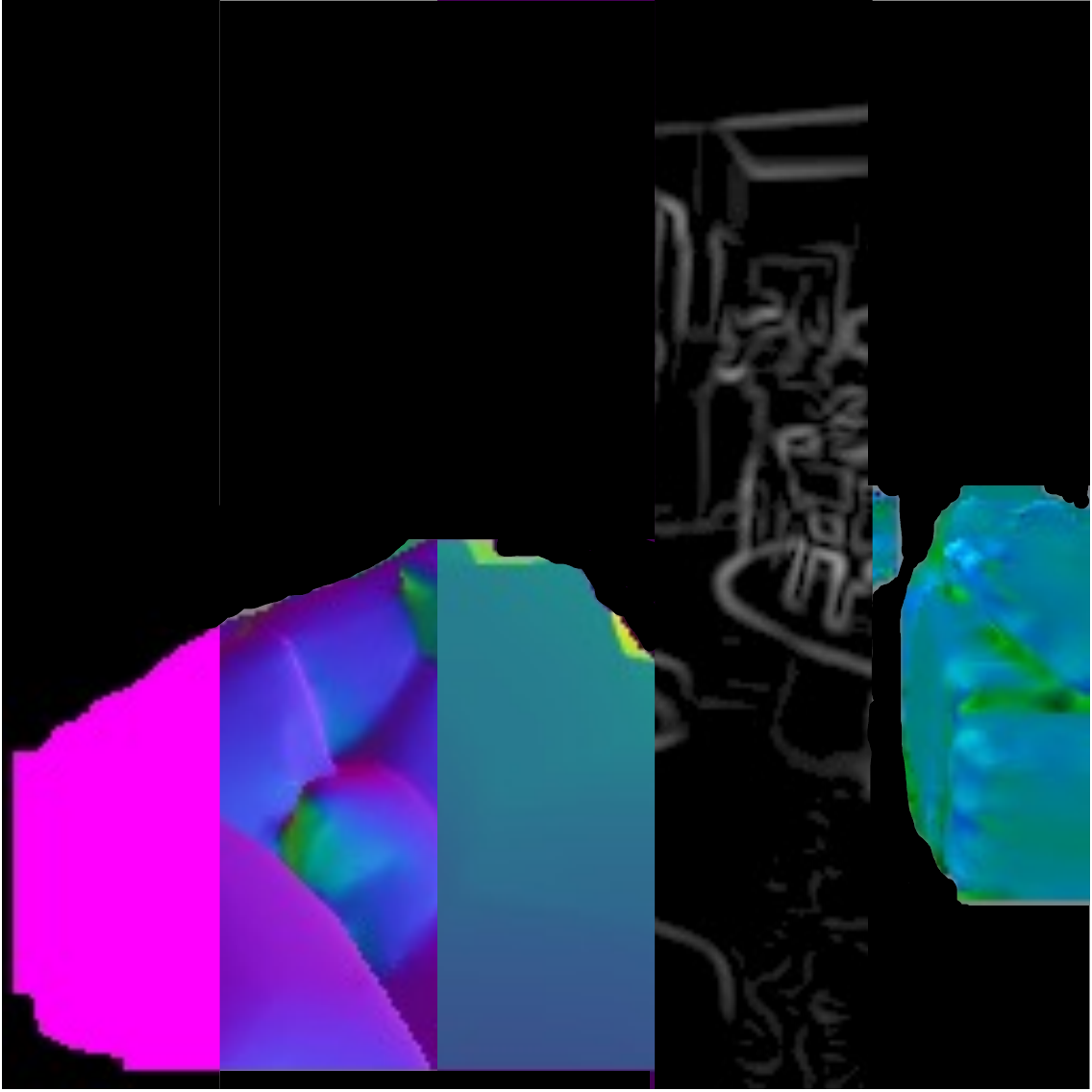} &
\includegraphics[width=0.10\linewidth]{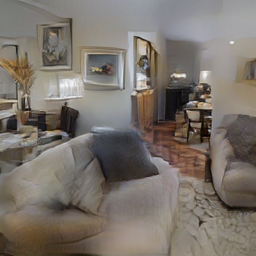} &   \includegraphics[width=0.10\linewidth]{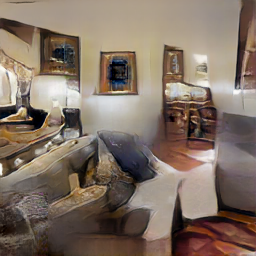} &
\includegraphics[width=0.10\linewidth]{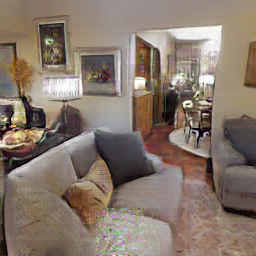} &  \includegraphics[width=0.10\linewidth]{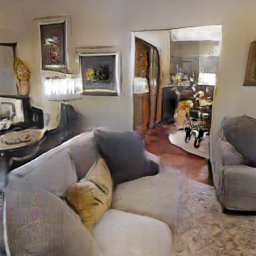} \\

\includegraphics[width=0.10\linewidth]{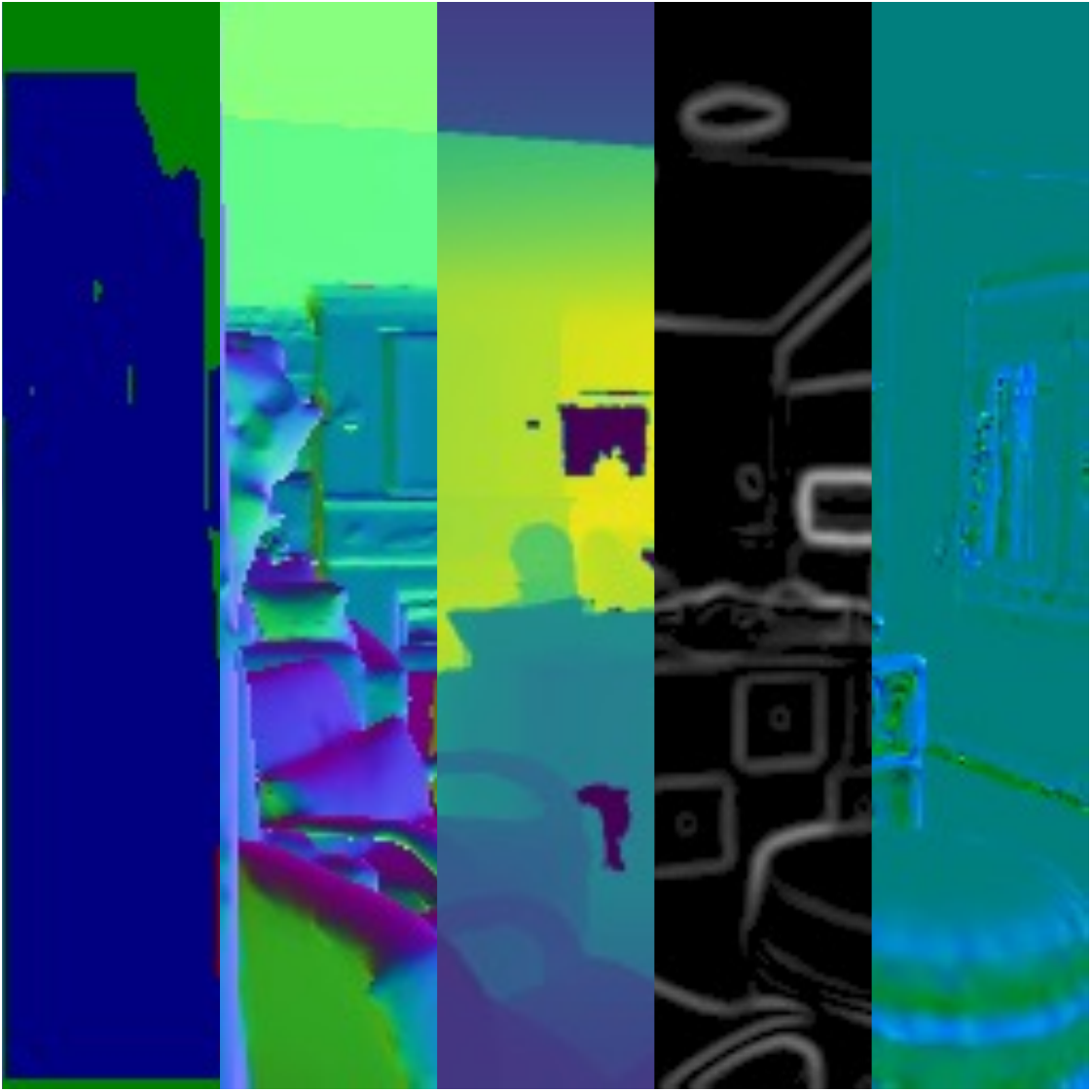}& 
 \includegraphics[width=0.10\linewidth]{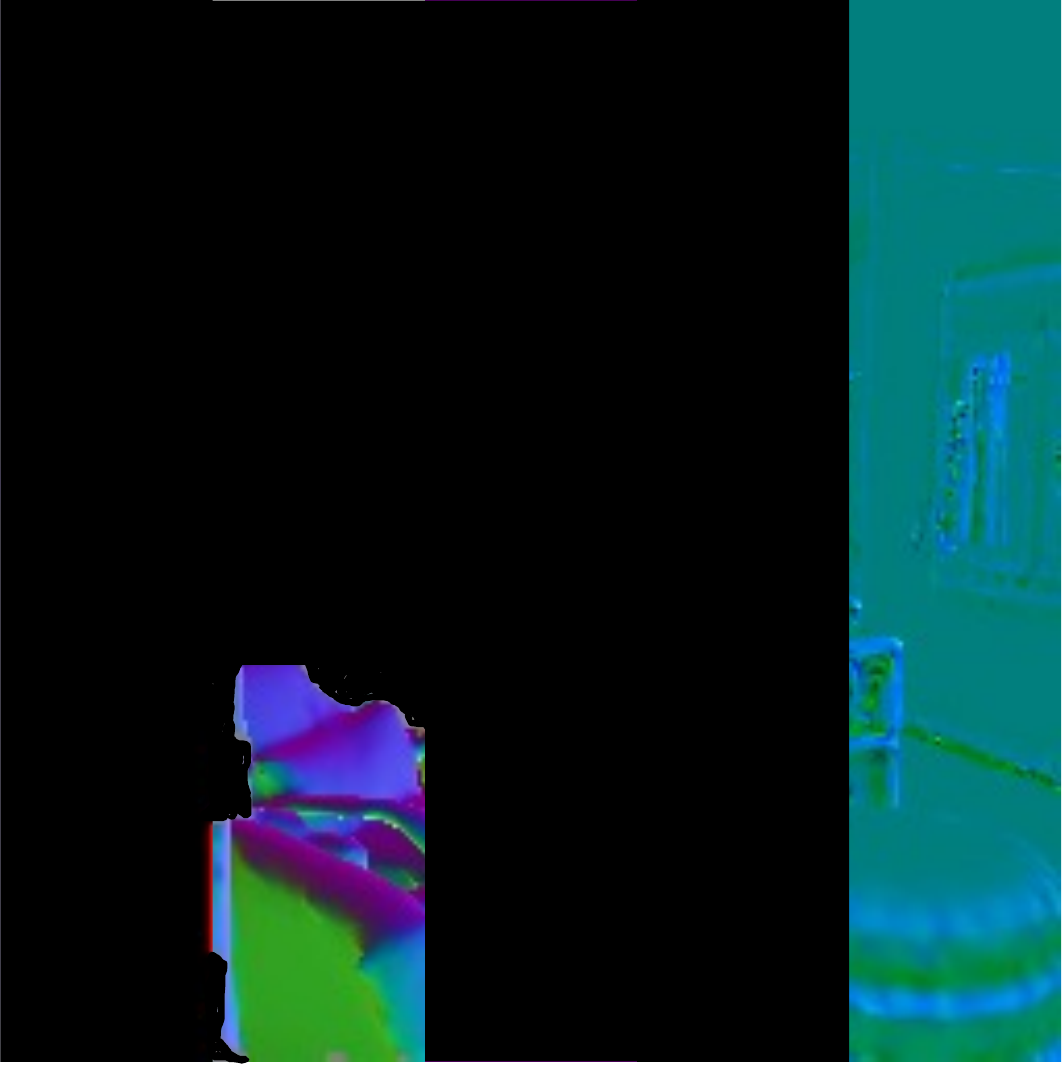} &
\includegraphics[width=0.10\linewidth]{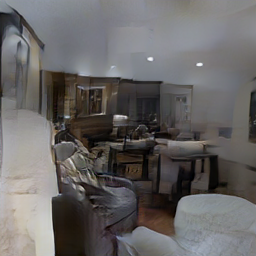} &   \includegraphics[width=0.10\linewidth]{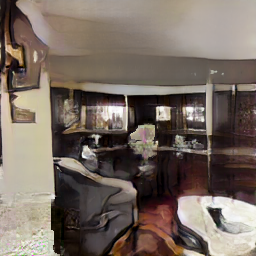} &
\includegraphics[width=0.10\linewidth]{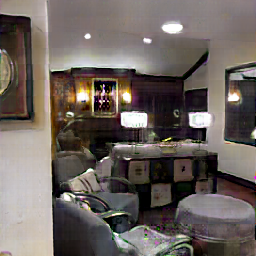} &  \includegraphics[width=0.10\linewidth]{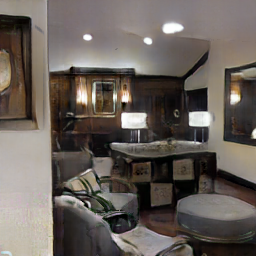} \\

\end{tabular}
\vspace{-2mm}
\caption{\textbf{Visual comparisons on the Taskonomy dataset.} Column 1 \& 2: five different labels tiled horizontally for dense and 50\% sparse cases. Column 5 \& 6: images generated by our  method using the labels from column 1 \& 2. Column 3 \& 4: images generated by the dense semantic methods. Our method generates more realistic images with fine geometric and visual details from both dense and sparse labels.}
\vspace{-6pt}
\label{fig::taskonomy}
\end{figure*}

\subsection{Network Overview}
\label{sec:network_overwiev}
The proposed model is divided into two components: the label merging component and the image generation component. This is depicted in Figure~\ref{fig:netOverview}. 

\noindent\textbf{Label Merging.} The label merging block takes the set of heterogeneous labels $\mathcal{X}$ as the input and merges them into a homogeneous space $\mathsf{Z} = \psi(\mathcal{X})$, where $\mathsf{Z} \in \mathbb{R}^{H \times W \times d}$ is the \emph{Concept Tensor}. It is important to note that the label merging block does not require the number of input labels to be the same for every input. Input labels $\mathcal{X}$ are heterogeneous because they come from different sources and can have different numbers of channels $C_k$, as well as different ranges of values. For example, semantic segmentation labels are represented using discrete values, while surface normals are continuous. The label merging block first translates all the provided labels $\mathsf{X}_k$ into the same dimensional embeddings $\mathsf{E}_k$, using the projection block. Then, it uses the concept generation block to translate embeddings $\mathsf{E}_k$ to the concept tensor $\mathsf{Z}$. In short, the label merging block fuses the heterogeneous set of input labels $\mathcal{X}$ into a homogeneous Concept Tensor $\mathsf{Z}$. This block is depicted in Figure~\ref{fig:LAME_block}.


\noindent\textbf{Generator.}
The task of the generator is to take the produced \emph{Concept Tensor} $\mathsf{Z}$ and generate the image $\mathsf{I} = g(\mathsf{Z})$, as shown in Figure~\ref{fig:netOverview}. As the generator, we employ the state-of-the-art ASAP model~\cite{RottShaham2020ASAP}. This model first synthesizes the high-resolution pixels using lightweight and highly parallelizable operators. ASAP performs most computationally expensive image analysis at a very coarse resolution. We modify the input to ASAP, by providing the \emph{Concept Tensor} $\mathsf{Z}$ as an input, instead of giving only one specific input labels $\mathsf{X}_k$. The generator can therefore exploit the merged information from all available input labels $\mathcal{X}$.

\noindent\textbf{Adversarial Training for Multi-conditioning.}
We follow the optimization protocol of ASAP-Net~\cite{RottShaham2020ASAP}.  We train our generator model adversarially with a multi-scale patch discriminator, as suggested by pix2pixHD~\cite{wang2018pix2pixHD}.  To achieve this, we modify the input to the discriminator by using the \emph{Concept Tensor} $\mathsf{Z}$ instead of a specific label $\mathsf{X}_k$, for example semantics used by the most existing works.

%% file: latex_submission/paper/04_experiments.tex
\section{Experiments}
 
\newcommand{\HalfSparsePic}{ \includegraphics[width=2.6mm]{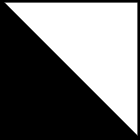}}
\newcommand{\DensePic}{ \includegraphics[width=2.6mm]{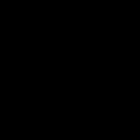}}
\newcommand{\NoTaskPic}{}
 \begin{table}[t!]
	\centering
    \tableFont
	\resizebox{\columnwidth}{!}{
    \begin{tabular}{c|c|c|c|c|c|c|c}
        \multirow{2}{*}{Backbone} & \multirow{2}{*}{Method}& \multicolumn{5}{|c|}{Label sparsity} & \multirow{2}{*}{FID} \\ 
		\cline{3-7}
		  &  & S & E & C & D & N & \\
		\hline \hline
		\multirow{2}{*}{SPADE~\cite{park2019semantic}} & Regular & 
		\DensePic & \NoTaskPic& \NoTaskPic & \NoTaskPic & \NoTaskPic & 72.3 \\ \cline{2-8}
		 & Naive-baseline & 
		\DensePic & \DensePic& \DensePic & \DensePic & \DensePic & 66.1 \\
		\hline
		\hline
		
    	 \multirow{6}{*}{ASAP~\cite{RottShaham2020ASAP}} & Regular & \DensePic & \NoTaskPic& \NoTaskPic & \NoTaskPic & \NoTaskPic & 73.8 \\ \cline{2-8}
    	 &  \makecell{Naive-baseline} & \DensePic & \DensePic& \DensePic & \DensePic & \DensePic & 74.6 \\
    	\cline{2-8} 
    	 & CLAM-baseline (Ours) & \HalfSparsePic & \HalfSparsePic& \HalfSparsePic & \HalfSparsePic & \HalfSparsePic& 43.8 \\
    	\cline{2-8}
         & TLAM (Ours) & \HalfSparsePic & \HalfSparsePic& \HalfSparsePic & \HalfSparsePic & \HalfSparsePic & 37.9 \\
    	\cline{2-8}
         & CLAM-baseline (Ours) & \DensePic & \DensePic& \DensePic & \DensePic & \DensePic & \underline{37.1}  \\
    	\cline{2-8}
         & TLAM (Ours) & \DensePic & \DensePic& \DensePic & \DensePic & \DensePic & \textbf{30.6} \\
    	\cline{2-8}
	\hline
	\end{tabular}
	}
	\caption{\textbf{FID scores on the Taskonomy dataset}. Our TLAM method generates images with significantly better visual quality for both sparse and dense labels. Symbol  {\DensePic} corresponds to dense labels, while {\HalfSparsePic}  corresponds to $50\%$ label sparsity. S stands for semantics, E for edges, C for curvature, D for depth and N for normals. Please, refer Figure~\ref{fig::taskonomy} for corresponding images. }\label{taskonomy-tbl}
	\vspace{-12pt}
 \end{table}
 
\noindent\textbf{Implementation Details.} 
We follow the optimization protocol of ASAP-Net~\cite{RottShaham2020ASAP}.  We train our generator model adversarially with a multi-scale patch discriminator, as suggested by pix2pixHD~\cite{wang2018pix2pixHD}.  
The training includes an adversarial hinge-loss, a perceptual loss and a discriminator feature matching loss. The learning rates for the generator and discriminator are $0.0001$ and $0.0004$, respectively.  We use the ADAM~\cite{Kingma2015AdamAM} solver with $\beta_1= 0$ and $\beta_2= 0.999$, following~\cite{park2019semantic,RottShaham2020ASAP}. 
To generate sparse labels, we look at spatial areas
corresponding to distinct semantic segmentation instances. For the sparsity of $S$, we randomly drop the labels with $S\%$ probability, independently for every label inside every area. Please turn to the supplementary material for a detailed visualization of sparse labels.

\subsection{Datasets}
We experiments on three different datasets to demonstrate the versatility and effectiveness of our approach.

\noindent\textbf{Taskonomy} dataset~\cite{taskonomy2018} is a multi-label annotated dataset of indoor scenes. The entire dataset consists of over 4 Million images from around 500 different buildings with high resolution RGB images, segmentation masks and other labels. We chose this dataset for our main experiments because of its wide selection of both semantic and geometric labels. In our experiments, we use the following labels: semantic segmentation, depth, surface normals, edges and curvature. We selected two buildings from the large dataset resulting a total of 18,246 images, split into 14,630/3,619 training/validation images.

\noindent\textbf{Cityscapes} dataset~\cite{Cordts_2016_CVPR} contains images of urban street scenes from 50 different cities and their dense pixel annotations for 30 classes. The training and validation split contains 3000 and 500 samples respectively. To obtain further labels, we use a state-of-the art depth estimation network~\cite{monodepth17}  for depth. The estimated depth, along with the camera intrinsics were used to compute the local patch-wise surface normals. Additionally, we use canny filters for edge detection. This resulted four labels for Cityscape. 

\noindent\textbf{NYU depth v2} dataset~\cite{Silberman:ECCV12} consists of 1449 densely labeled pairs of aligned RGB images and depth, normals, semantic segmentation and edges maps of indoor scenes. We split the data into 1200/249 training/validation sets.

The presented qualitative and quantitative results are generated on the respective hold-out test sets, with the exception of Figure~\ref{fig::nyu}, where we follow the protocol of~\cite{wang2018pix2pixHD} for comparison with the other methods.

\begin{figure}[t!]
    \centering
    \includegraphics[width=7.4cm]{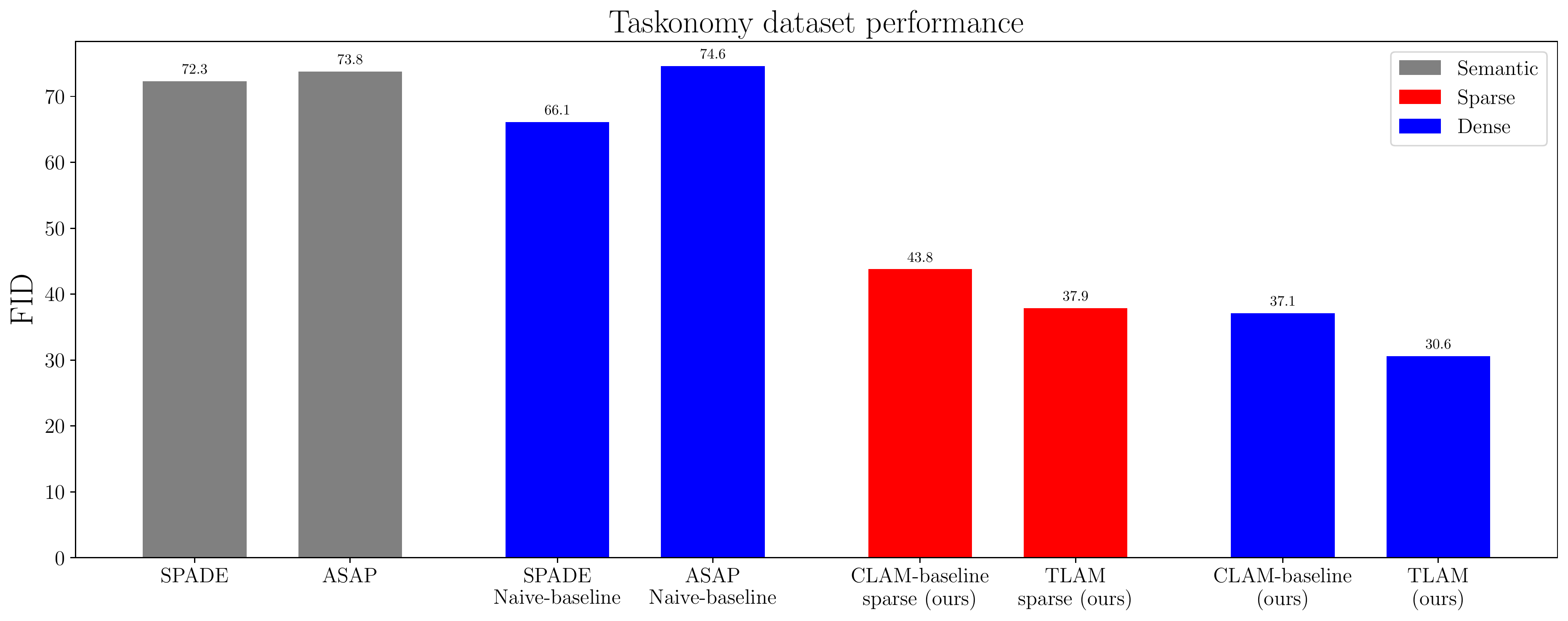}
    \vspace{-2mm}
    \caption{\textbf{Model comparison.} Our TLAM method preforms better compared to the established baselines as well as to the SotA models which use semantic segmentation labels only.}
    \label{fig:bar-fid}
    \vspace{-5pt}
\end{figure}

\subsection{Baselines and Metrics}
Since this is the first work for spatially multi-conditional image generation, we construct our own baselines. 

\noindent\textbf{Naive-baseline} takes all the available inputs and concatenates them along the channel dimension to create the input, which is then fed as an input to the ASAP-Net or SPADE backbone. This also serves as an ablation to study the efficacy of the concept generation component in TLAM.
 
\noindent\textbf{Convolutional Label Merging (CLAM-baseline)} is another baseline that stacks multiple consecutive blocks similar to the projection block from Section~\ref{sec:projection_block}. The first block is exactly the projection block~\eqref{eq:projection_block}, while the following $l$ blocks perform the same operation with $\mathsf{A}^{l}_k \in \mathbb{R}^{d \times d}, \mathsf{b^{l}_k} \in \mathbb{R}^{d}$, preserving the dimensionality $d$. After the final block, all output elements corresponding to the same spatial location are averaged to produce the \emph{Concept Tensor} $\mathsf{Z}$, just like at the end of the TLAM block. 

\noindent\textbf{SotA semantic-only methods.} We also compare compare our method with state-of-the-art semantic image synthesis models, including SPADE \cite{park2019semantic}, ASAP-Net~\cite{RottShaham2020ASAP}, CRN~\cite{crn}, SIMS~\cite{sims}, Pix2Pix~\cite{pix2pix2017} and Pix2PixHD~\cite{wang2018pix2pixHD}. 

\noindent\textbf{Performance Metrics.} We follow the evaluation protocol of previous methods \cite{RottShaham2020ASAP,park2019semantic}. We measure the quality of generated images using the Fréchet Inception Distance (FID) \cite{Heusel2017GANsTB}, which compares the distribution between generated data and real data. The FID summarizes how similar two groups of images are in terms of visual feature statistics.
The second metric is the segmentation score, obtained by evaluating mean-Intersection-over-Union (mIoU) and pixel accuracy (accu) of a semantic segmentation model applied to the generated images.
We use state-of-the-art semantic segmentation networks DRN-D-105~\cite{Yu2017} for Cityscapes and DeepLabv3plus~\cite{cao2021shapeconv} for NYU depth v2 dataset.

\subsection{Quantitative Results} 
\noindent\textbf{Image Generation.} 
Table~\ref{taskonomy-tbl} reports the FID scores obtained on the Taskonomy dataset. We compare our method with several baselines and with different label sparsity. 
Our TLAM significantly outpreforms SPADE and ASAP SotA methods, which use only semantic segmentation maps as inputs. This shows that using different spatial input labels can indeed improve the generation quality.
Moreover, when SPADE and ASAP merge multiple labels as an input, they do not preform significantly better than when using just the semantics. This emphasizes the difficulty of merging multiple spatial labels, which are heterogeneous in nature. Some labels represent semantic image properties, while others represent geometric properties. Also, some labels are continuous, while others are discrete.
Furthermore, our TLAM preforms better than the CLAM-baseline, showing the value of having a better label merging network to deal with the heterogeneity present in the input labels.
Finally, we compare TLAM and the CLAM-baseline with dense and sparse labels. As expected, having dense labels achieves better image quality. The visual quality when using 50\% sparse labels is close to when using dense labels. This is interesting and desirable, since in practical scenarios one often ends up having sparse labels~\cite{popovic2021compositetasking}.


The results on Cityscapes and NYU are summarized in Tables~\ref{cityscales-tbl} and~\ref{nyu-tbl}. 
On the Cityscape dataset, we compare our method with the SotA methods and report FID, and segmentation mIoU and accuracy. Our method achieves better accuracy compared to the other methods. As reported in~\cite{park2019semantic} the SIMS model produces a lower FID, but has poor segmentation accuracy on the Cityscapes.  This is because SIMS synthesizes an image by first stitching image patches from the training dataset. 
On the NYU dataset, our method achieves better mIoU and accuracy. Unfortunately, we are unable to report the FID, due to the small size of only 249 images in the validation set.  

\begin{figure*}[t!]
    \label{fig:lab}
    \centering
    \captionsetup{font=small}
    \begin{subfigure}[b]{0.31\linewidth}
        \centering
        \includegraphics[width=\linewidth]{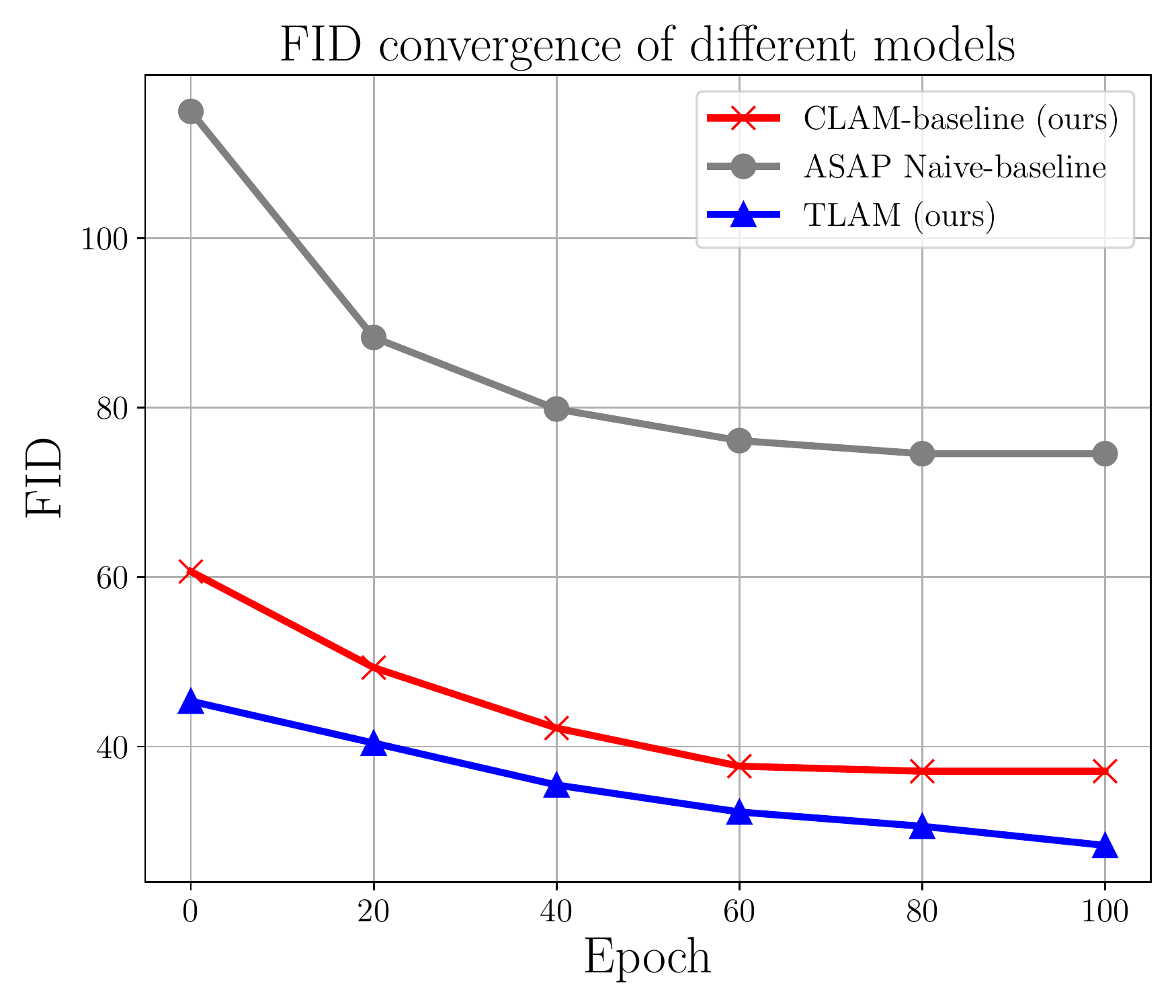}
        \captionof{figure}{\textbf{Convergence of different models.} We compare FID scores during training of our TLAM method to the Naive and CLAM baselines. Our method converges faster and achieves better FID compared to those of the baselines at every point during training.}
        \label{fig:convergence-plot}
    \end{subfigure}
    \hfill 
    \begin{subfigure}[b]{0.31\linewidth}
        \centering
        \includegraphics[width=\linewidth]{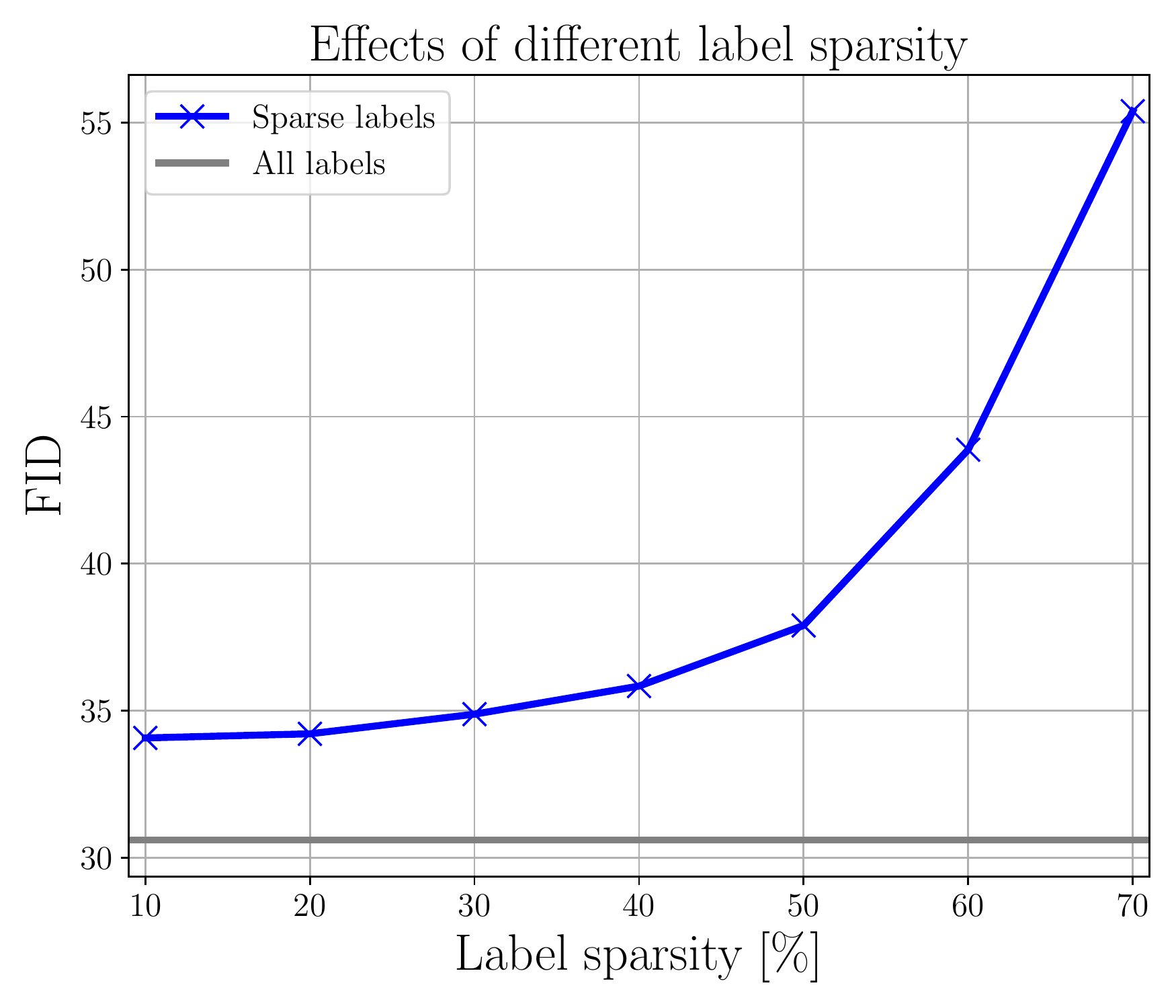}
        \captionof{figure}{\textbf{Effects of different label sparsity.} Our model achieves good FID score even when a fraction of labels is presented during inference. With more labels present, our model achieves better performance, even though it was trained for $50 \%$ label sparsity.
        }
        \label{fig:sparsity}
    \end{subfigure}
    \hfill    
    \begin{subfigure}[b]{0.31\linewidth}
        \centering
        \includegraphics[width=\linewidth]{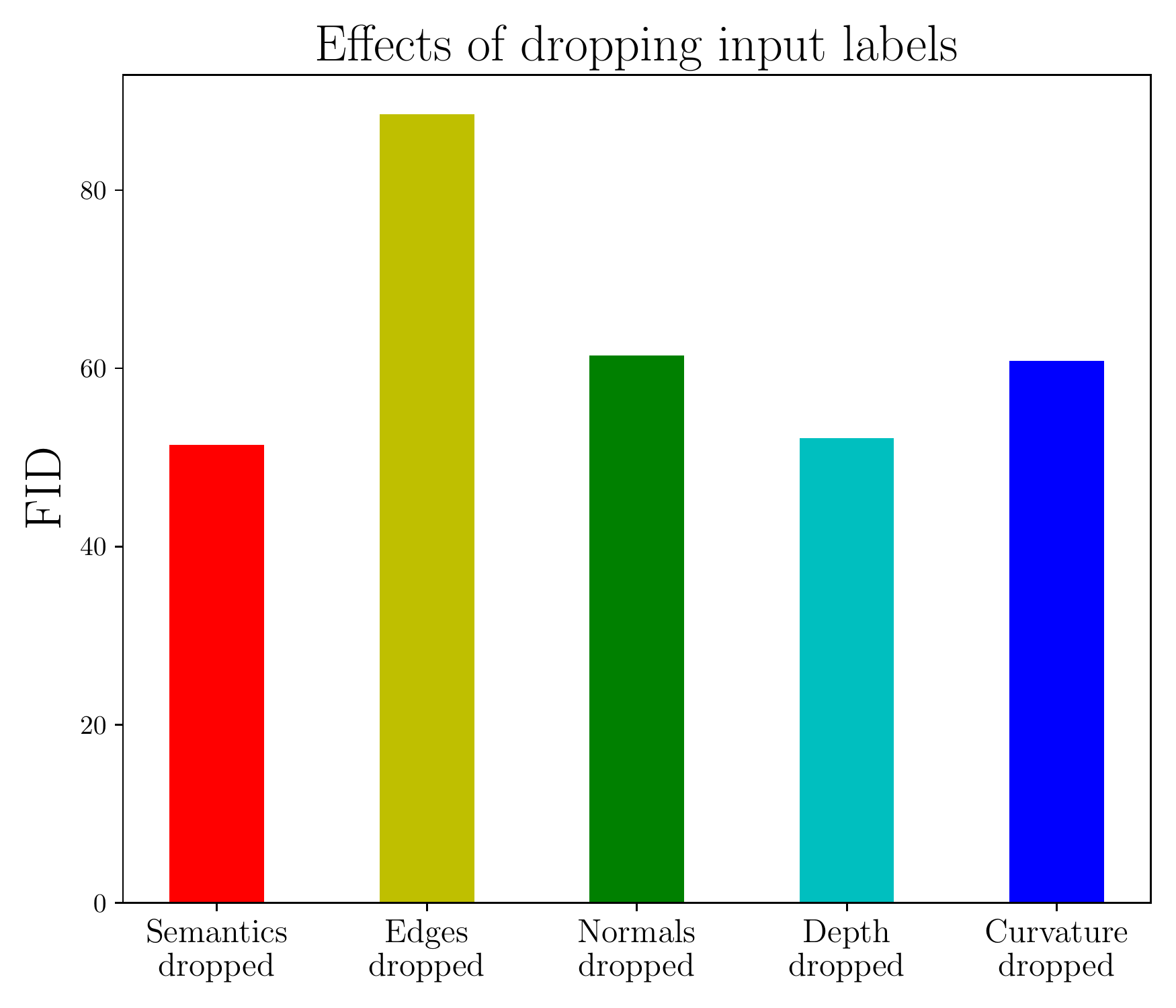}
        \captionof{figure}{\textbf{Effects of dropping input labels.} We examine how dropping a specific label affects the image quality of our TLAM model. Dropping any label degrades FID which leads to the conclusion that all labels provide useful information for image synthesis.}
        \label{fig:l3}
    \end{subfigure}
    \caption{\textbf{Training convergence and label sensitivity.} 
    We analyse behavioural aspects of our models with regards to training and labels.}
    \vspace{-6pt}
\end{figure*}

\noindent\textbf{Training Convergence.} 
Figure~\ref{fig:convergence-plot} shows the evolution of FID during training on Taskonomy. We evaluate the FID on the validation set every 20 epochs for TLAM, and the ASAP naive and CLAM baselines. 
One can observe that TLAM quickly achieves a very good FID, even after 20 epochs. At this instance, TLAM has 2.5 and 1.3 times better FID than those of the naive and CLAM baseline, respectively. 
We can also see that TLAM converges faster than the other models. 
One training epoch in Figure~\ref{fig:convergence-plot} takes about 1.7hrs for our method on one GeForce GTX TITAN X GPU. 

\begin{table}[t!]
\centering
\tableFont
\begin{tabular}{r||c|c|c}
	{Method} & mIoU  & Accuracy & FID\\
	\Xhline{2\arrayrulewidth}
	CRN~\cite{crn} & 52.4 & 77.1 & 104.7\\
	SIMS~\cite{sims} & 47.2 & 75.5 & \textbf{49.7}\\
	Pix2Pix~\cite{pix2pix2017} & 39.5 & 78.3 & 80.7 \\
	Pix2PixHD~\cite{wang2018pix2pixHD} & \underline{58.3} & 81.4 & 95.0 \\
SPADE~\cite{park2019semantic} & \textbf{62.3} & \underline{81.9} & 71.8 \\
\hline
ASAP~\cite{RottShaham2020ASAP} & 44.9 & 78.6 & 72.5 \\
TLAM (Ours) & 45.5 & \textbf{85.3} & \underline{68.3} \\
\hline
\end{tabular}
\caption{\textbf{Quantitative results on Cityscapes.} 
Our method uses ASAP as backbone and improves it by a significant margin.}\label{cityscales-tbl}
\vspace{-4px}
\end{table}

\begin{table}[t!]
	\centering
	\tableFont
    \begin{tabular}{r||c|c}
		Method & mIoU  & Accuracy \\
		SPADE~\cite{park2019semantic} & 33.1 & 47.4 \\
				\hline
		ASAP~\cite{RottShaham2020ASAP} & \underline{36.2} & \underline{49.1} \\
		TLAM (Ours) & \textbf{38.3} & \textbf{53.1} \\
		\hline
	\end{tabular}
	\caption{\textbf{Quantitative results on NYU.} Our method demonstrates the benefit of multi-conditioning using the ASAP backbone.}\label{nyu-tbl}	
	\vspace{-6pt}
 \end{table}

\subsection{Qualitative Results} 
To visually compare our TLAM label merging, we present qualitative results for both dense and sparse labels on the Taskonomy dataset in Figure~\ref{fig::taskonomy}, together with the semantic-only baselines. TLAM with all-dense and all-sparse labels generates high-fidelity images. Our method generates fine structural details such as lights, decorations, and even mirror reflections clearly better than SPADE and ASAP. 
The results on the Cityscapes dataset are shown in Figure~\ref{fig::cityscapes}. Notably, our method with sparse labels achieves similar visual quality as other methods.
Figure~\ref{fig::nyu} shows qualitative results on the NYU dataset, where Pix2PixHD also renders images with good quality, however it fails to capture the lighting conditions in contrast to our method. Notably, our method captures the rich geometric structures (such as on the ceiling), thanks to the geometric labels. Please, refer to our supplementary material for more visual results.
Overall, the qualitative results demonstrate the effectiveness of TLAM, which exploits novel pixel-wise label transformers, even for sparse labels.

\subsection{Label Sensitivity Study} 
We analyse the sensitivity of our method with regards to the provided labels on the Taskonomy dataset.

\noindent\textbf{Label Sparsity.}
Figure \ref{fig:sparsity} shows how the FID is affected by label sparsity. Experiments conducted on increasing sparsity from 10\% to 70\%, show a steady degradation of FID with less available labels. 
\emph{Note that our model already achieves good FID using only 30\% of available labels}. 
The experiments were conducted using a single model trained with 50\% sparsity. It also shows the generalizability of our method across various levels of sparsity.

\noindent\textbf{Removal of Labels.} In Figure~\ref{fig:l3}, we plot the FID after removing each label from the input, using the TLAM model trained on Taskonomy with 50\% sparsity. We observe that among the five labels, edges play the most significant role. On the other hand, the semantics and depth are the most dispensable. Nevertheless, the removal of any label results into a worsening of the FID at least by  a factor of 1.3. This suggests that all labels provide different information crucial for image generation, while being mutually complimentary. 

We conclude that the proposed label merging block can successfully deal with incomplete labels and is able to exploit information from all available labels.



\subsection{Concept Visualization and Image Editing}
\begin{figure*}[t!]
    \addtolength{\tabcolsep}{-4.7pt}
    \centering
    \begin{tabular} {ccccccc}
    Concept & Image & Semantics & Normals & Edges & Depth & Curvature \\
    \includegraphics[width=0.137\linewidth]{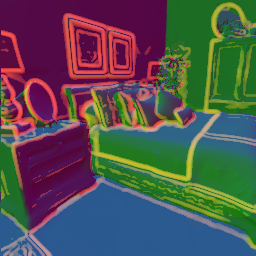} &
    \includegraphics[width=0.137\linewidth]{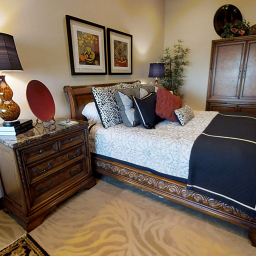}&
    \includegraphics[width=0.137\linewidth]{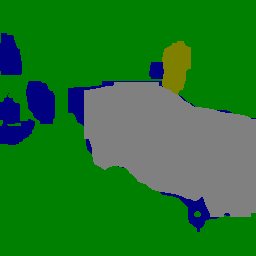}&
    \includegraphics[width=0.137\linewidth]{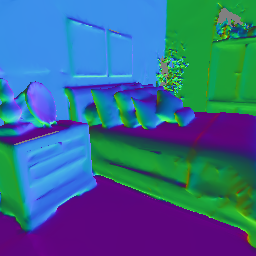}&
    \includegraphics[width=0.137\linewidth]{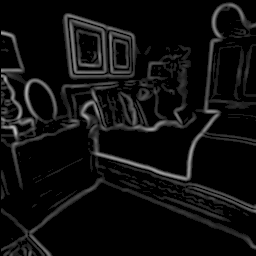}&
    \includegraphics[width=0.137\linewidth]{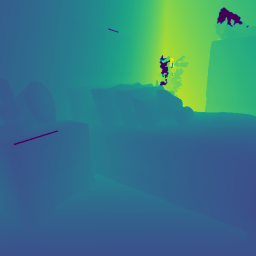}&
    \includegraphics[width=0.137\linewidth]{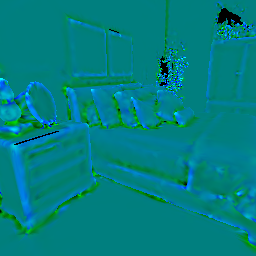}
    \end{tabular}
	\caption{\textbf{Concept Tensor visualization.} From left to right: \emph{Concept Tensor} projected to RGB; original image; five different input labels.}
	\label{fig::pca}
\end{figure*}

\textbf{Concept Visualization.} To visualize the concept tensor $\mathsf{Z}\in\mathbb{R}^{H\times W \times 96}$, we project it to 3 channels, using  Principal Component Analysis~\cite{FRSLIIIOL}, and present it in  
Figure~\ref{fig::pca}, along with the corresponding image and labels. One can observe that the visualized concept tensor indeed resembles different aspects of the input labels (e.g. edges and normals). 


 


\input{figures/paper/cityscapes}
\input{figures/paper/nyu}

\begin{figure*}[t!]
    \centering
    \begin{tabular} {ccccc}
    Original & Mask & SPADE~\cite{park2019semantic} & ASAP~\cite{RottShaham2020ASAP}  & TLAM (Ours)  \\
    
    \includegraphics[width=0.135\linewidth]{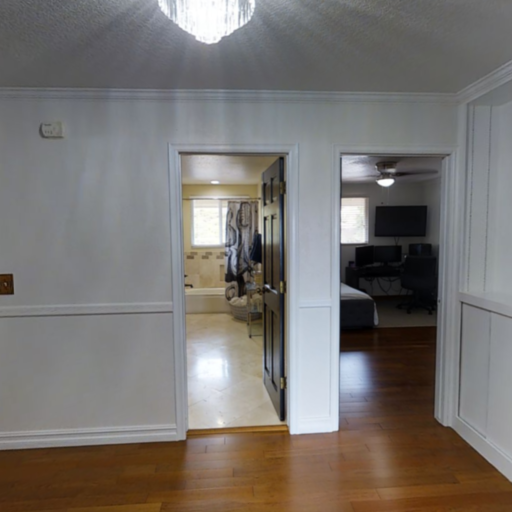} &
    \includegraphics[width=0.135\linewidth]{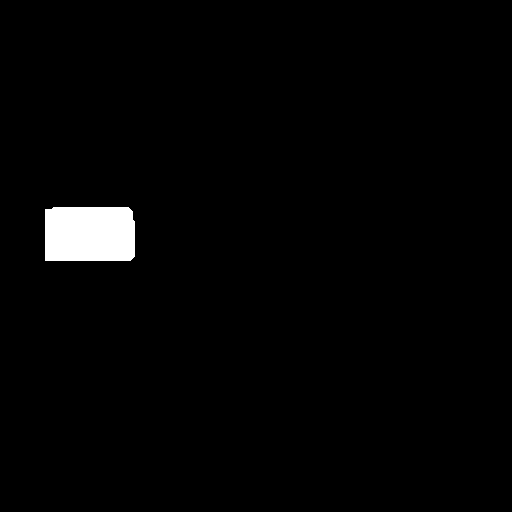} &
    \includegraphics[width=0.135\linewidth]{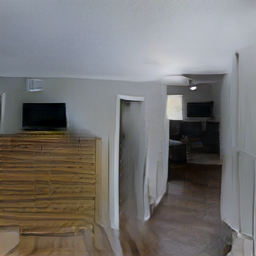} &
    \includegraphics[width=0.135\linewidth]{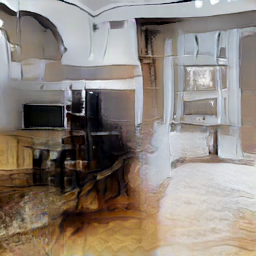} &
    \includegraphics[width=0.135\linewidth]{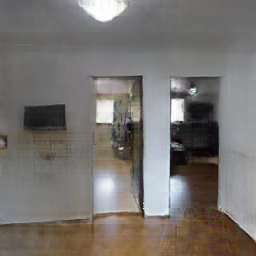}
    \end{tabular}
	\caption{\textbf{Object insertion.} 
	From left to right: original image; mask of the inserted object; generated image using SPADE, ASAP and our method, respectively.
	This figure shows insertion of a TV, by inserting labels provided by the user at the location she/he chooses.
	}
	\vspace{8pt}
	\label{fig:object_removal_insertion}
\end{figure*}

\noindent\textbf{Geometric Image Editing with User Inputs.}
In order to demonstrate an intuitive application of our method, we perform image editing by inserting a new object into the scene. Figure~\ref{fig:object_removal_insertion} shows how our method can mimic rendering while allowing the geometric manipulation of an image. We render a TV in the given image, by simply augmenting different labels with user-provided labels for the TV. 

%% file: figures/paper/cityscapes.tex
\begin{figure*}[t!]
\centering
\addtolength{\tabcolsep}{-4.5pt}
\begin{tabular} {cc|cc|cc}
 Label & Reference & SPADE~\cite{park2019semantic} & ASAP~\cite{RottShaham2020ASAP} & \makecell{TLAM \\(Ours)} & \makecell{Sparse-TLAM \\(Ours)}\\
 
 \includegraphics[width=0.16\linewidth]{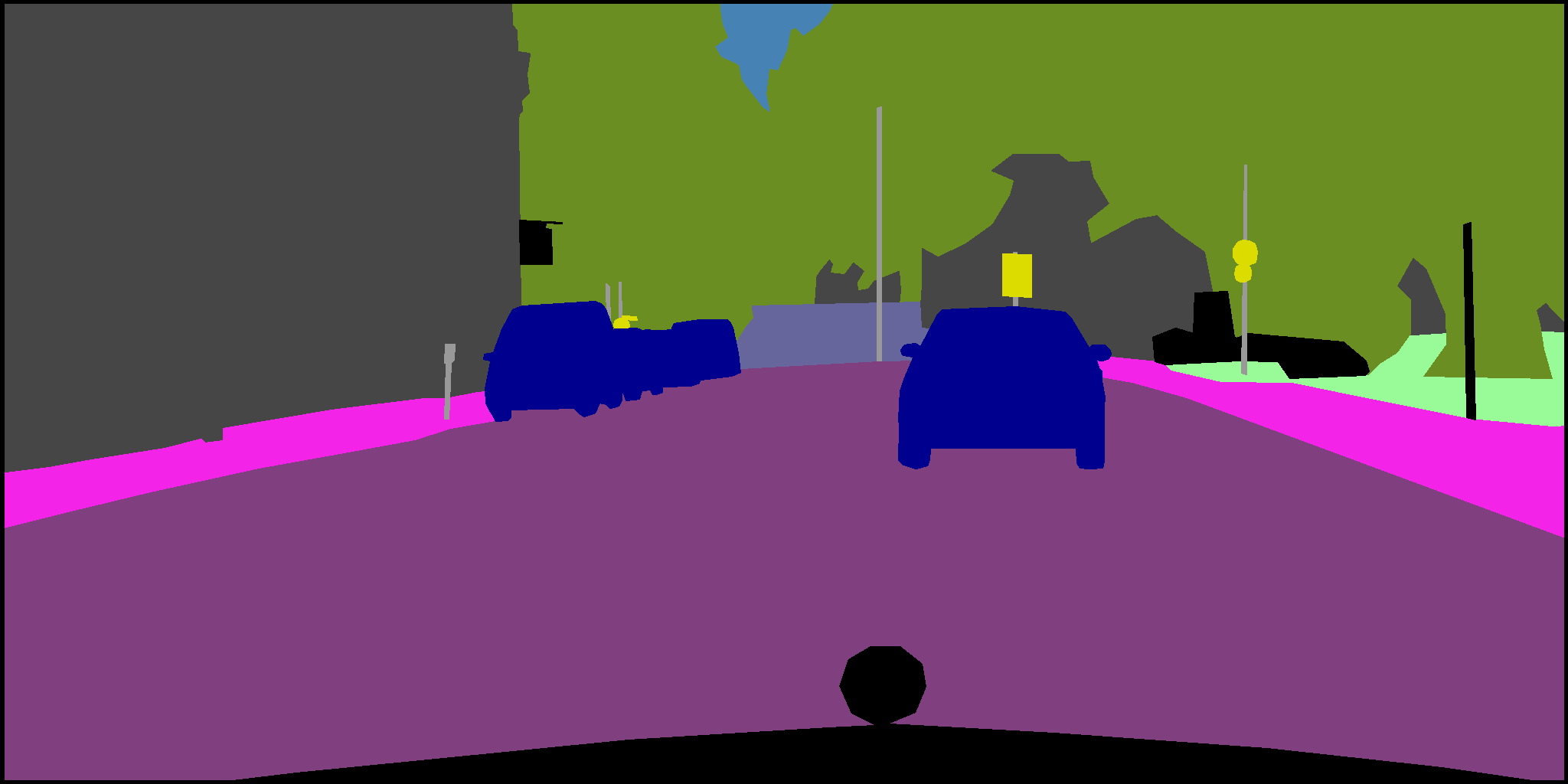}&
 \includegraphics[width=0.16\linewidth]{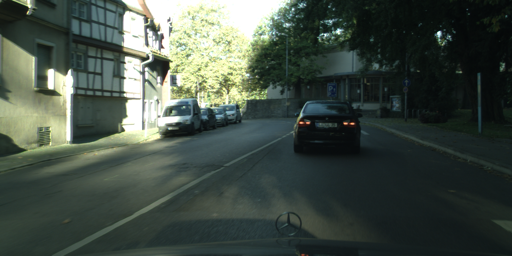} &
\includegraphics[width=0.16\linewidth]{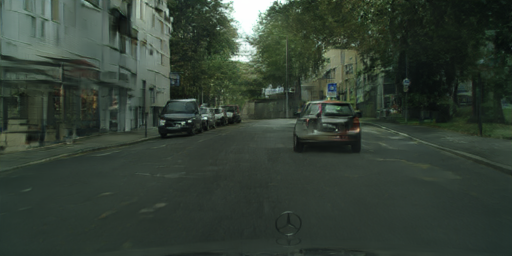} &   
\includegraphics[width=0.16\linewidth]{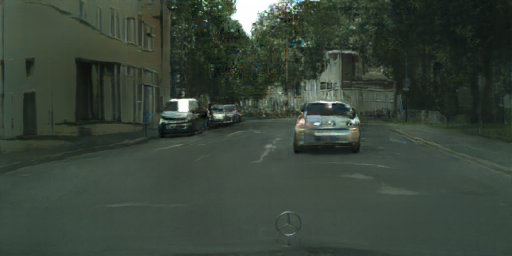} &
\includegraphics[width=0.16\linewidth]{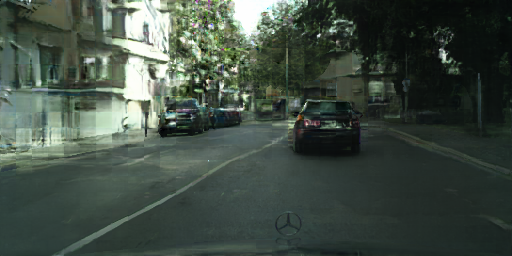} &  
\includegraphics[width=0.16\linewidth]{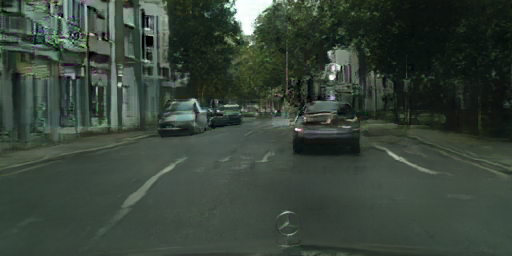} \\

\includegraphics[width=0.16\linewidth]{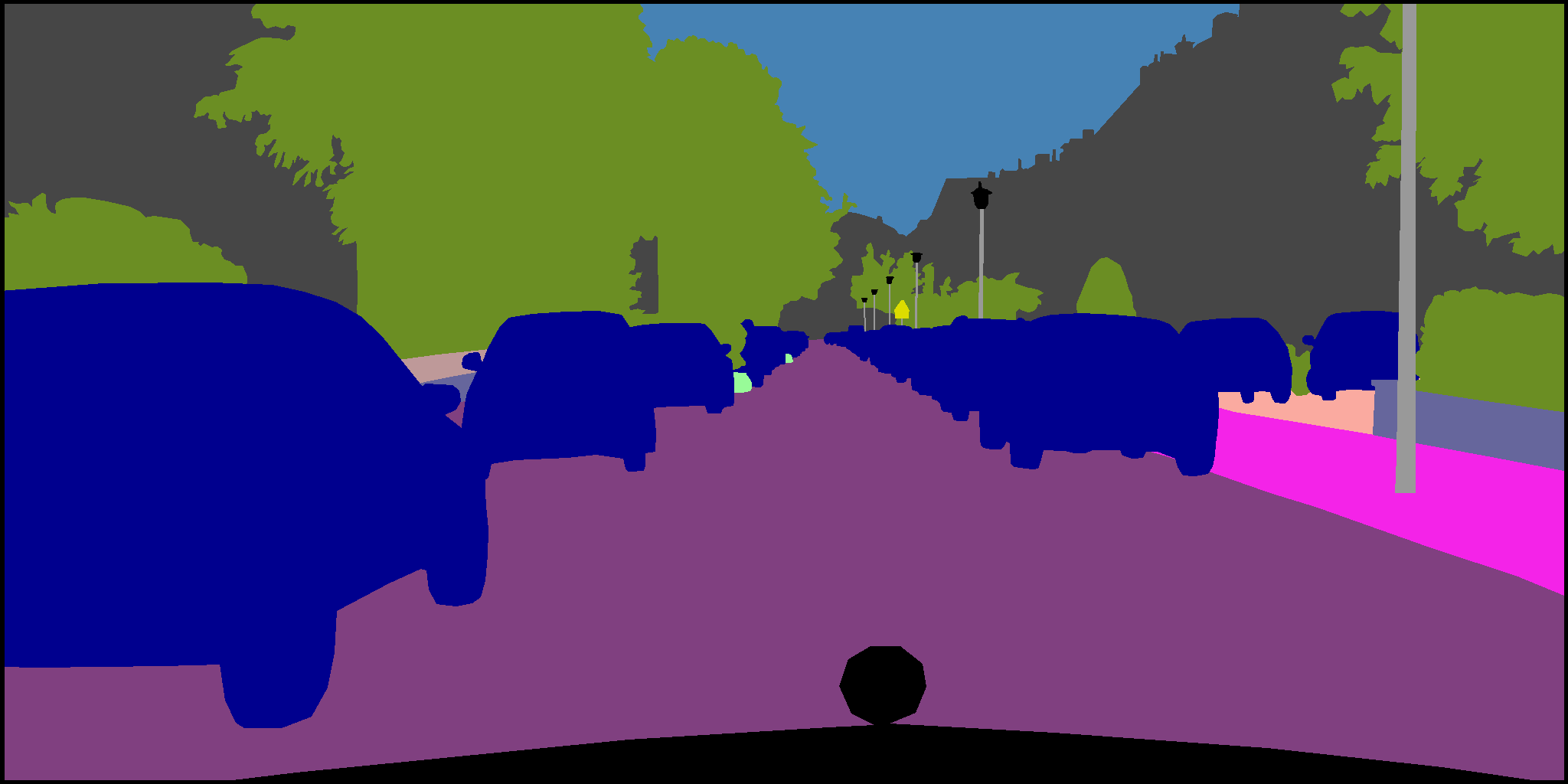}& 
\includegraphics[width=0.16\linewidth]{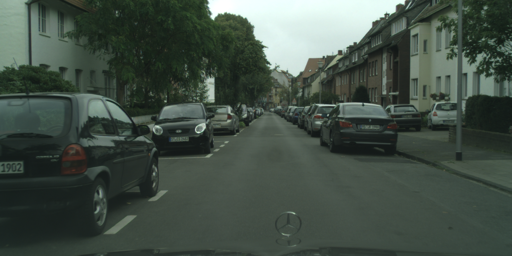} &
\includegraphics[width=0.16\linewidth]{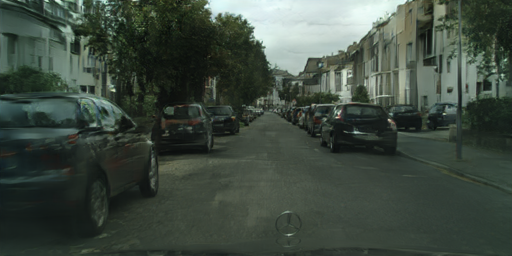} &   
\includegraphics[width=0.16\linewidth]{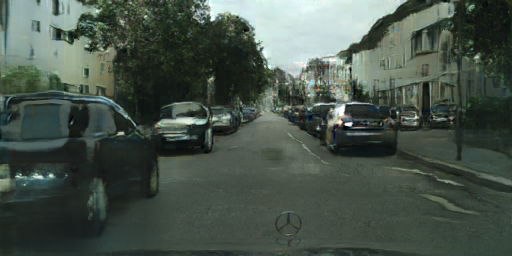} &
\includegraphics[width=0.16\linewidth]{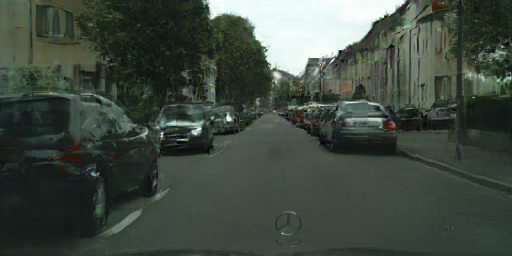} &  
\includegraphics[width=0.16\linewidth]{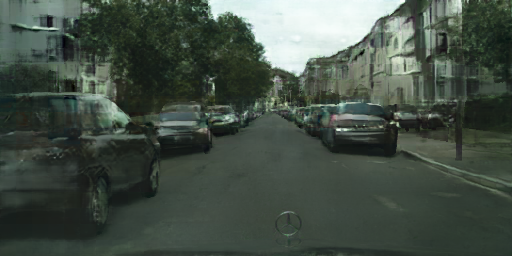} \\

\includegraphics[width=0.16\linewidth]{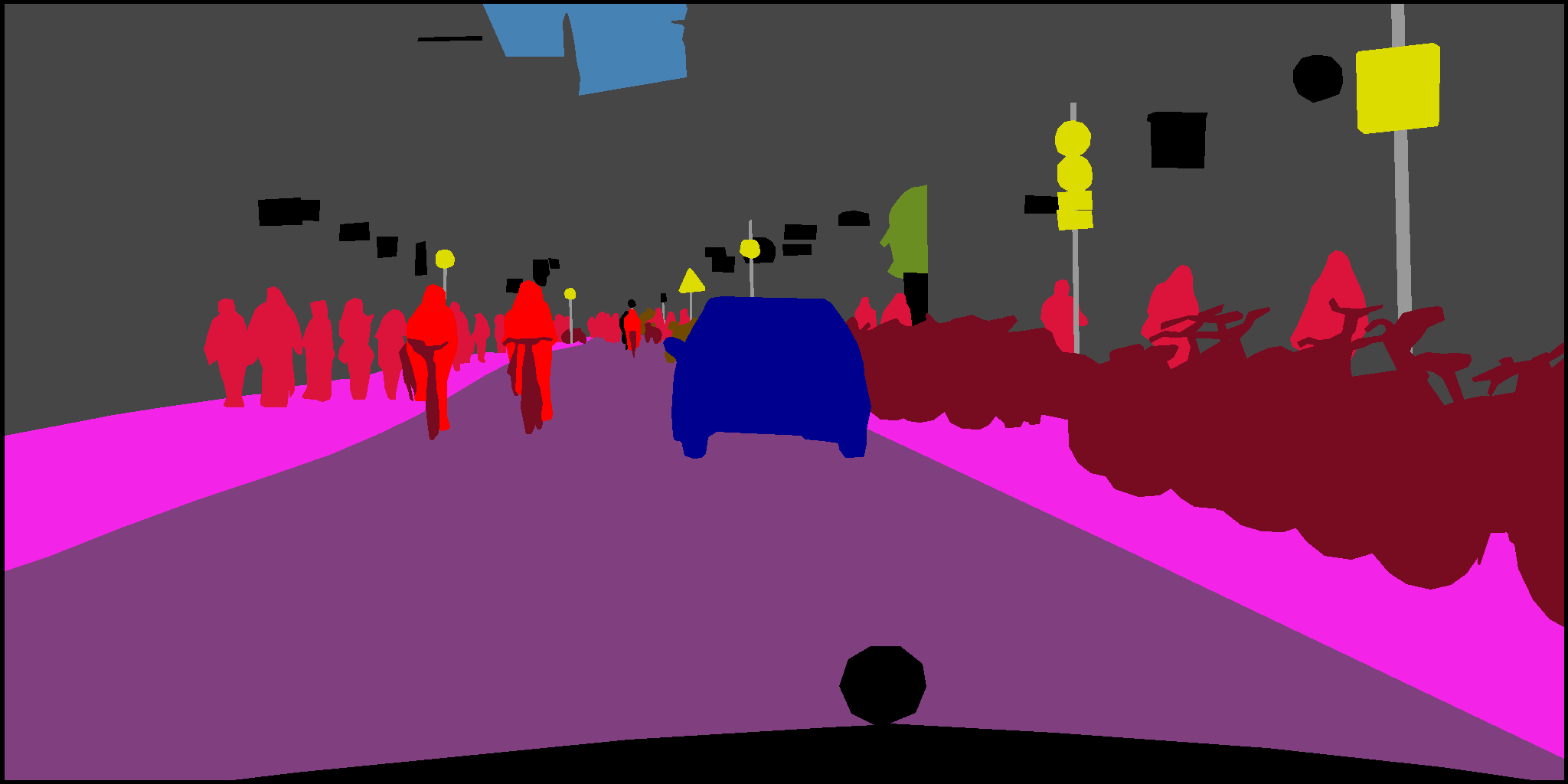}& 
\includegraphics[width=0.16\linewidth]{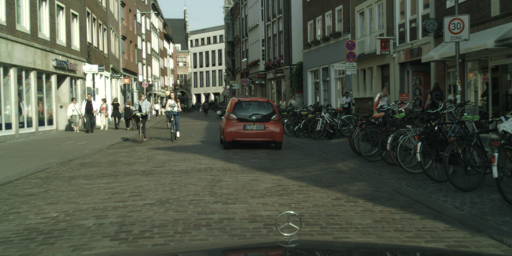} &
\includegraphics[width=0.16\linewidth]{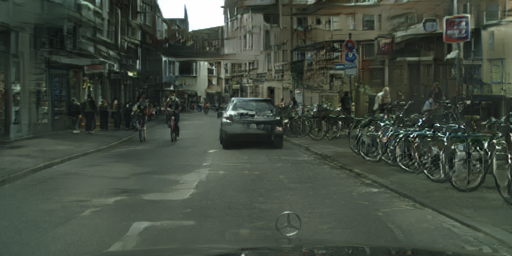} &   
\includegraphics[width=0.16\linewidth]{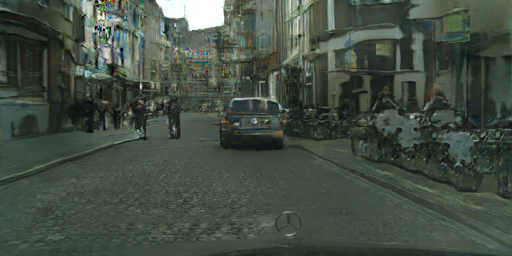} &
\includegraphics[width=0.16\linewidth]{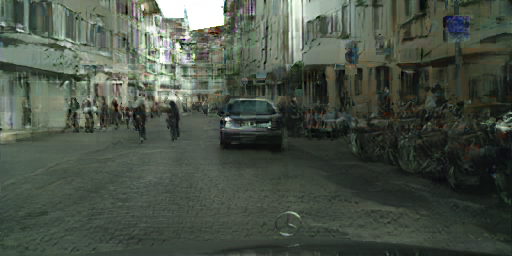} &  
\includegraphics[width=0.16\linewidth]{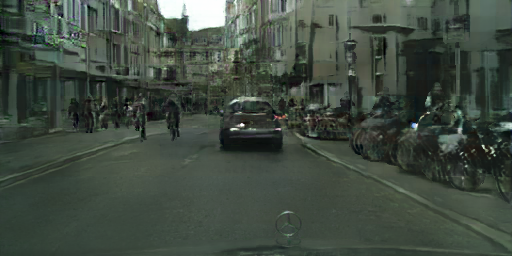}
\end{tabular}
	\caption{\textbf{Visual comparison on the Cityscapes.} Our approach achieves a visual quality on par with the compared methods.}
	\label{fig::cityscapes}
\end{figure*}

%% file: figures/paper/nyu.tex
\begin{figure*}[t!]
\centering
\footnotesize
\addtolength{\tabcolsep}{-3.5pt} 
\begin{tabular} {cccccc|c}
Reference & Pix2pix~\cite{pix2pix2017} & CRN~\cite{crn} & pix2pixHD~\cite{wang2018pix2pixHD} & SPADE~\cite{park2019semantic} & ASAP~\cite{RottShaham2020ASAP} & TLAM (Ours) \\
 
\includegraphics[width=0.132\linewidth]{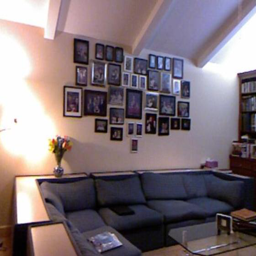} &
\includegraphics[width=0.132\linewidth]{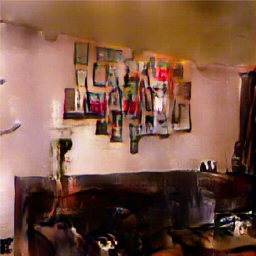} &   \includegraphics[width=0.132\linewidth]{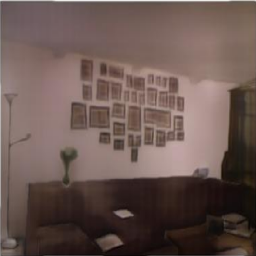} &
\includegraphics[width=0.132\linewidth]{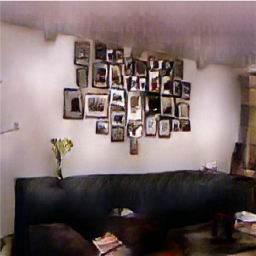} &  \includegraphics[width=0.132\linewidth]{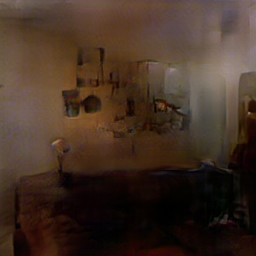} &
\includegraphics[width=0.132\linewidth]{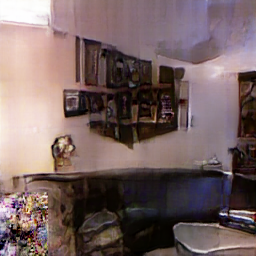} &
\includegraphics[width=0.132\linewidth]{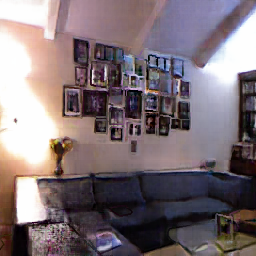} 
\\
\includegraphics[width=0.132\linewidth]{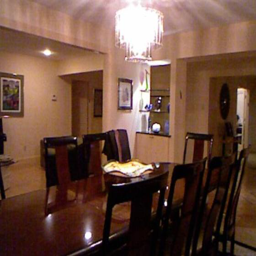} &
\includegraphics[width=0.132\linewidth]{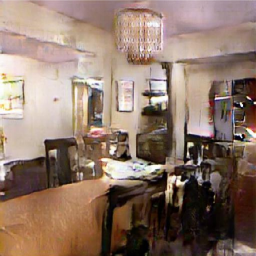} &   \includegraphics[width=0.132\linewidth]{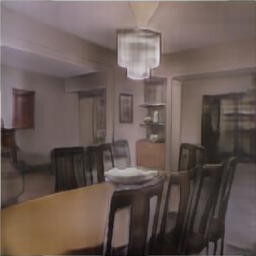} &
\includegraphics[width=0.132\linewidth]{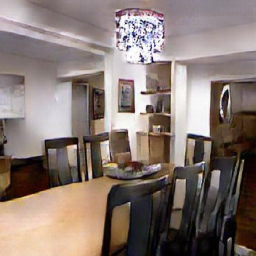} &  \includegraphics[width=0.132\linewidth]{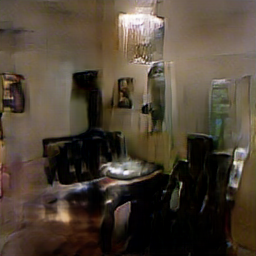} &
\includegraphics[width=0.132\linewidth]{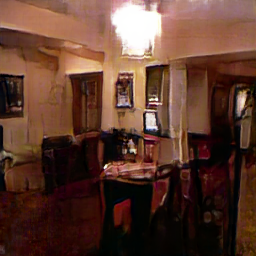} &
\includegraphics[width=0.132\linewidth]{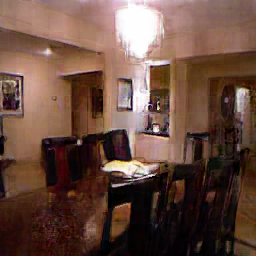} 
\end{tabular}
	\caption{\textbf{Visual comparison on NYU.} Our method generates images that better capture the lighting and geometry with  more details.}
	\label{fig::nyu}
\end{figure*}

%% file: latex_submission/paper/0x_conclusion.tex
\section{Conclusion}
\label{sec:conclusion}
In this work, we offer a new perspective on image generation as inverse of image understanding. In the same way as image understanding involves the solving of multiple different tasks, we desire the control over the generation process to include multiple input labels of various kinds. 
To this end, we design a neural network architecture that is capable of handling sparse and heterogeneous labels, by mapping them to a homogeneous concept space in a pixel-wise fashion. 
With our proposed module, we can equip spatially conditioned generators with the desired properties. From our experiments on challenging datasets, we conclude that the benefits and flexibility of this additional layer of control gives way to exciting results beyond the state-of-the-art.



%% file: latex_submission/supplementary/all-tlam.tex
This document provides additional details, which complement the main paper. We provide the complete network diagram of our generator in Section~\ref{sec:generator_overview_supp}. More implementation details are contained in in Section~\ref{sec:implementation_details_supp}. The visualization of input labels under different sparsity can be found in Section~\ref{sec:input_labels_sup}. In Section~\ref{sec:geometric_editing_supp}, we demonstrate an intuitive application of our method. Additional qualitative results are presented in Section~\ref{sec:qualitative_res_sup}. Finally, a discussion about ethical and societal impacts is contained in Section~\ref{sec:ethical_sup}.
Also, please refer to our video supplementary for the example of geometric editing. 


\section{Generator Overview}

\label{sec:generator_overview_supp}
\noindent{\textbf{Whole Network.}} The network diagram of our generator is presented in Figure~\ref{fig:network-overview}.
The figure shows how our proposed Label Merging Block TLAM is connected to the ASAP-Net~\cite{RottShaham2020ASAP} generator, which we use as the backbone.

\noindent{\textbf{ASAP-Net Generator.}} The generator takes the \emph{Concept Tensor} $\mathsf{Z} \in \mathbb{R}^{H \times W \times d}$ as an input, similar to taking the semantic labels in the original design. It then outputs a tensor of weights, which parameterize the pixelwise spatially-varying multi-layer perceptrons (MLPs). The MLPs, with infered weights, compute the final output image by taking the \emph{Concept Tensor} $\mathsf{Z}$ as their input.

\section{Implementation Details}
\label{sec:implementation_details_supp}
\noindent{\textbf{Projection Block.}} The Projection Block projects each input label into an embedding space with the same dimensionality. Every input label is processed by one 1x1 convolution layer, followed by a nonlinear activation. The projection is a 96-dimensional tensor for each label.

\noindent{\textbf{Concept generation Block.}} The Concept Generation Block takes the projected tensors as input, and operates over them in a pixel-wise fashion. For each pixel, we thus have an embedding vector for each task. We add a label specific-encoding to the embedding vector of each label, as a way of signalizing which task the embedding corresponds to. For every pixel, the concept generation block processes each set of encoded labels with a transformer. The transformer consists of 3 layers, where each uses 3 heads. 

\section{Visualization of input Labels}
\label{sec:input_labels_sup}
In Figure~\ref{fig:all-lbl} we show examples of the input labels from the Taskonomy dataset. These labels include semantic segmentation, normals, depth, edges and curvature.
Furthermore, in Figure~\ref{fig:sparse-lbl1}, \ref{fig:sparse-lbl2} and \ref{fig:sparse-lbl3} we show examples of labels from the Taskonomy dataset, with 70\%, 50\% and 30\% label sparsity.
To generate sparse labels, we look at spatial areas
corresponding to distinct semantic segmentation instances. For the sparsity of $S$, we randomly drop the labels with $S\%$ probability, independently for every label inside every area.
Higher sparsity means there is a higher probability that a semantic region will be masked out for all labels.

\section{Geometric Image Editing with User Inputs}
\label{sec:geometric_editing_supp}
In order to demonstrate an intuitive application of our method, we perform image editing by inserting a new object into the scene. Figure~\ref{tab::geom_edit} shows how our method can mimic rendering while allowing the geometric manipulation of an image. In this application, we render a table in the given image, by simply augmenting different labels to include label information derived from a 3D model. Our method is able to render the table realistically within the image, while ASAP and SPADE perform unsatisfactorily.
Figure~\ref{fig:object_removal_insertion} shows an example of removing a certain object from the image by augmenting the labels.

\input{figures/supplementary/all-lbl-examples}

\input{figures/supplementary/sparse-lbl-example}

\section{Additional Qualitative Results}
\label{sec:qualitative_res_sup}
In  Figure~\ref{fig:all-tlam} we show images synthesized with our TLAM model using dense input labels (semantics, curvature, edges, normals and depth), on the Taskonomy dataset. 
In Figure~\ref{fig:sparse-tlam} we show images synthesized with our TLAM model using sparse input labels (50\% sparsity), on the Taskonomy dataset. Finally, Figure~\ref{fig:cityscapes} shows additional visual comparison on the Cityscapes dataset, where we compare our TLAM and Sparse-TLAM methods with SPADE~\cite{park2019semantic} and ASAP-Net~\cite{RottShaham2020ASAP}.

\section{Ethical and Societal Impact}
\label{sec:ethical_sup}
This work is going one step further into the image generation. While bringing great potential artistic  value to the general public, such technology can be misused for fraudulent  purposes. Despite the generated images looking realistic, this issue can be partially mitigated by learning-based fake detection~\cite{Guarnera2020DeepFakeDB,Wang2020CNNGeneratedIA,Verdoliva2020MediaFA}.

In regards to limitations, our method is not designed with a particular focus on balanced representation of appearance and labels. Therefore, image generation may behave unexpected on certain conditions, groups of objects or people. We recommend application-specific strategies for data collection and training to ensure the desired outcome~\cite{Alvi2018TurningAB,Wang2020TowardsFI}.

\input{figures/supplementary/big_diagram}

\input{figures/supplementary/tlam-all-lbl}
\input{figures/supplementary/tlam-sparse}

\input{figures/supplementary/image_editing}
\input{figures/supplementary/cityscapes}

%% file: figures/supplementary/all-lbl-examples.tex
\begin{figure*}[h]

\addtolength{\tabcolsep}{-9pt}  
\centering
\begin{tabular} {cccccc}
 Image & Semseg & Normals & Depth & Edges & Curvature\\
 
\includegraphics[width=0.13\linewidth]{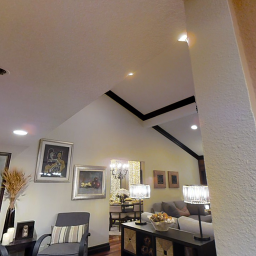}&
\hspace{2.5mm} \includegraphics[width=0.13\linewidth]{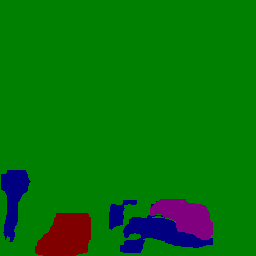} &
\hspace{2.5mm} \includegraphics[width=0.13\linewidth]{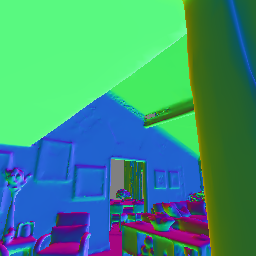} &   
\hspace{2.5mm} \includegraphics[width=0.13\linewidth]{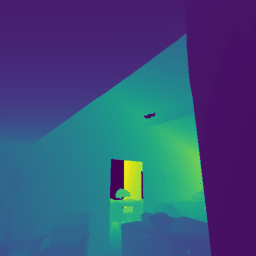} &
\hspace{2.5mm} \includegraphics[width=0.13\linewidth]{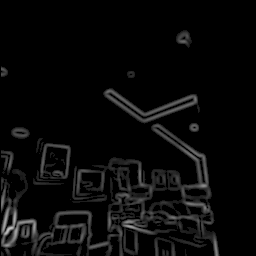} &  \hspace{2.5mm} \includegraphics[width=0.13\linewidth]{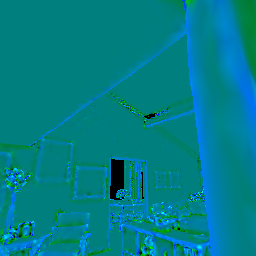} \\

\includegraphics[width=0.13\linewidth]{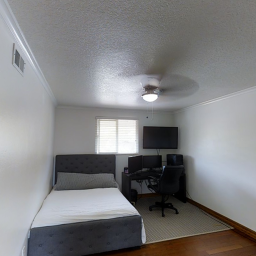}&
\hspace{2.5mm} \includegraphics[width=0.13\linewidth]{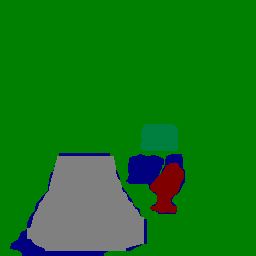} &
\hspace{2.5mm} \includegraphics[width=0.13\linewidth]{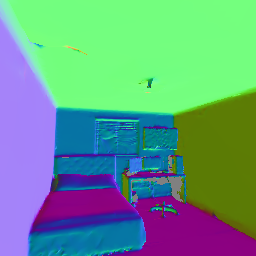} &  
\hspace{2.5mm} \includegraphics[width=0.13\linewidth]{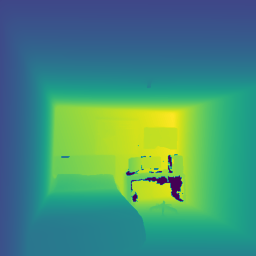} &
\hspace{2.5mm} \includegraphics[width=0.13\linewidth]{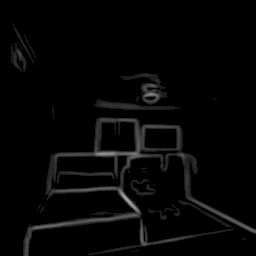} & 
\hspace{2.5mm} \includegraphics[width=0.13\linewidth]{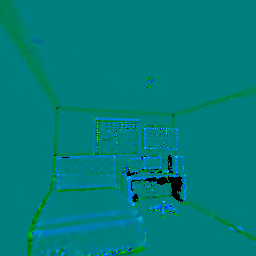} \\

\includegraphics[width=0.13\linewidth]{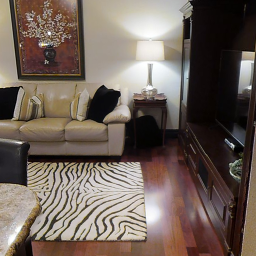}&
\hspace{2.5mm} \includegraphics[width=0.13\linewidth]{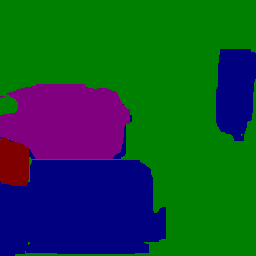} &
\hspace{2.5mm} \includegraphics[width=0.13\linewidth]{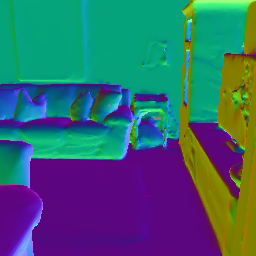} &  
\hspace{2.5mm} \includegraphics[width=0.13\linewidth]{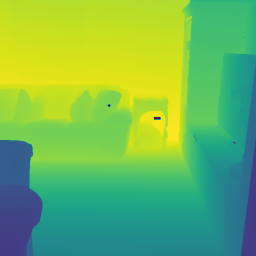} &
\hspace{2.5mm} \includegraphics[width=0.13\linewidth]{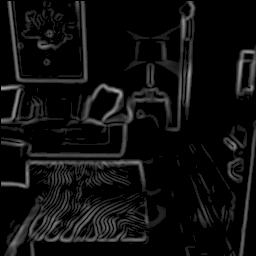} & 
\hspace{2.5mm} \includegraphics[width=0.13\linewidth]{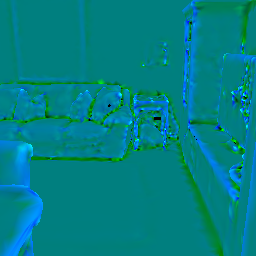} \\

\includegraphics[width=0.13\linewidth]{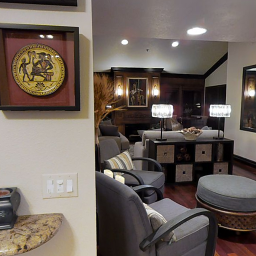}&
\hspace{2.5mm} \includegraphics[width=0.13\linewidth]{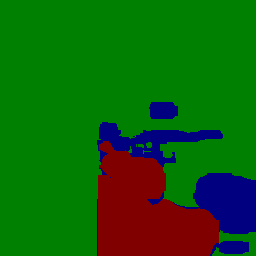} &
\hspace{2.5mm} \includegraphics[width=0.13\linewidth]{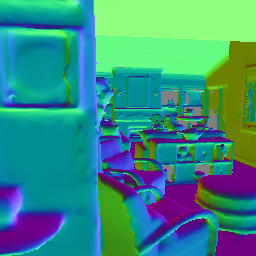} &
\hspace{2.5mm} \includegraphics[width=0.13\linewidth]{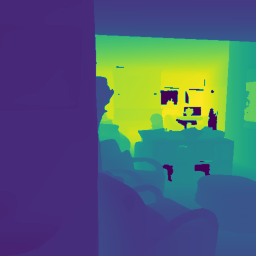} &
\hspace{2.5mm} \includegraphics[width=0.13\linewidth]{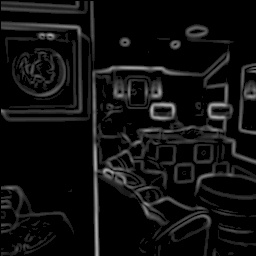} & 
\hspace{2.5mm} \includegraphics[width=0.13\linewidth]{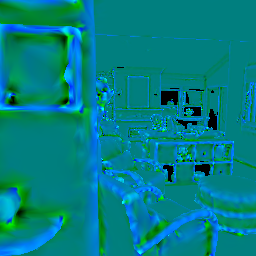} \\

\end{tabular}
	\caption{\textbf{Examples of dense input labels.} Samples of the Taskonomy dataset with all input labels visualized.}
	\label{fig:all-lbl}
\end{figure*}

%% file: figures/supplementary/sparse-lbl-example.tex
\begin{figure*}[h]
\addtolength{\tabcolsep}{-8pt}  
\centering
\begin{tabular}{c}
    Image   \\
    \includegraphics[width=0.14\linewidth]{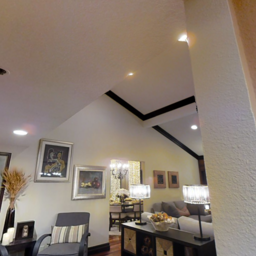}
    \\
\end{tabular}
\\

\begin{tabular} {ccccc}
 Semseg & Normals & Depth & Edges & Curvature\\
 
\multirow{1}{*}[33pt]{{\rotatebox{90}{\makecell{70 \% \\ sparsity}}}}
\includegraphics[width=0.14\linewidth]{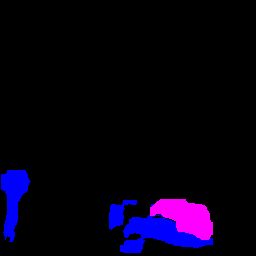} &
\hspace{2.5mm} \includegraphics[width=0.14\linewidth]{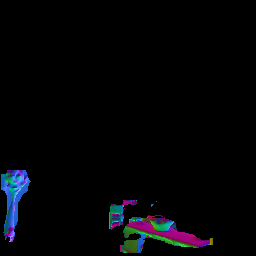} &   
\hspace{2.5mm} \includegraphics[width=0.14\linewidth]{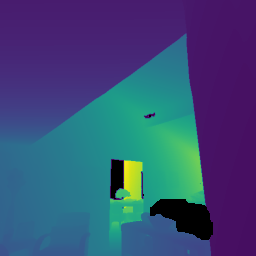} &
\hspace{2.5mm} \includegraphics[width=0.14\linewidth]{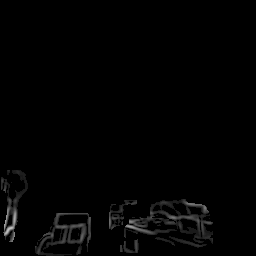} & 
\hspace{2.5mm} \includegraphics[width=0.14\linewidth]{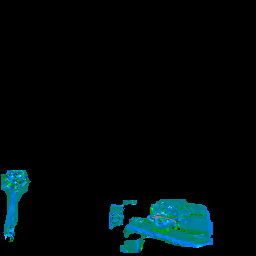} \\

\multirow{1}{*}[33pt]{{\rotatebox{90}{\makecell{50 \% \\ sparsity}}}}
\includegraphics[width=0.14\linewidth]{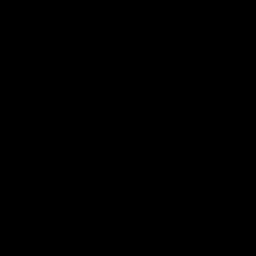} &
\hspace{2.5mm} \includegraphics[width=0.14\linewidth]{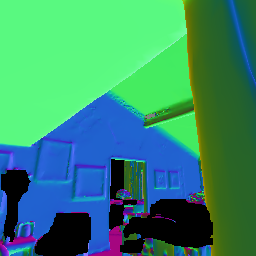} &  
\hspace{2.5mm} \includegraphics[width=0.14\linewidth]{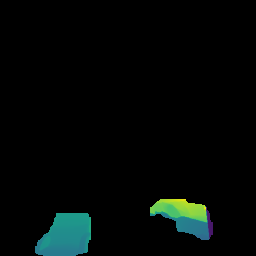} &
\hspace{2.5mm} \includegraphics[width=0.14\linewidth]{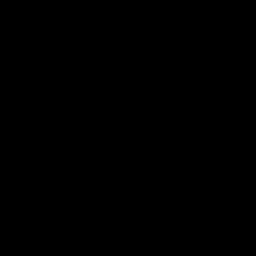} & 
\hspace{2.5mm} \includegraphics[width=0.14\linewidth]{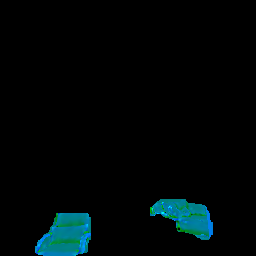} \\

\multirow{1}{*}[33pt]{{\rotatebox{90}{\makecell{30 \% \\ sparsity}}}}
\includegraphics[width=0.14\linewidth]{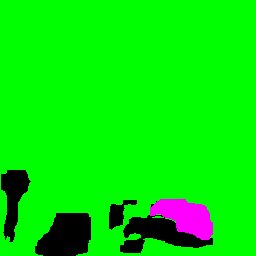} &
\hspace{2.5mm} \includegraphics[width=0.14\linewidth]{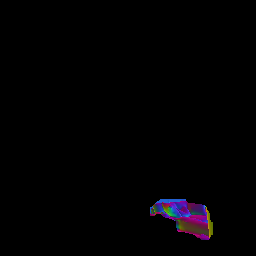} &  
\hspace{2.5mm} \includegraphics[width=0.14\linewidth]{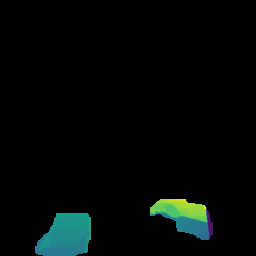} &
\hspace{2.5mm} \includegraphics[width=0.14\linewidth]{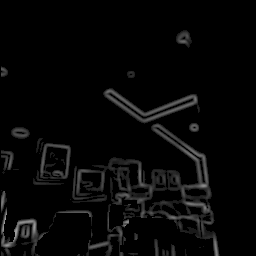} & 
\hspace{2.5mm} \includegraphics[width=0.14\linewidth]{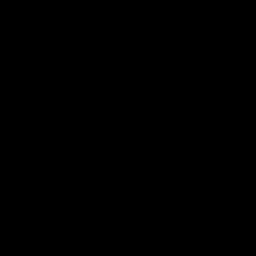} \\

\end{tabular}
	\caption{\textbf{One example of sparse input labels.} One sample from the Taskonomy dataset where input labels have the sparsity of 70\%, 50\% and 30\%.}
	\label{fig:sparse-lbl1}
\end{figure*}

\begin{figure*}[h]
\addtolength{\tabcolsep}{-8pt}  
\centering
\begin{tabular}{c}
    Image   \\
    \includegraphics[width=0.14\linewidth]{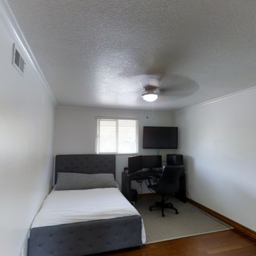}  
    \\
\end{tabular}
\\

\begin{tabular} {ccccc}
 Semseg & Normals & Depth & Edges & Curvature\\

\multirow{1}{*}[33pt]{{\rotatebox{90}{\makecell{70 \% \\ sparsity}}}}
\includegraphics[width=0.14\linewidth]{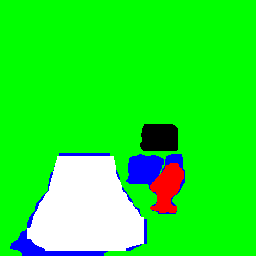} &
\hspace{2.5mm} \includegraphics[width=0.14\linewidth]{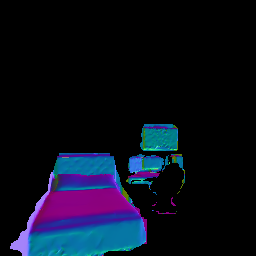} &  
\hspace{2.5mm} \includegraphics[width=0.14\linewidth]{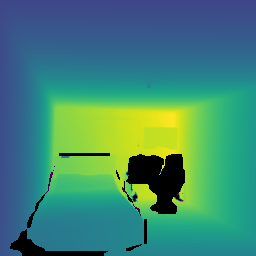} &
\hspace{2.5mm} \includegraphics[width=0.14\linewidth]{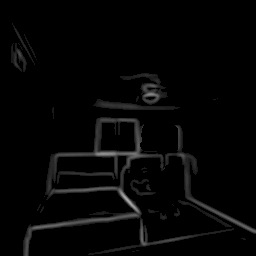} & 
\hspace{2.5mm} \includegraphics[width=0.14\linewidth]{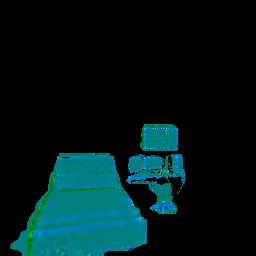} \\

\multirow{1}{*}[33pt]{{\rotatebox{90}{\makecell{50 \% \\ sparsity}}}}
\includegraphics[width=0.14\linewidth]{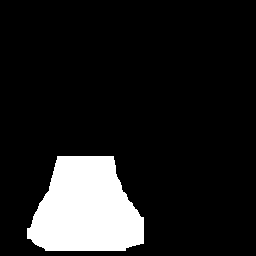} &
\hspace{2.5mm} \includegraphics[width=0.14\linewidth]{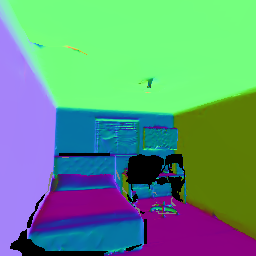} &  
\hspace{2.5mm} \includegraphics[width=0.14\linewidth]{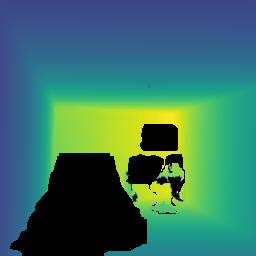} &
\hspace{2.5mm} \includegraphics[width=0.14\linewidth]{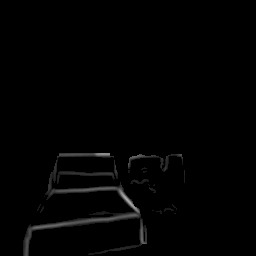} & 
\hspace{2.5mm} \includegraphics[width=0.14\linewidth]{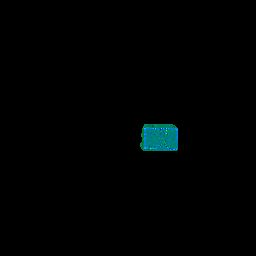} \\

\multirow{1}{*}[33pt]{{\rotatebox{90}{\makecell{30 \% \\ sparsity}}}}
\includegraphics[width=0.14\linewidth]{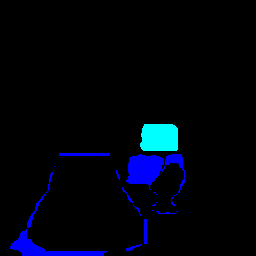} &
\hspace{2.5mm} \includegraphics[width=0.14\linewidth]{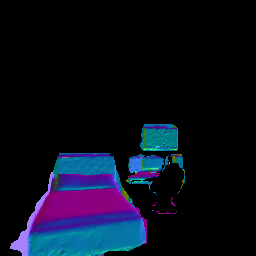} &  
\hspace{2.5mm} \includegraphics[width=0.14\linewidth]{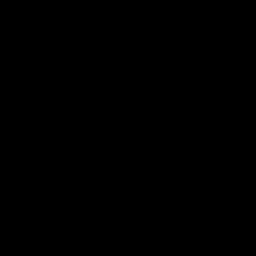} &
\hspace{2.5mm} \includegraphics[width=0.14\linewidth]{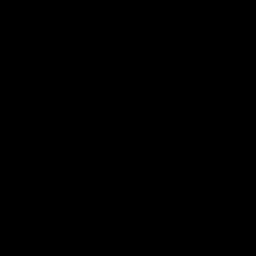} & 
\hspace{2.5mm} \includegraphics[width=0.14\linewidth]{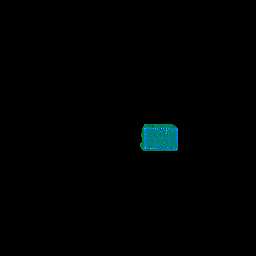} \\

\end{tabular}
	\caption{\textbf{One example of sparse input labels.} One sample from the Taskonomy dataset where input labels have the sparsity of 70\%, 50\% and 30\%.}
	\label{fig:sparse-lbl2}
\end{figure*}

\begin{figure*}[h]
\addtolength{\tabcolsep}{-8pt}  
\centering
\begin{tabular}{c}
    Image   \\
    \includegraphics[width=0.14\linewidth]{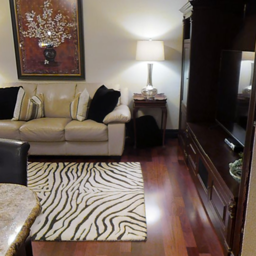}
    \\
\end{tabular}
\\

\begin{tabular} {ccccc}
Semseg & Normals & Depth & Edges & Curvature\\

\multirow{1}{*}[33pt]{{\rotatebox{90}{\makecell{70 \% \\ sparsity}}}}
\includegraphics[width=0.14\linewidth]{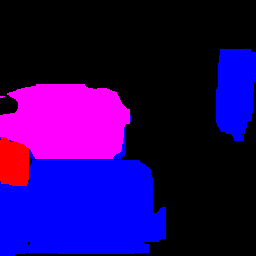} &
\hspace{2.5mm} \includegraphics[width=0.14\linewidth]{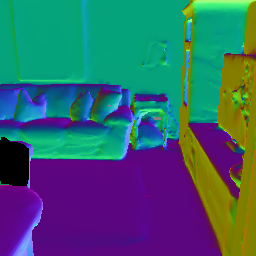} & 
\hspace{2.5mm} \includegraphics[width=0.14\linewidth]{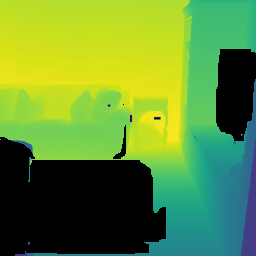} &
\hspace{2.5mm} \includegraphics[width=0.14\linewidth]{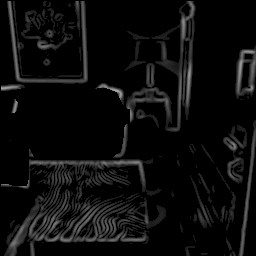} & 
\hspace{2.5mm} \includegraphics[width=0.14\linewidth]{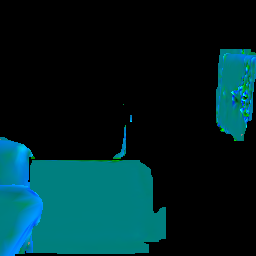} \\

\multirow{1}{*}[33pt]{{\rotatebox{90}{\makecell{50 \% \\ sparsity}}}}
\includegraphics[width=0.14\linewidth]{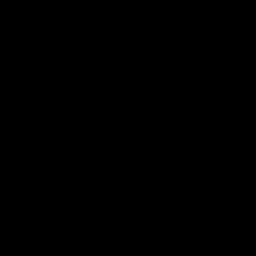} &
\hspace{2.5mm} \includegraphics[width=0.14\linewidth]{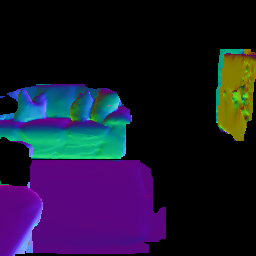} &  
\hspace{2.5mm} \includegraphics[width=0.14\linewidth]{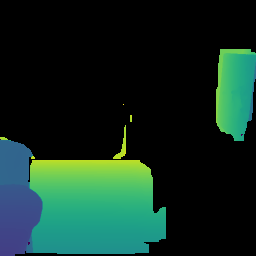} &
\hspace{2.5mm} \includegraphics[width=0.14\linewidth]{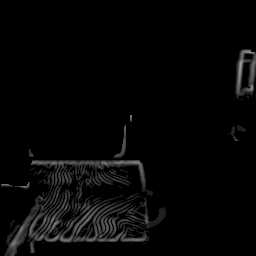} & 
\hspace{2.5mm} \includegraphics[width=0.14\linewidth]{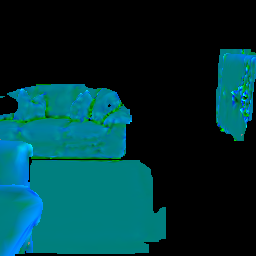} \\

\multirow{1}{*}[33pt]{{\rotatebox{90}{\makecell{30 \% \\ sparsity}}}}
\includegraphics[width=0.14\linewidth]{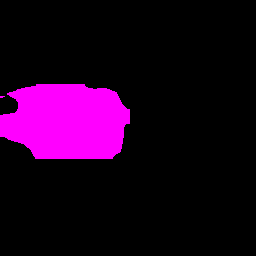} &
\hspace{2.5mm} \includegraphics[width=0.14\linewidth]{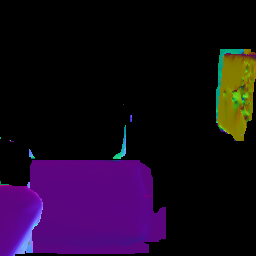} &  
\hspace{2.5mm} \includegraphics[width=0.14\linewidth]{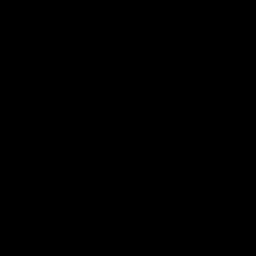} &
\hspace{2.5mm} \includegraphics[width=0.14\linewidth]{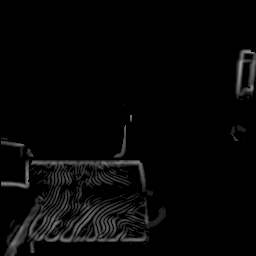} & 
\hspace{2.5mm} \includegraphics[width=0.14\linewidth]{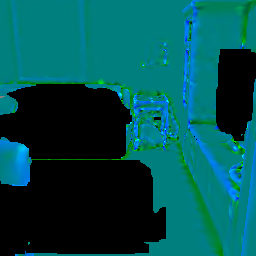} \\

\end{tabular}
	\caption{\textbf{One example of sparse input labels.} One sample from the Taskonomy dataset where input labels have the sparsity of 70\%, 50\% and 30\%.}
	\label{fig:sparse-lbl3}
\end{figure*}

%% file: figures/supplementary/big_diagram.tex
\begin{figure*}[h]
    \centering
    \includegraphics[width=\linewidth]{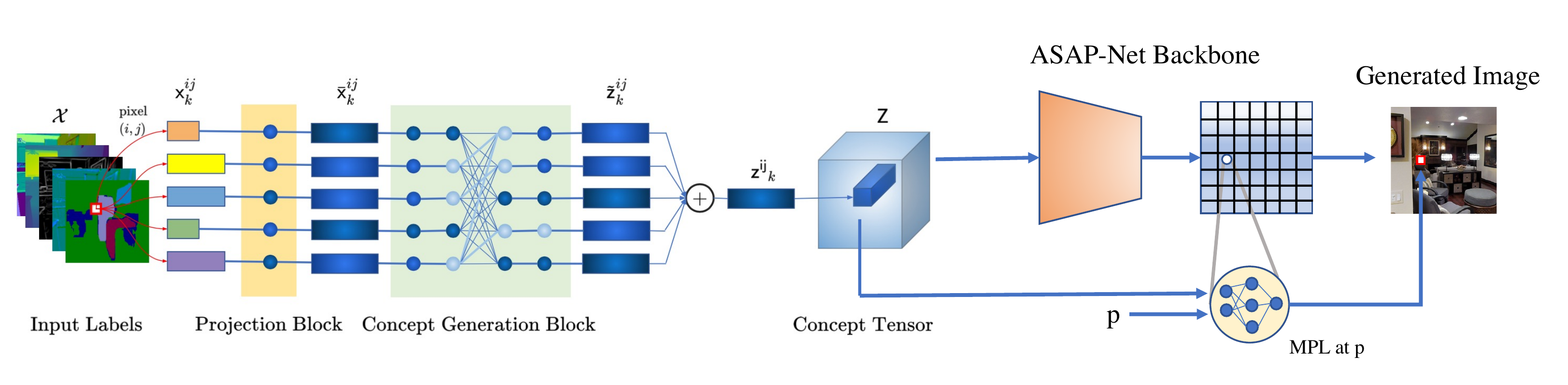}
    \vspace{-12pt}
    \caption{\textbf{Label Merging TLAM block and ASAP-Net Backbone.} The input labels are processed by the label merging network to output the \emph{Concept Tensor}. The ASAP generator takes the \emph{Concept Tensor} $\mathsf{Z} \in \mathbb{R}^{H \times W \times d}$ as an input and outputs a tensor of weights. Those weights are parameters of pixelwise, spatially-varying, MLPs, which compute the final output image from the \emph{Concept Tensor} $\mathsf{Z}$.}
    \label{fig:network-overview}
\end{figure*}

%% file: figures/supplementary/tlam-all-lbl.tex
\begin{figure*}[h]
\addtolength{\tabcolsep}{-8pt} 
\centering
\begin{tabular} {cccc}
 
\includegraphics[width=0.24\textwidth]{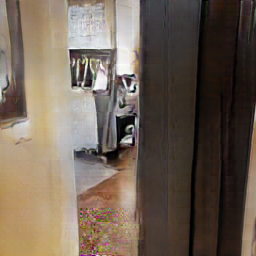}&
\hspace{2mm} \includegraphics[width=0.24\textwidth]{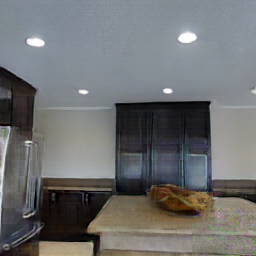} &
\hspace{2mm} \includegraphics[width=0.24\textwidth]{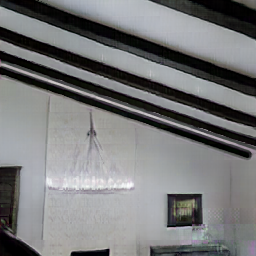} &   
\hspace{2mm} \includegraphics[width=0.24\textwidth]{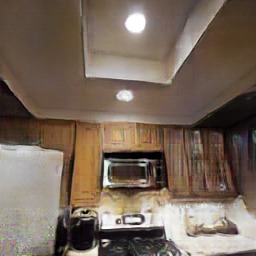} \\

\includegraphics[width=0.24\textwidth]{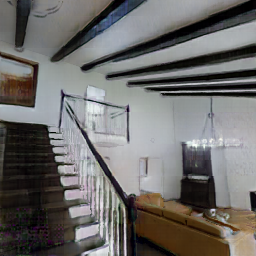} &  
\hspace{2mm} \includegraphics[width=0.24\textwidth]{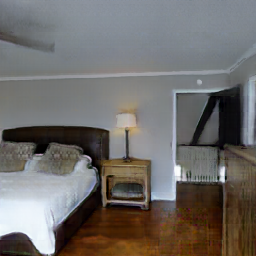} &
\hspace{2mm} \includegraphics[width=0.24\textwidth]{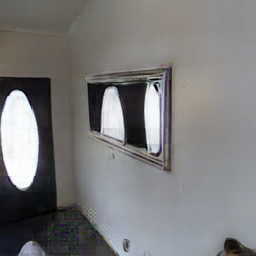}&
\hspace{2mm} \includegraphics[width=0.24\textwidth]{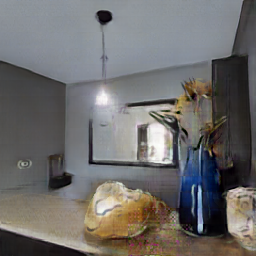} \\ 

\includegraphics[width=0.24\textwidth]{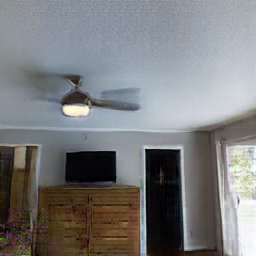} &  
\hspace{2mm} \includegraphics[width=0.24\textwidth]{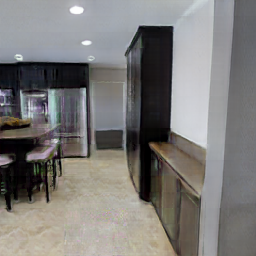} &
\hspace{2mm} \includegraphics[width=0.24\textwidth]{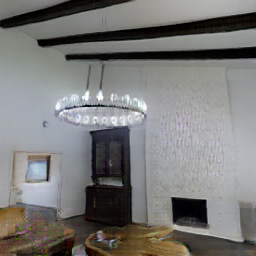}&
\hspace{2mm} \includegraphics[width=0.24\textwidth]{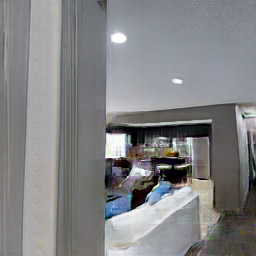} \\ 

\includegraphics[width=0.24\textwidth]{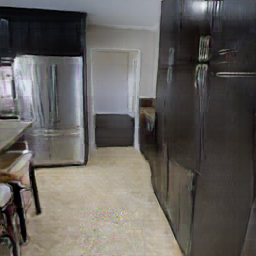} &  
\hspace{2mm} \includegraphics[width=0.24\textwidth]{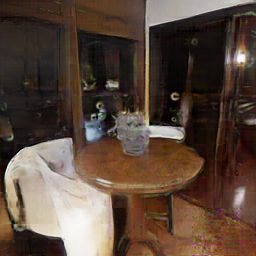} &
\hspace{2mm} \includegraphics[width=0.24\textwidth]{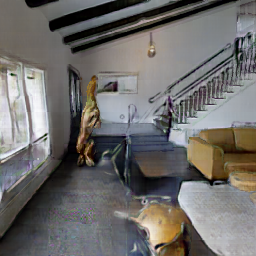}&
\hspace{2mm} \includegraphics[width=0.24\textwidth]{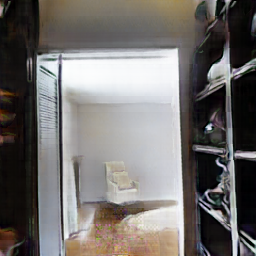} \\ 

\includegraphics[width=0.24\textwidth]{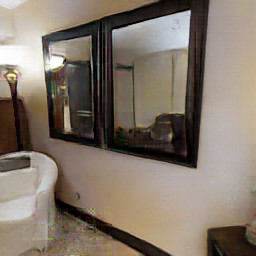} & 
\hspace{2mm} \includegraphics[width=0.24\textwidth]{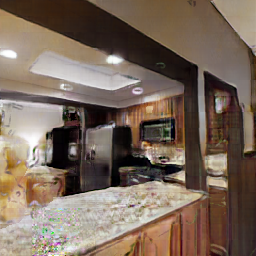} &
\hspace{2mm} \includegraphics[width=0.24\textwidth]{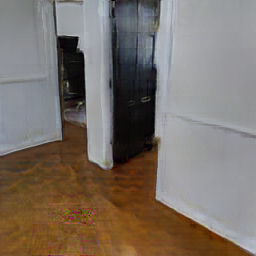}&
\hspace{2mm} \includegraphics[width=0.24\textwidth]{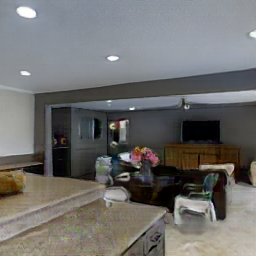} \\ 

\end{tabular}
	\caption{\textbf{Images generated using dense input labels.} Here we see images generated  with our proposed TLAM label merging model, using dense input labels from the Taskonomy dataset.}
	\label{fig:all-tlam}
\end{figure*}

%% file: figures/supplementary/tlam-sparse.tex
\begin{figure*}[h]
\addtolength{\tabcolsep}{-8pt}  
\centering
\begin{tabular} {cccc}
 
\includegraphics[width=0.24\textwidth]{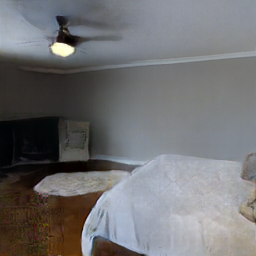}&
\hspace{2mm} \includegraphics[width=0.24\textwidth]{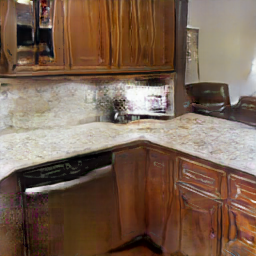} &
\hspace{2mm} \includegraphics[width=0.24\textwidth]{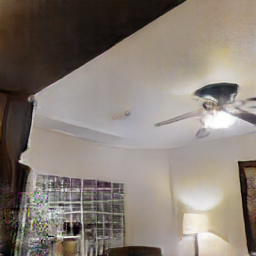} &  
\hspace{2mm} \includegraphics[width=0.24\textwidth]{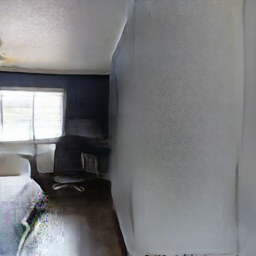} \\

\includegraphics[width=0.24\textwidth]{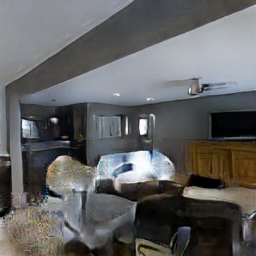} & 
\hspace{2mm} \includegraphics[width=0.24\textwidth]{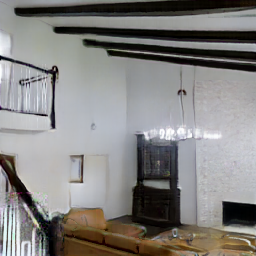} &
\hspace{2mm} \includegraphics[width=0.24\textwidth]{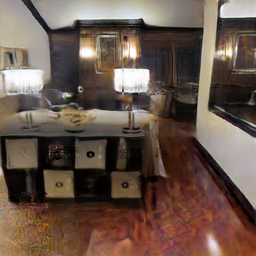}&
\hspace{2mm} \includegraphics[width=0.24\textwidth]{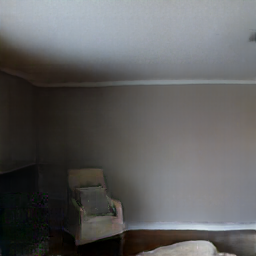} \\ 

\includegraphics[width=0.24\textwidth]{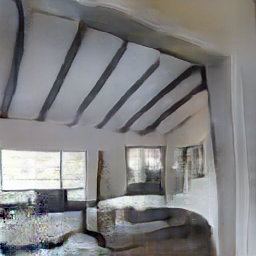} &  
\hspace{2mm} \includegraphics[width=0.24\textwidth]{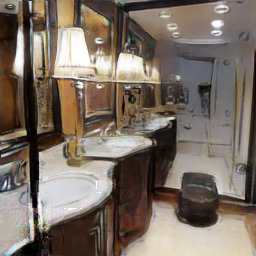} &
\hspace{2mm} \includegraphics[width=0.24\textwidth]{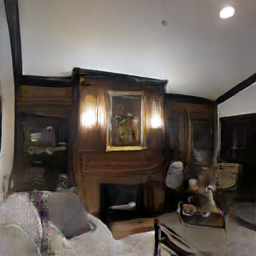}&
\hspace{2mm} \includegraphics[width=0.24\textwidth]{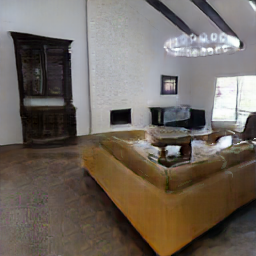} \\ 

\includegraphics[width=0.24\textwidth]{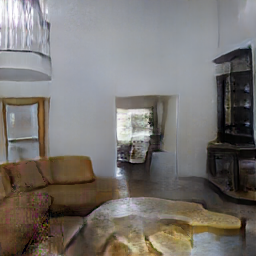} & 
\hspace{2mm} \includegraphics[width=0.24\textwidth]{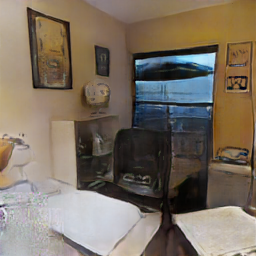} &
\hspace{2mm} \includegraphics[width=0.24\textwidth]{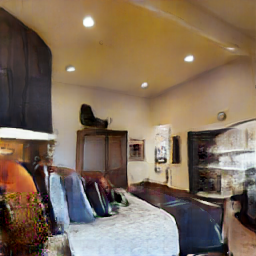}&
\hspace{2mm} \includegraphics[width=0.24\textwidth]{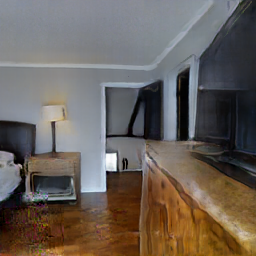} \\ 

\includegraphics[width=0.24\textwidth]{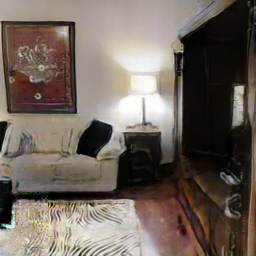} & 
\hspace{2mm} \includegraphics[width=0.24\textwidth]{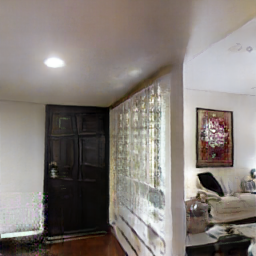} &
\hspace{2mm} \includegraphics[width=0.24\textwidth]{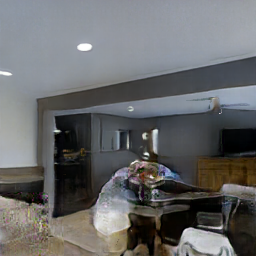}&
\hspace{2mm} \includegraphics[width=0.24\textwidth]{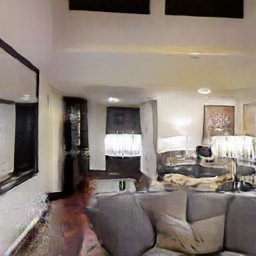} \\ 

\end{tabular}
	\caption{\textbf{Images generated using 50\% sparse input labels.} Here we see images generated  with our proposed TLAM label merging model, using input labels from the Taskonomy dataset with 50\% sparsity.}
	\label{fig:sparse-tlam}
\end{figure*}

%% file: figures/supplementary/image_editing.tex
\begin{figure*}[h]
    \centering
    \begin{tabular} {cccccc}
    \addtolength{\tabcolsep}{-4.5pt}  
\makecell{Original \\ image} & \makecell{Curvature \\ \emph{user-provided}}  & \makecell{Normals \\ \emph{mask by user}} & \makecell{SPADE \\ \cite{park2019semantic}} & \makecell{ASAP \\\cite{RottShaham2020ASAP}} & \makecell{TLAM \\ (Ours)} \\
     \includegraphics[width=0.14\linewidth]{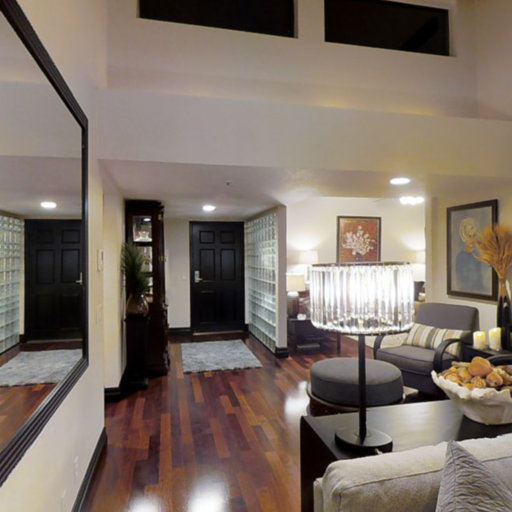}  &
    \includegraphics[width=0.14\linewidth]{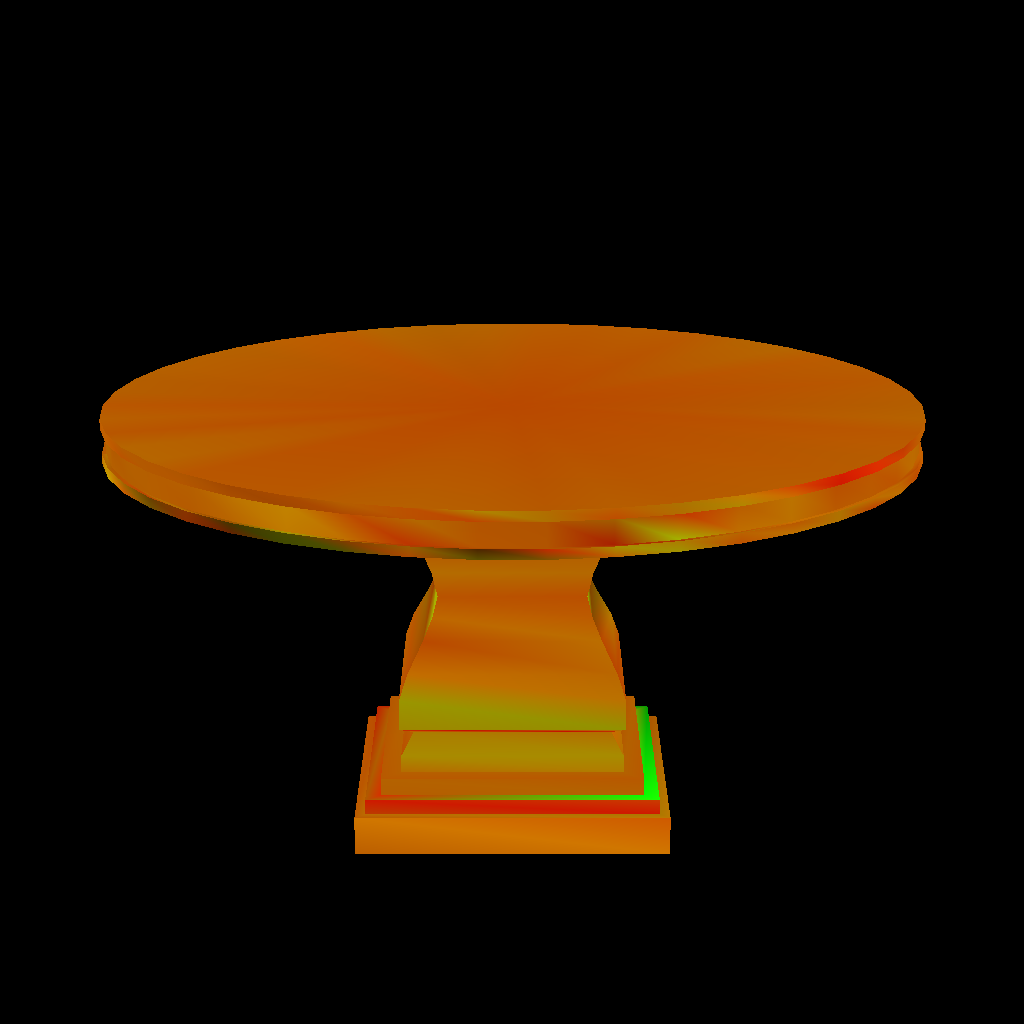}  &
    \includegraphics[width=0.14\linewidth]{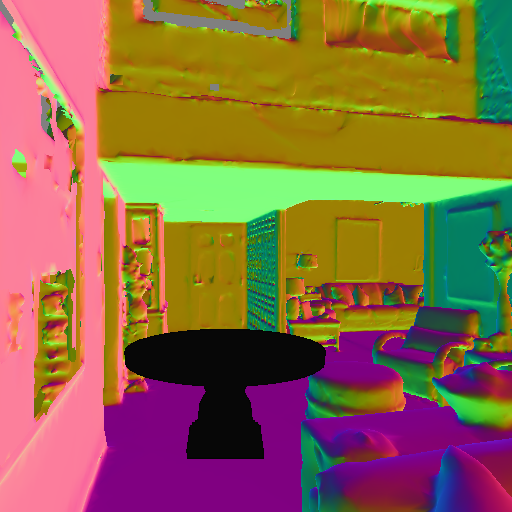}  &
    \includegraphics[width=0.14\linewidth]{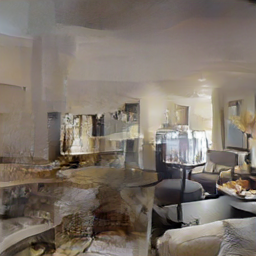} &
    \includegraphics[width=0.14\linewidth]{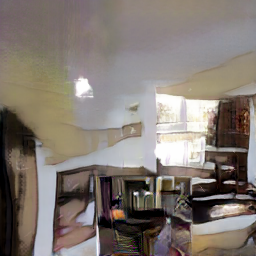} &
    \includegraphics[width=0.14\linewidth]{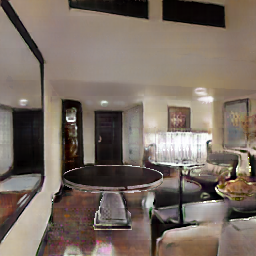} 
    \end{tabular}
	\caption{\textbf{Geometric image editing with user inputs.}
	From left to right: original image; curvature label (provided among five) of the table to be inserted into the image;  semantics of the table on the normal map of the scene; generated image using SPADE, ASAP and our method, respectively. During editing, we use dense labels of the original image, where we introduce five labels derived from a texture-less 3D model of a table (object of interest)-- labels are first edited to introduce the object, followed by the image generation using the proposed TLAM method.}
	\label{tab::geom_edit}
\end{figure*}

\begin{figure*}[t!]
    \addtolength{\tabcolsep}{-4.7pt}
    \centering
    \begin{tabular} {ccccc}
    Original & Mask & SPADE~\cite{park2019semantic} & ASAP~\cite{RottShaham2020ASAP}  & TLAM (Ours)  \\
    \includegraphics[width=0.14\linewidth]{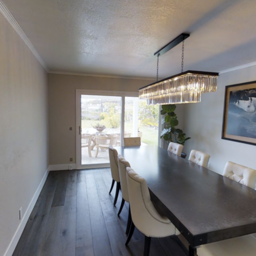} &
    \includegraphics[width=0.14\linewidth]{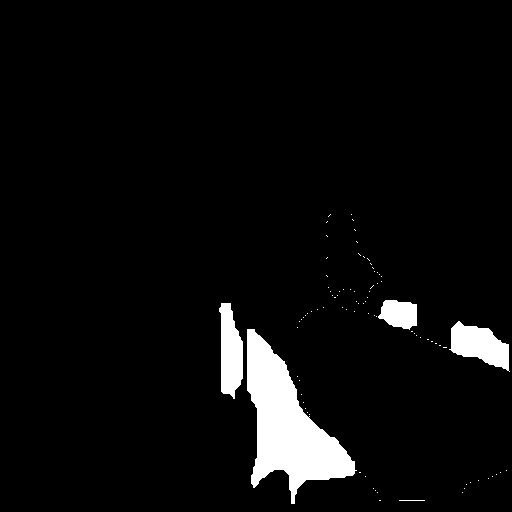} &
    \includegraphics[width=0.14\linewidth]{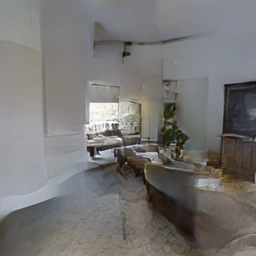} &
    \includegraphics[width=0.14\linewidth]{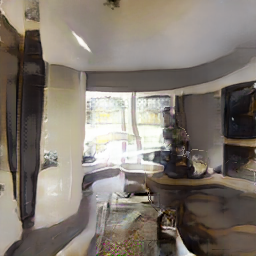} &
    \includegraphics[width=0.14\linewidth]{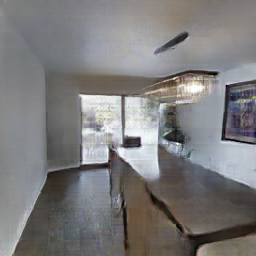}
    \end{tabular}
	\caption{\textbf{Object  removal.} 
	From left to right: original image; mask of the removed object; generated image using SPADE, ASAP and our method, respectively.
	This figure shows removal of chairs from the image, by removing their labels.
	The mask for removal is \emph{manually} chosen by the user. 
	}
	\label{fig:object_removal_insertion}
\end{figure*}

%% file: figures/supplementary/cityscapes.tex
\begin{figure*}[h]
\addtolength{\tabcolsep}{-6pt}  
\centering
\begin{tabular} {cccccc}
 Label & Image & \makecell{SPADE \\ \cite{park2019semantic}} & \makecell{ASAP \\ \cite{RottShaham2020ASAP}} & \makecell{TLAM \\(Ours)} & \makecell{Sparse-TLAM \\ (Ours)}\\
 
\includegraphics[width=0.15\linewidth]{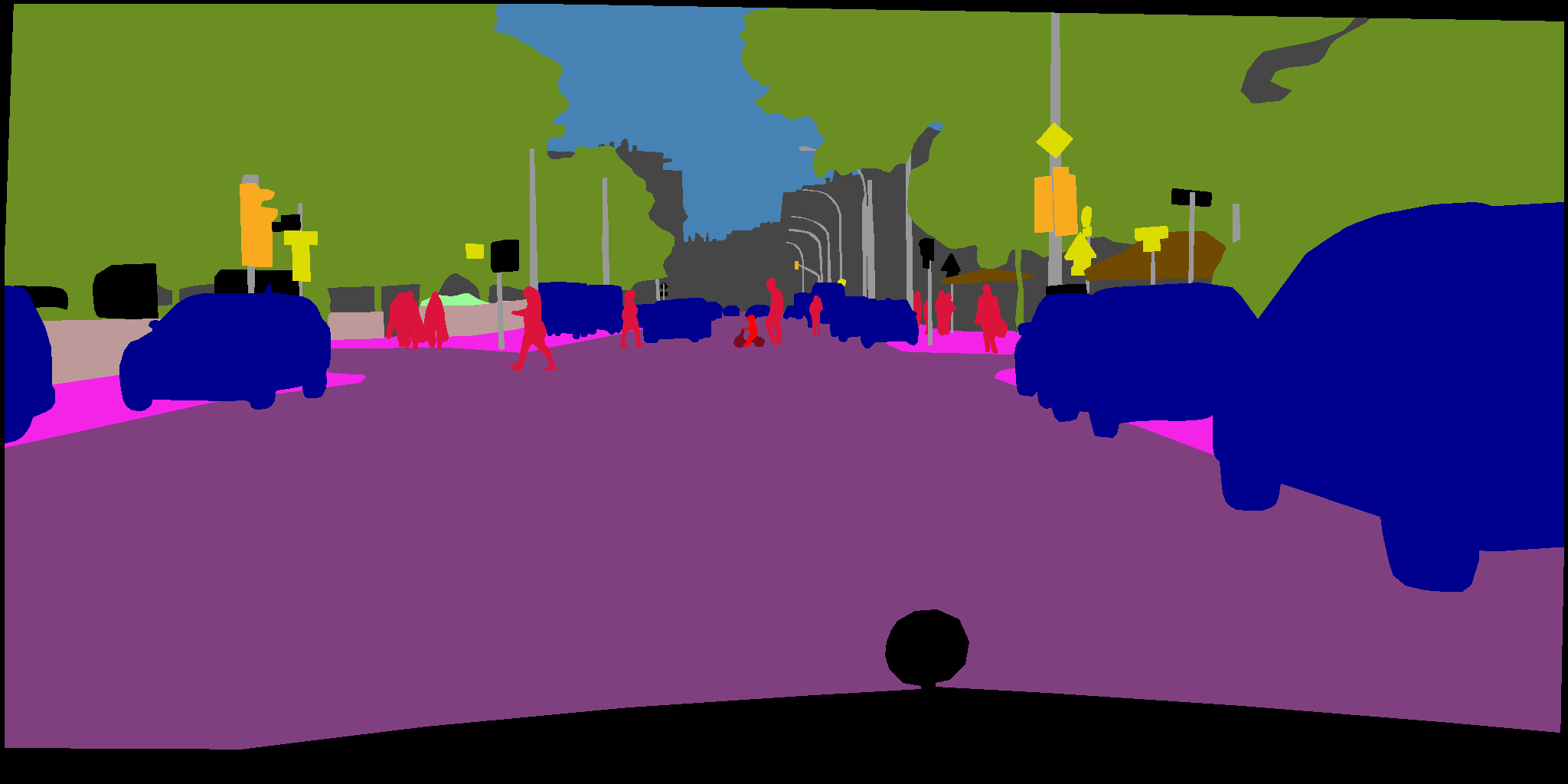}&
\hspace{2mm} \includegraphics[width=0.15\linewidth]{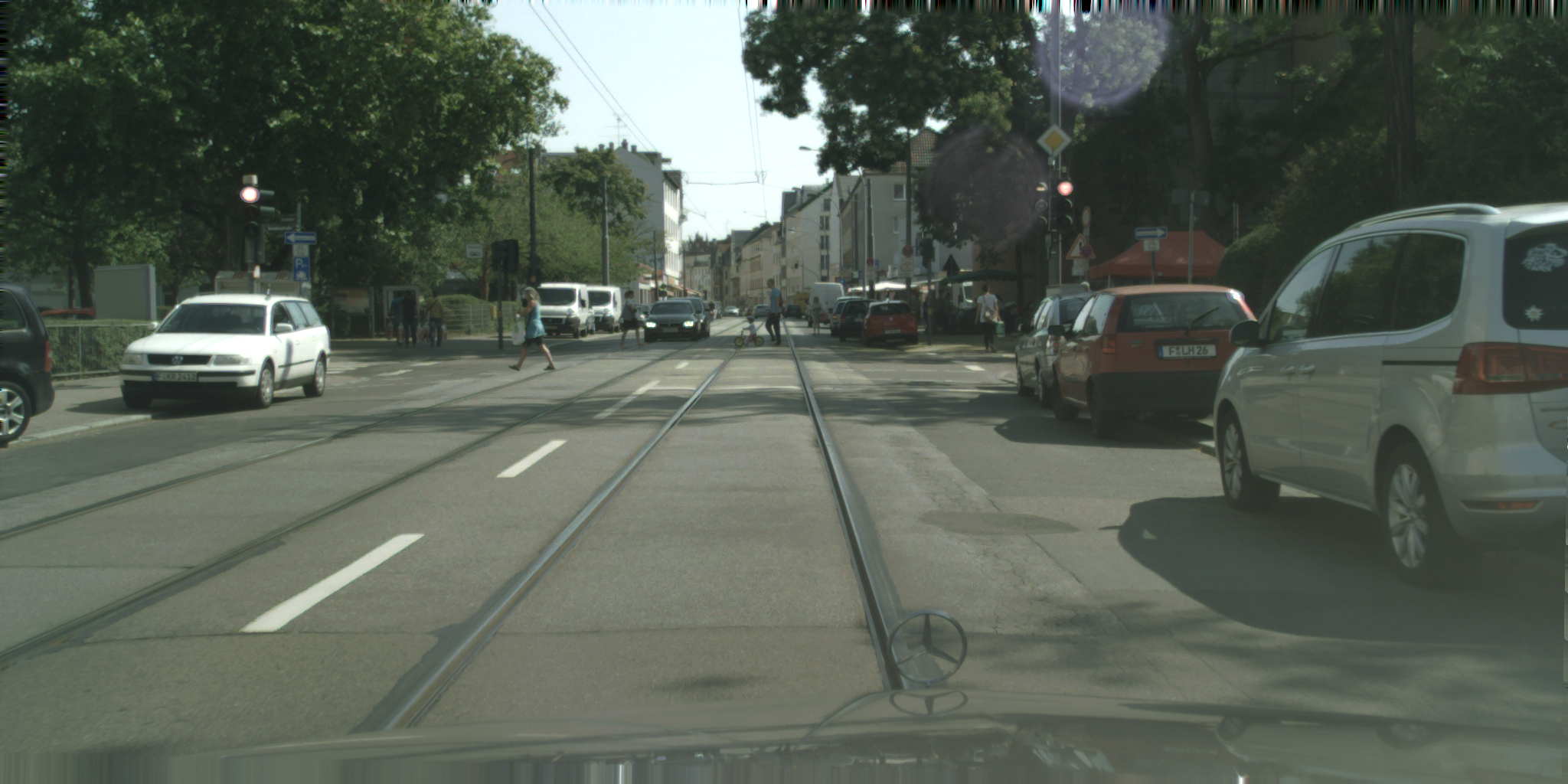} &
\hspace{2mm} \includegraphics[width=0.15\linewidth]{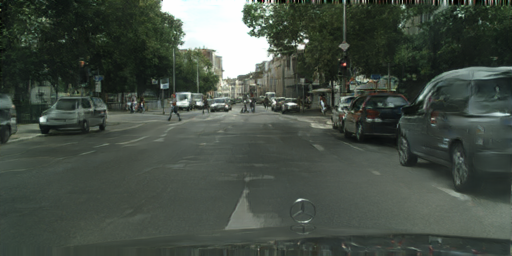} &   \hspace{2mm} \includegraphics[width=0.15\linewidth]{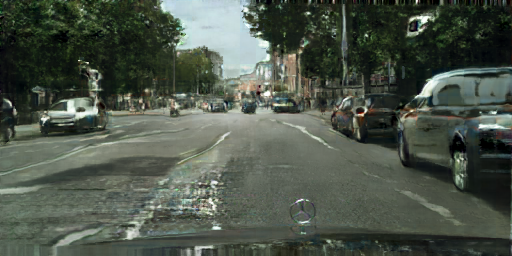} &
\hspace{2mm} \includegraphics[width=0.15\linewidth]{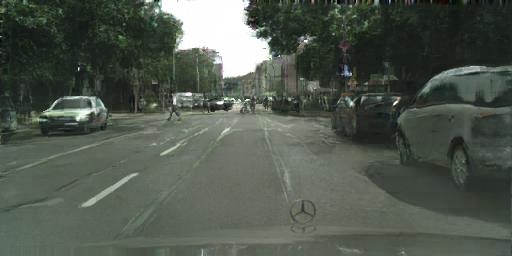} &  
\hspace{2mm} \includegraphics[width=0.15\linewidth]{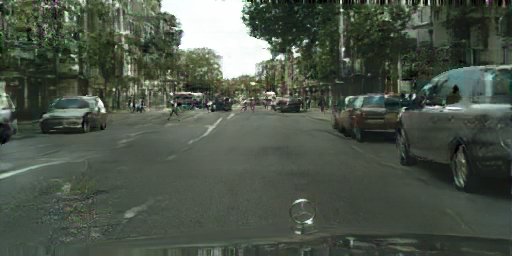} \\

\includegraphics[width=0.15\linewidth]{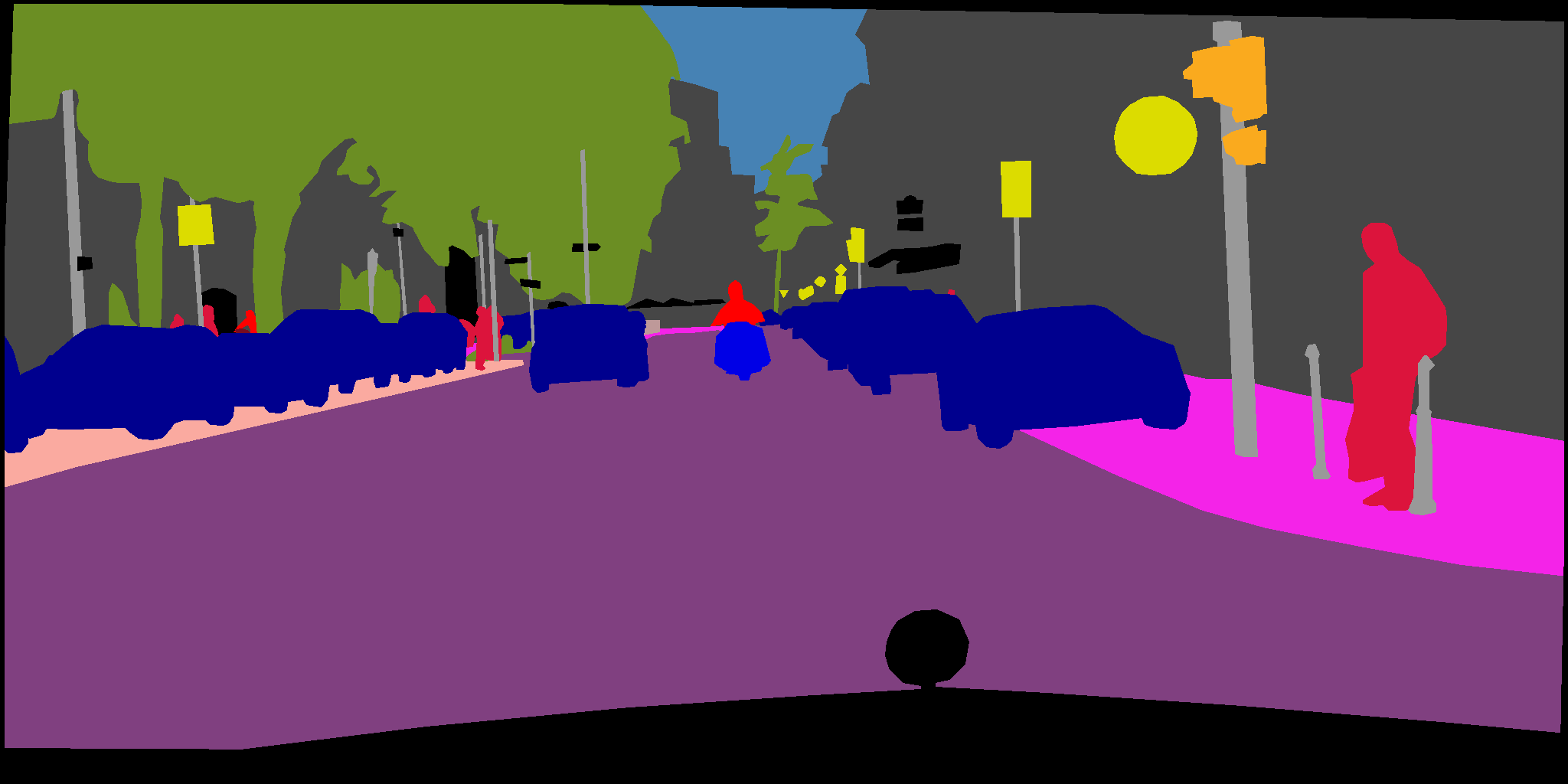}&
\hspace{2mm} \includegraphics[width=0.15\linewidth]{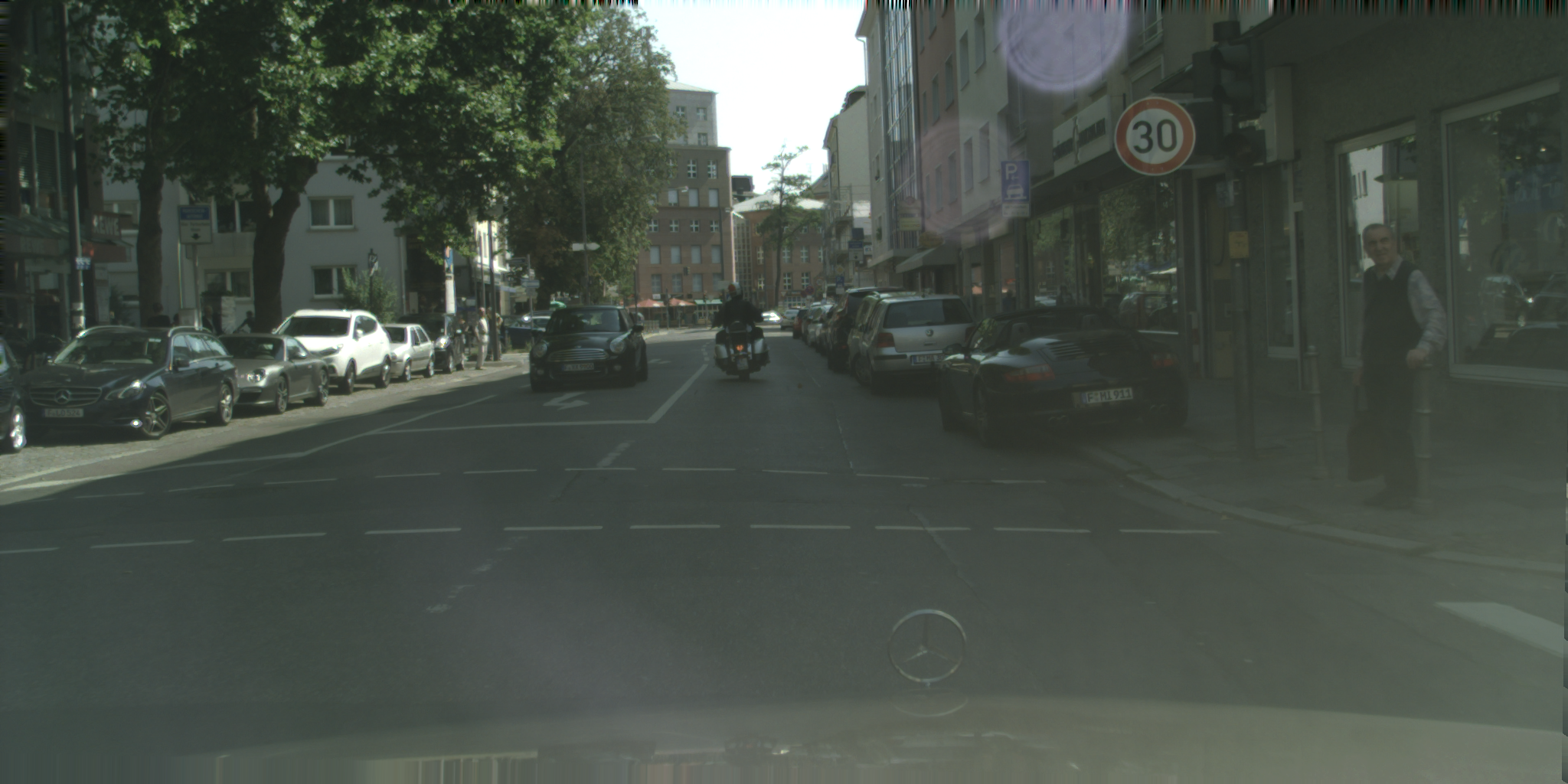} &
\hspace{2mm} \includegraphics[width=0.15\linewidth]{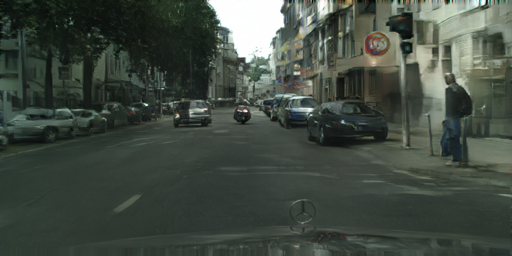} &   \hspace{2mm} \includegraphics[width=0.15\linewidth]{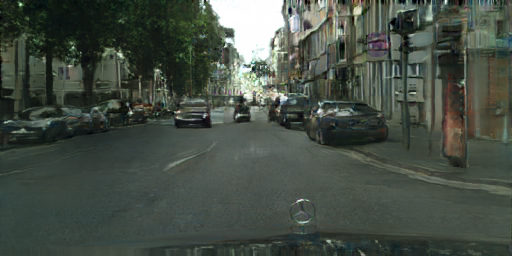} &
\hspace{2mm} \includegraphics[width=0.15\linewidth]{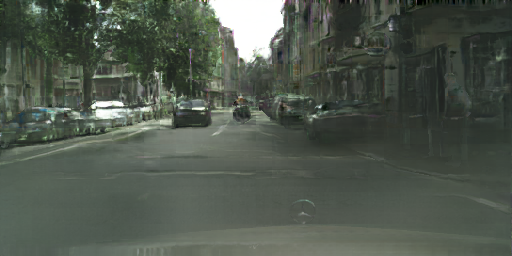} &  
\hspace{2mm} \includegraphics[width=0.15\linewidth]{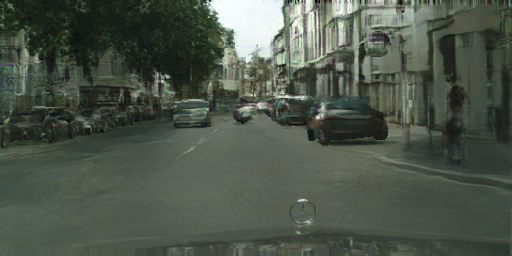} \\

\includegraphics[width=0.15\linewidth]{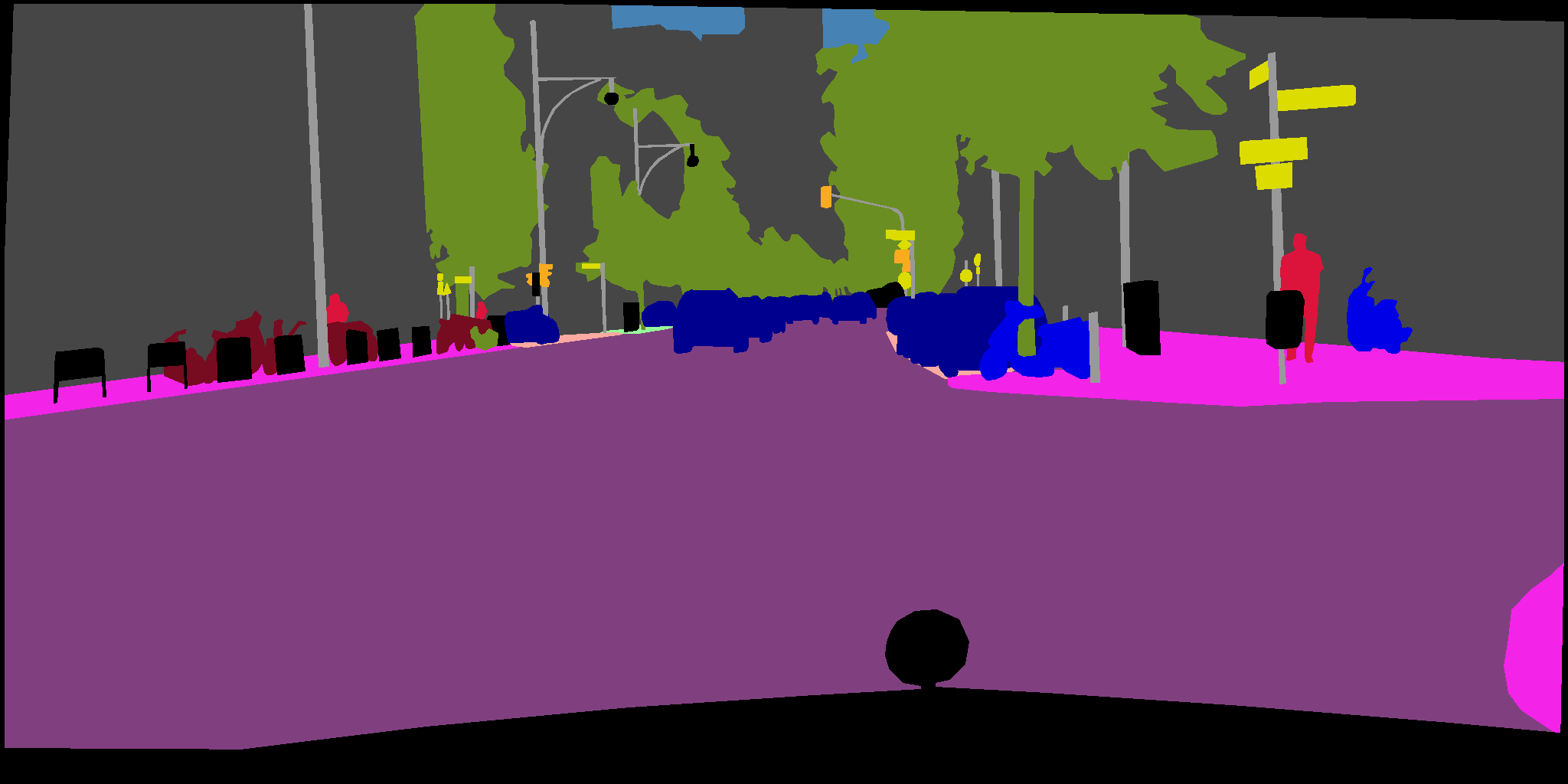}&
\hspace{2mm} \includegraphics[width=0.15\linewidth]{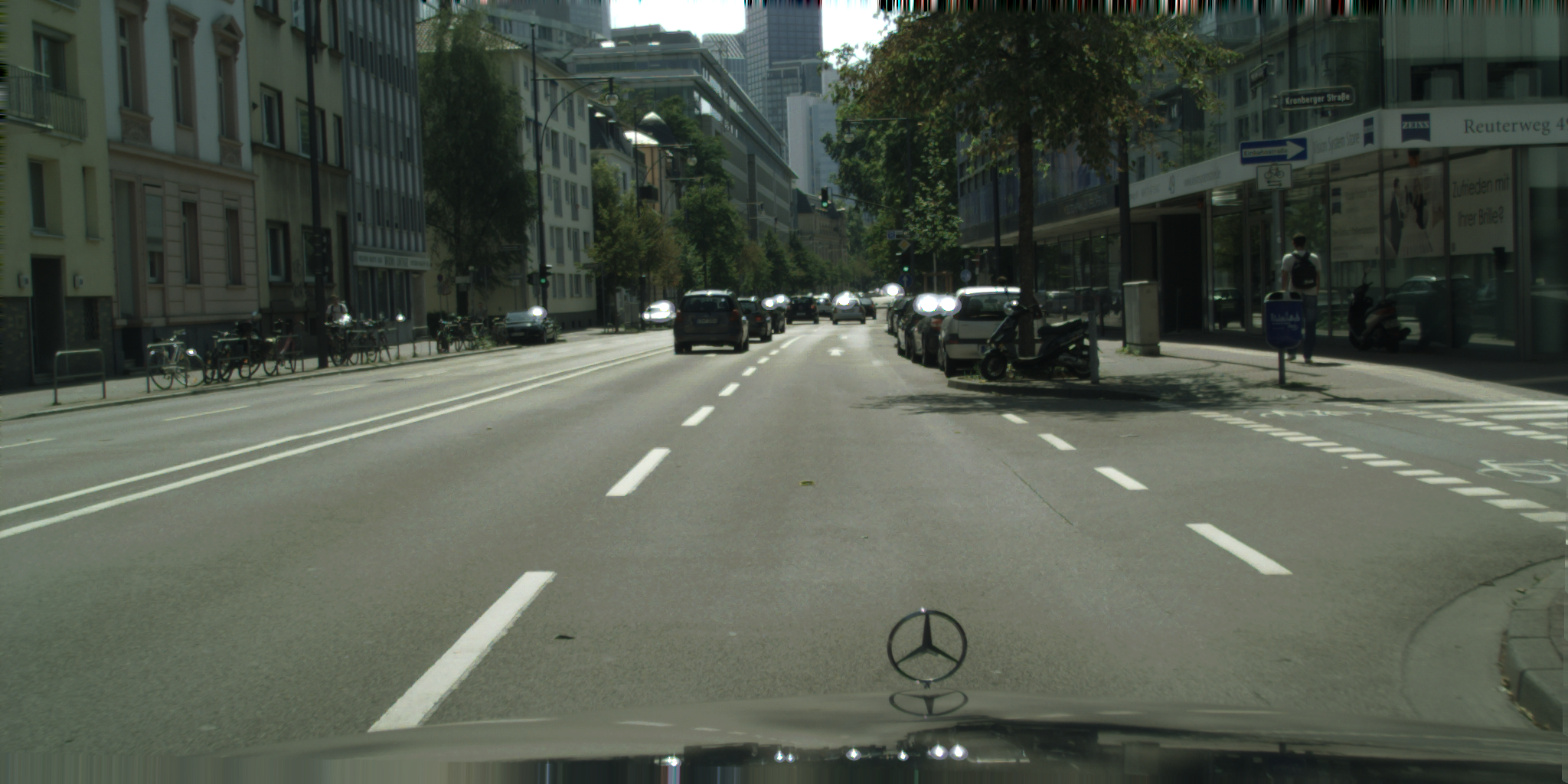} &
\hspace{2mm} \includegraphics[width=0.15\linewidth]{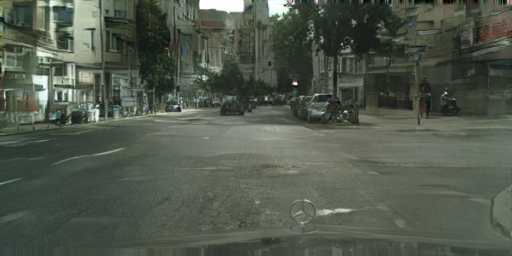} &    \hspace{2mm} \includegraphics[width=0.15\linewidth]{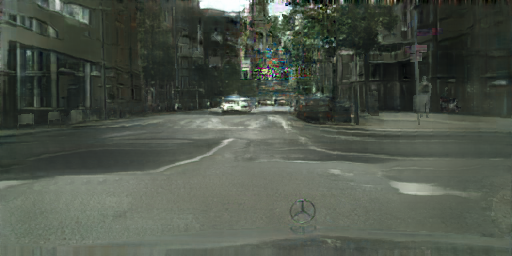} &
\hspace{2mm} \includegraphics[width=0.15\linewidth]{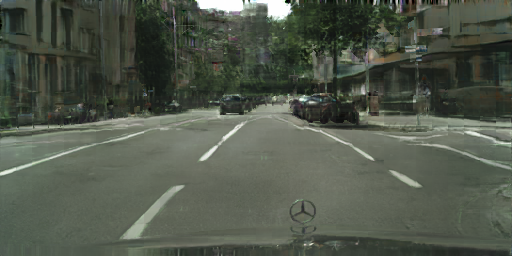} &  
\hspace{2mm} \includegraphics[width=0.15\linewidth]{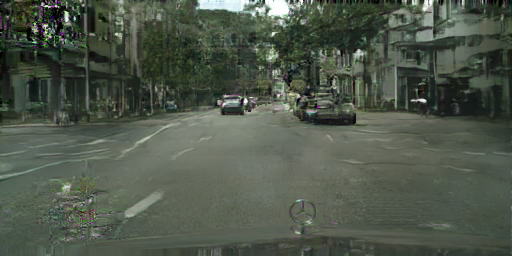} \\

\includegraphics[width=0.15\linewidth]{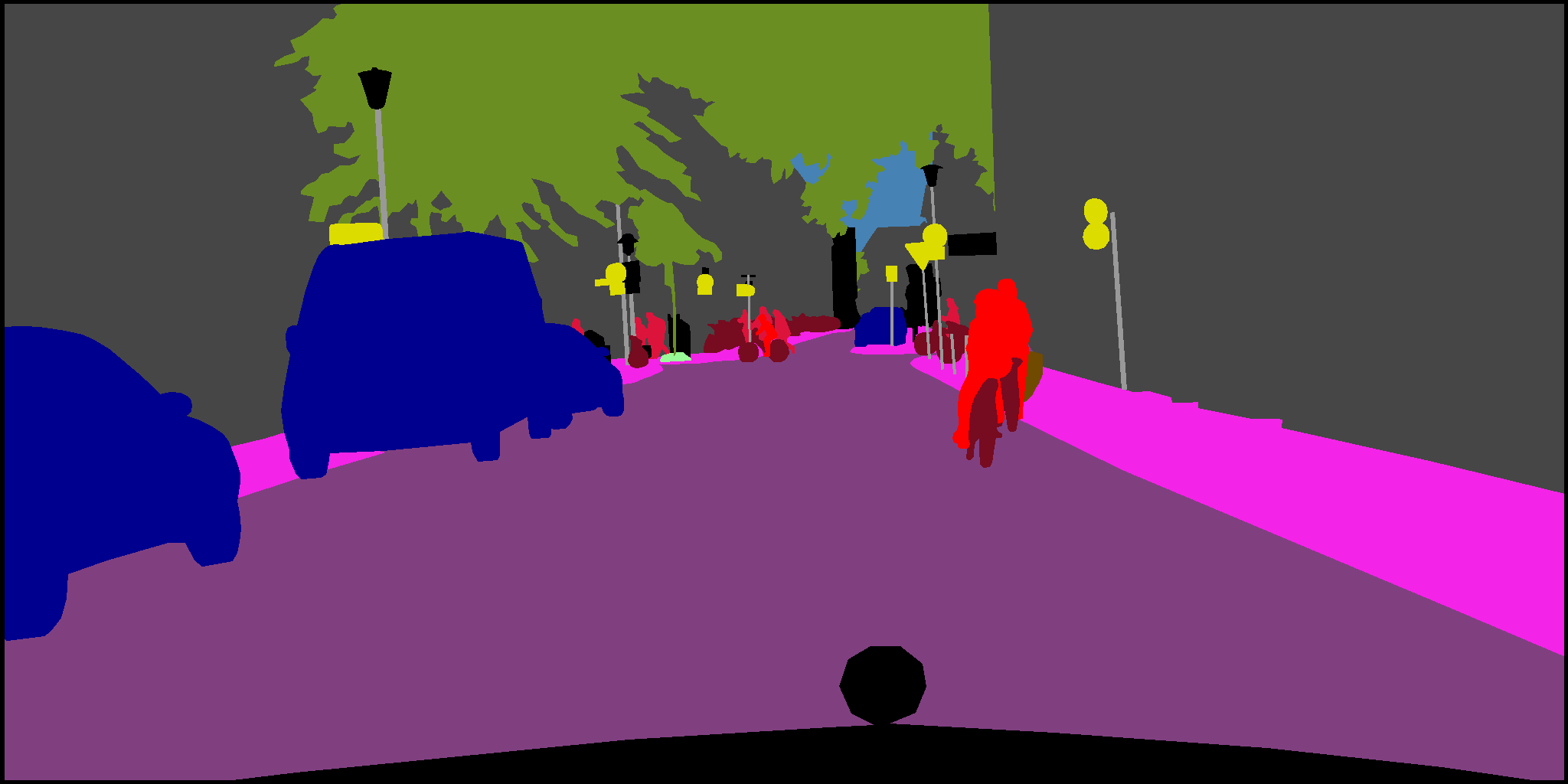}&
\hspace{2mm} \includegraphics[width=0.15\linewidth]{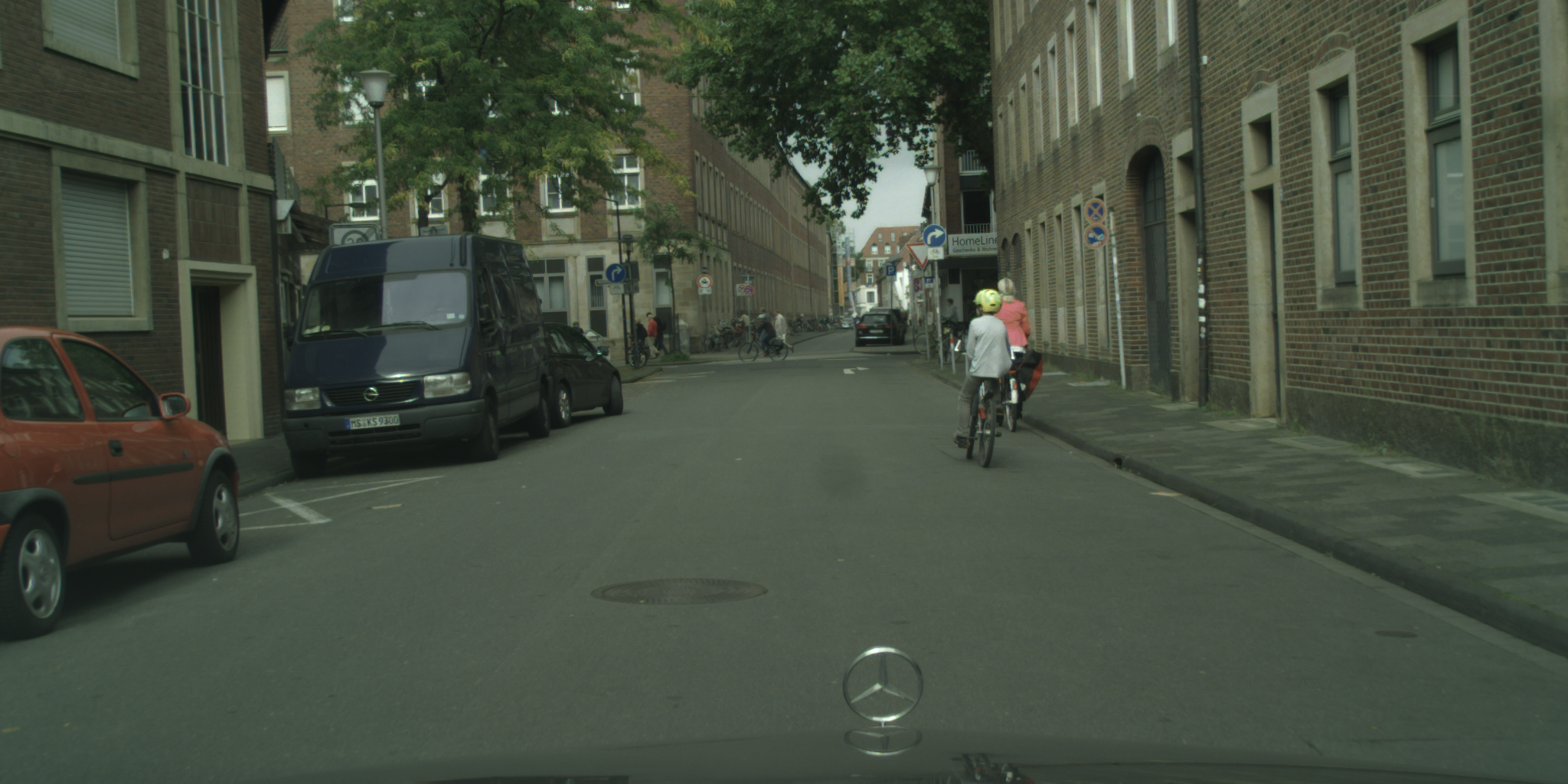} &
\hspace{2mm} \includegraphics[width=0.15\linewidth]{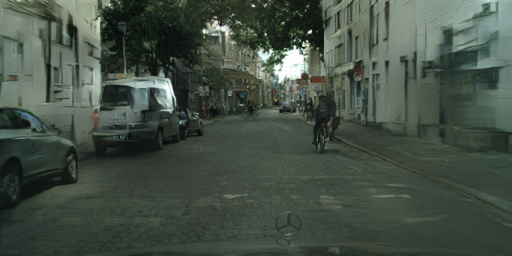} &   \hspace{2mm} \includegraphics[width=0.15\linewidth]{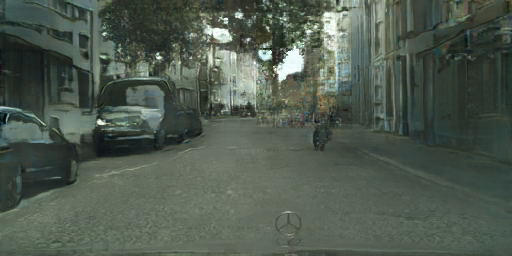} &
\hspace{2mm} \includegraphics[width=0.15\linewidth]{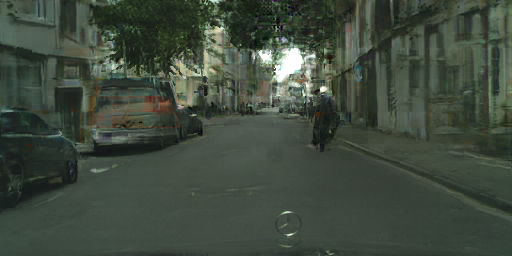} &  
\hspace{2mm} \includegraphics[width=0.15\linewidth]{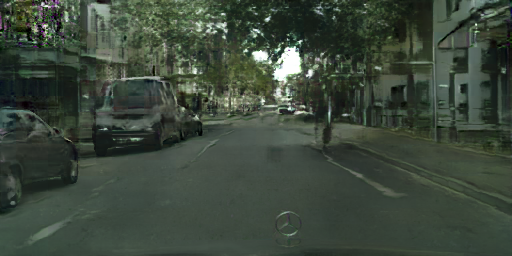} \\

\includegraphics[width=0.15\linewidth]{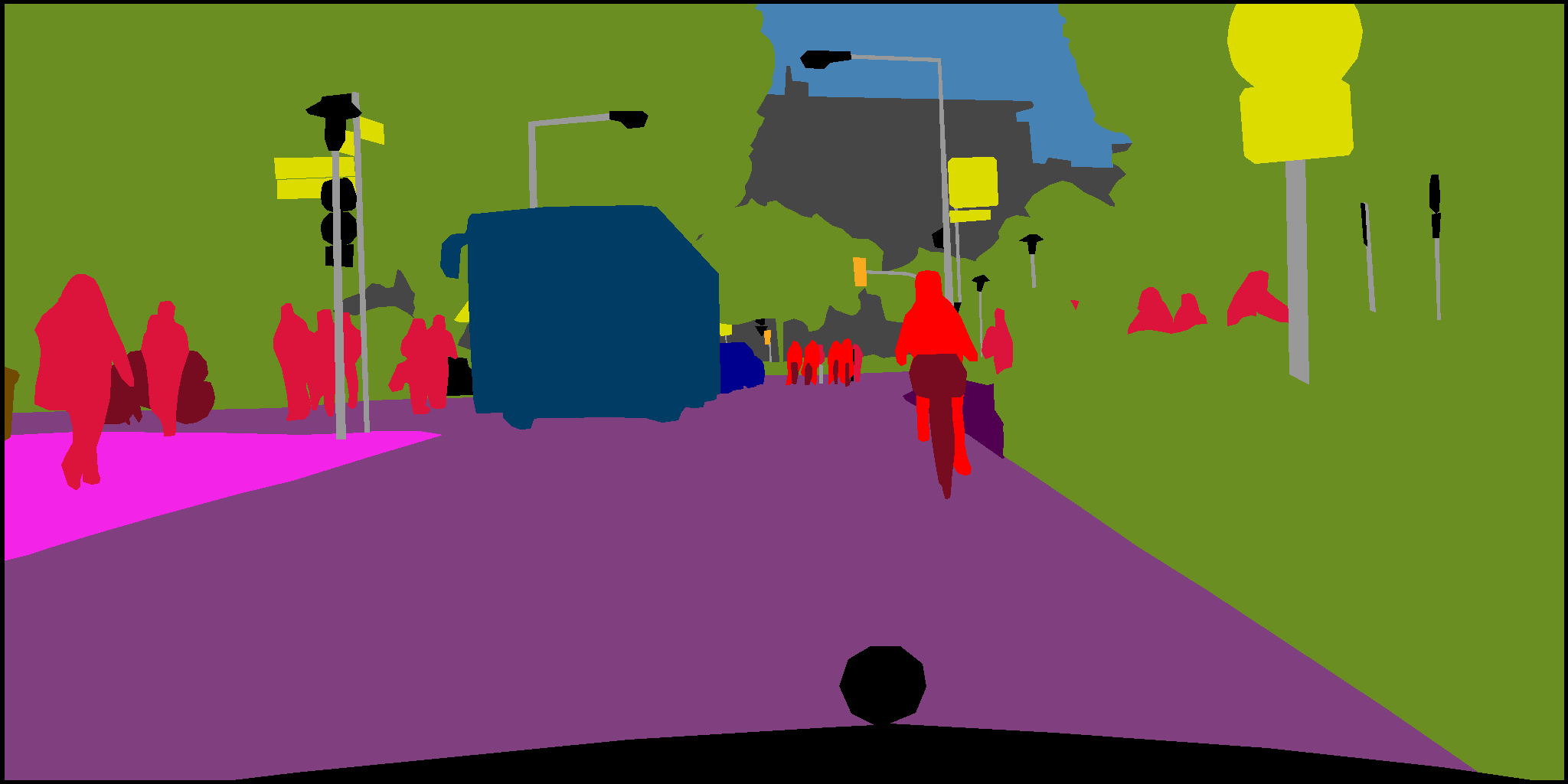}&
\hspace{2mm} \includegraphics[width=0.15\linewidth]{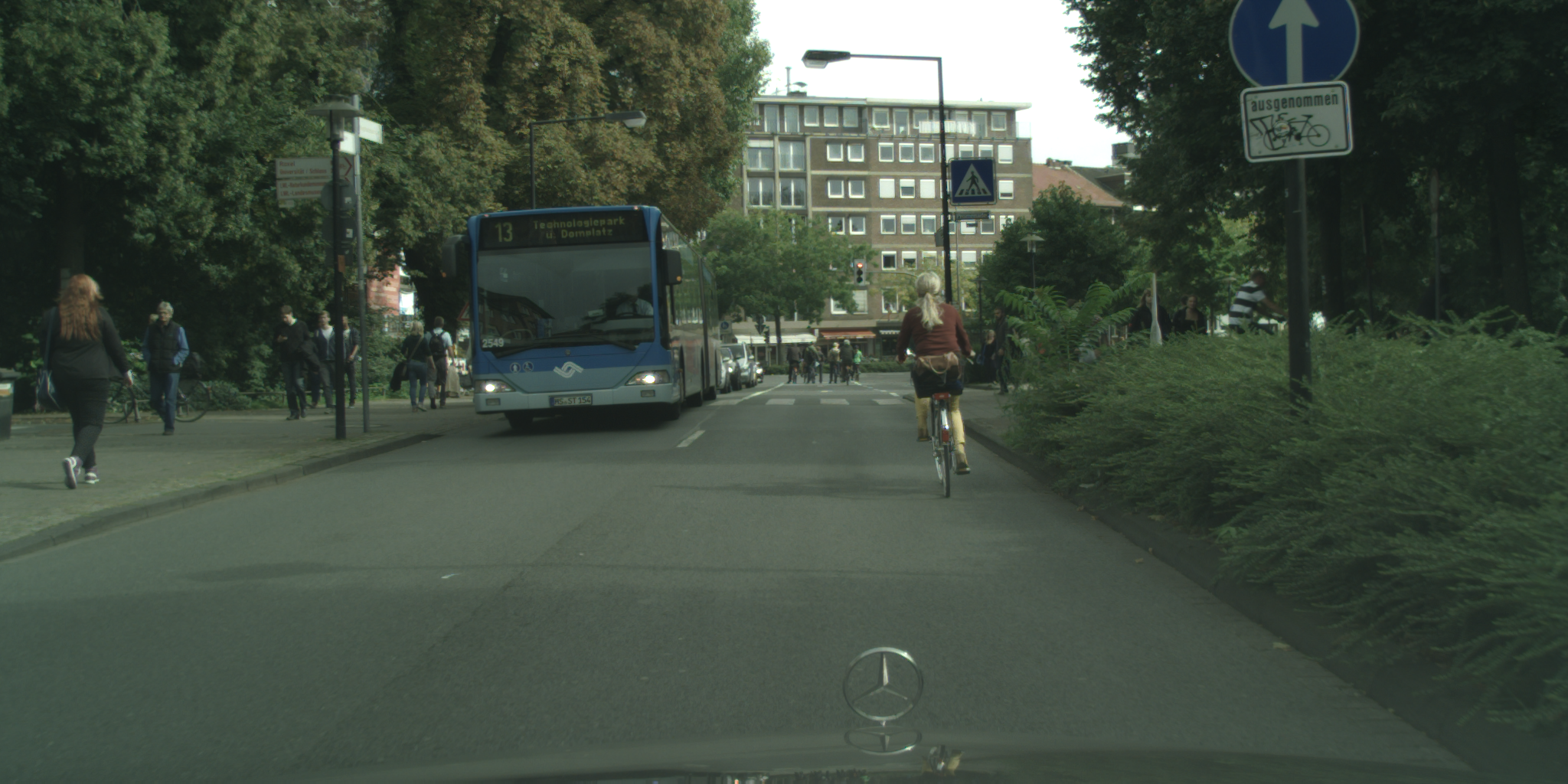} &
\hspace{2mm} \includegraphics[width=0.15\linewidth]{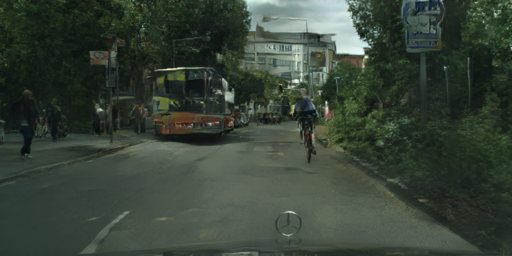} &   \hspace{2mm} \includegraphics[width=0.15\linewidth]{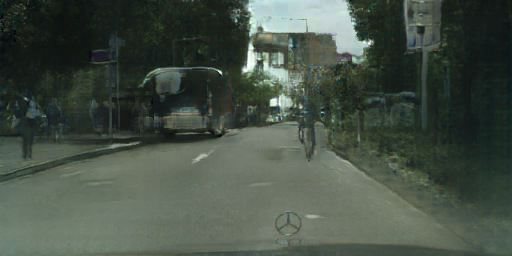} &
\hspace{2mm} \includegraphics[width=0.15\linewidth]{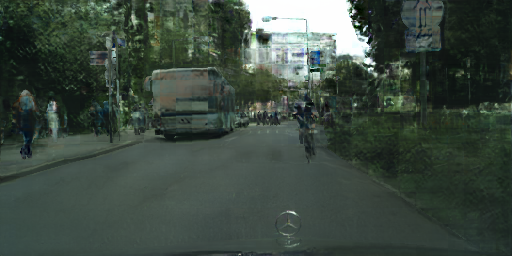} &  
\hspace{2mm} \includegraphics[width=0.15\linewidth]{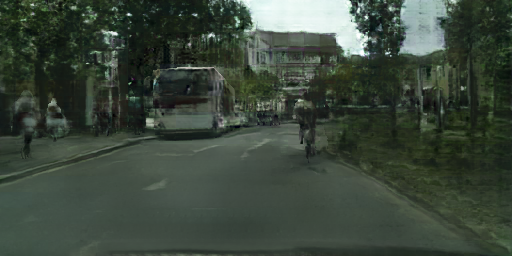} \\

\includegraphics[width=0.15\linewidth]{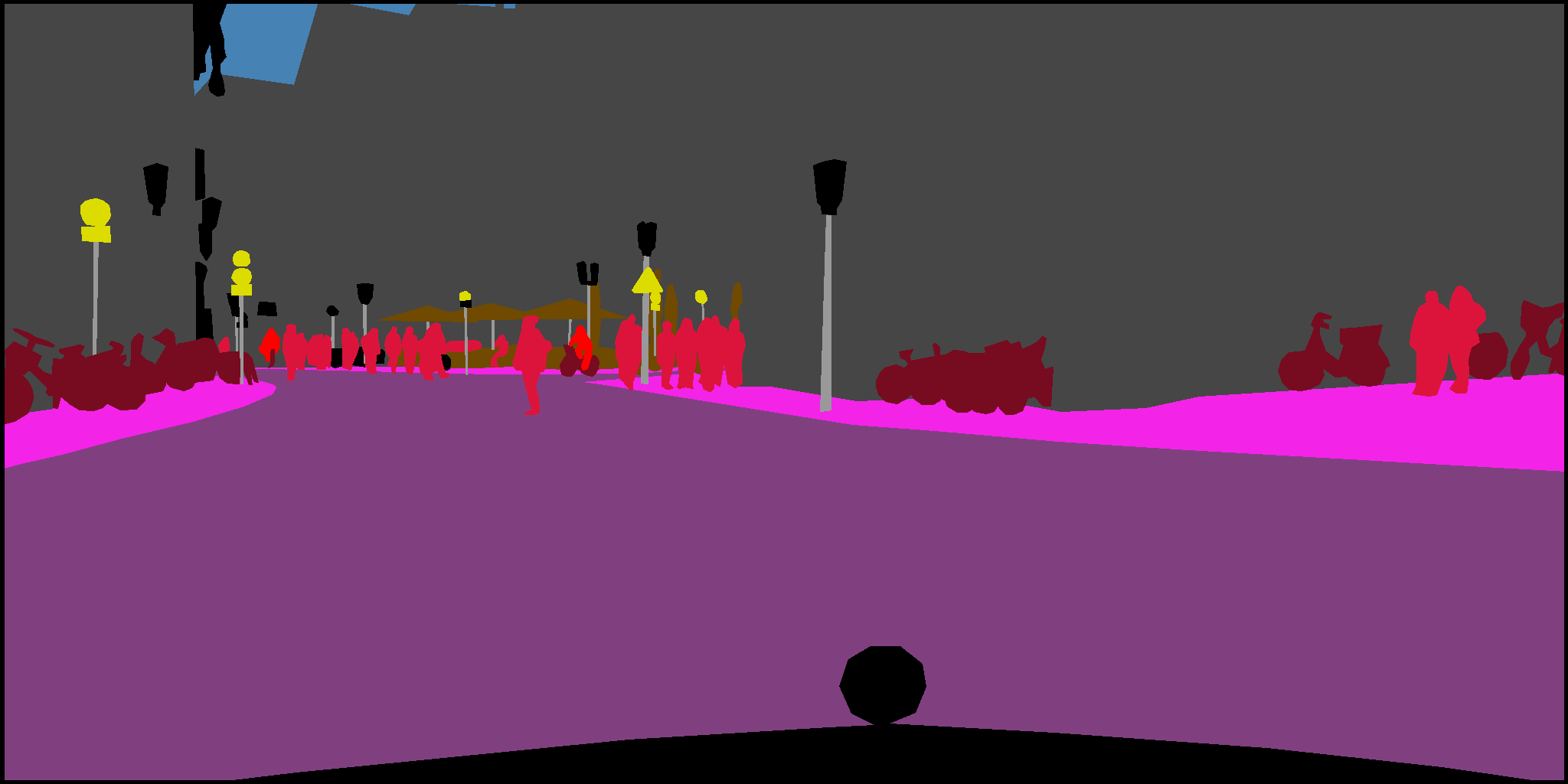}&
\hspace{2mm} \includegraphics[width=0.15\linewidth]{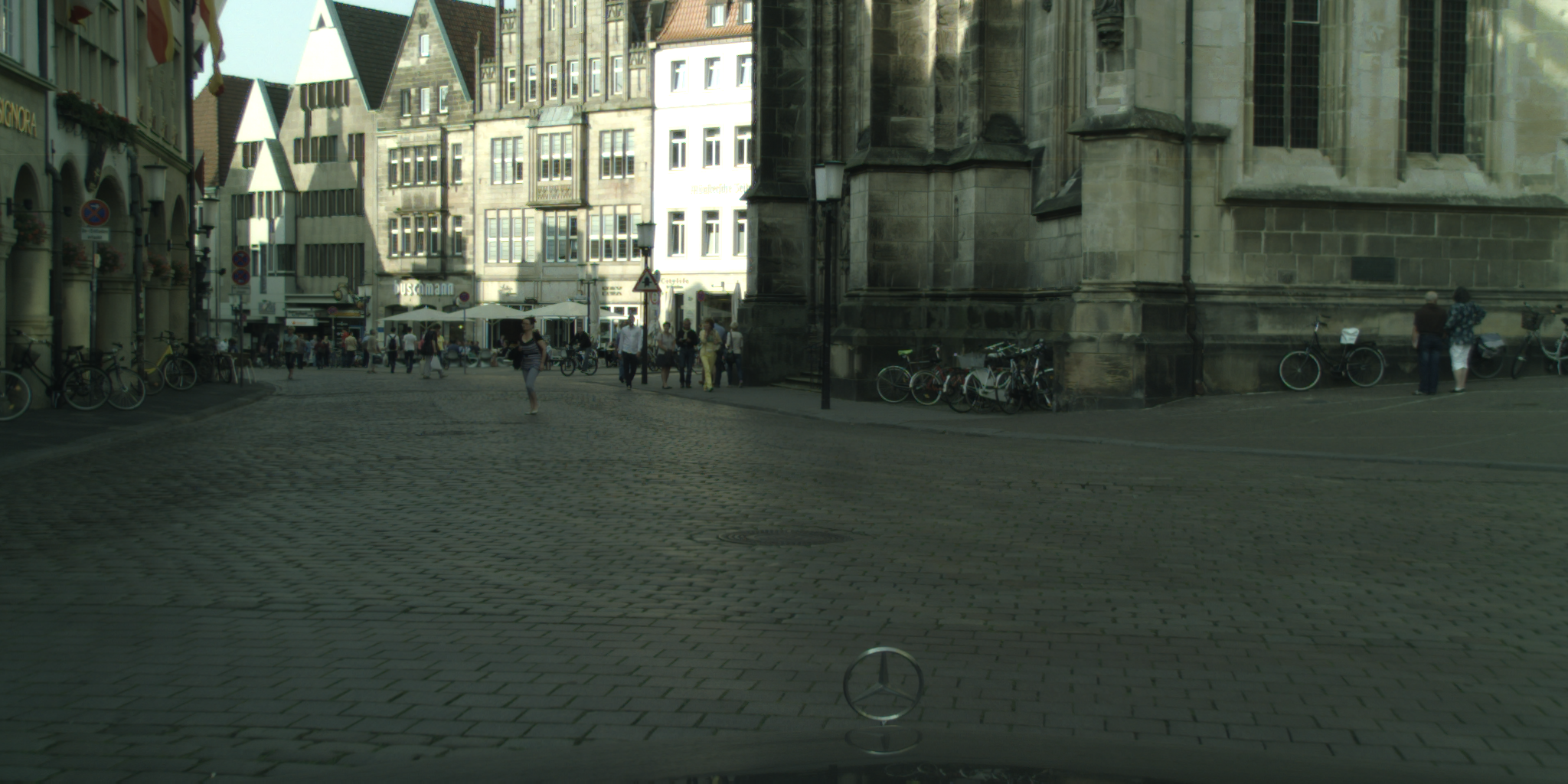} &
\hspace{2mm} \includegraphics[width=0.15\linewidth]{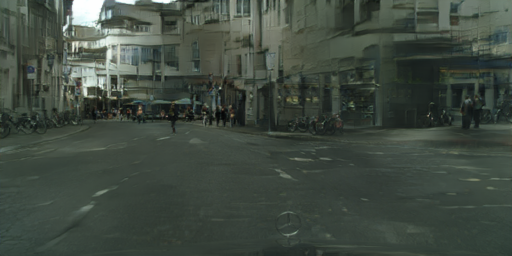} &   \hspace{2mm} \includegraphics[width=0.15\linewidth]{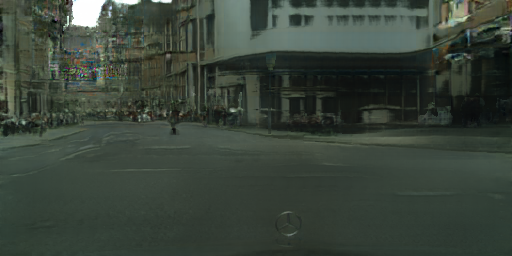} &
\hspace{2mm} \includegraphics[width=0.15\linewidth]{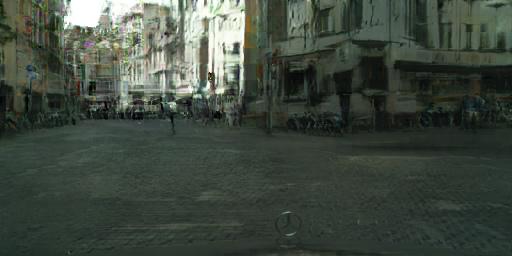} &  
\hspace{2mm} \includegraphics[width=0.15\linewidth]{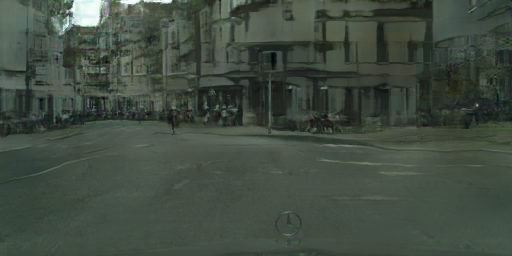}
\end{tabular}
	\caption{\textbf{Image generation of different methods on Cityscapes.} Here we compare generated results of our methods TLAM and Sparse-TLAM to SPADE and ASAP-Net.}
	\label{fig:cityscapes}
\end{figure*}